\documentclass[journal,onecolumn]{IEEEtran}
\ifCLASSINFOpdf
\else
\fi
\usepackage{graphics} 
\usepackage{amsmath} 
\usepackage{amssymb}  
\usepackage{textglos}
\usepackage[utf8]{inputenc} 
\usepackage[T1]{fontenc}    
\usepackage{hyperref}       
\usepackage{url}            
\usepackage{booktabs}       
\usepackage{amsfonts}       
\usepackage{nicefrac}       
\usepackage{microtype}      
\usepackage{lipsum}
\usepackage{tabularx} 
\usepackage{amsmath}  
\usepackage{graphicx} 
\usepackage[margin=1in,letterpaper]{geometry} 
\usepackage{cite} 
\usepackage{titlesec}
\usepackage{mathtools, amssymb, nccmath}
\usepackage{wasysym}
\usepackage{algorithm}
\usepackage{algorithmic}
\usepackage{multicol}

\usepackage{enumitem}
\usepackage{float}
\usepackage{mathtools}
\usepackage{amsmath,amsfonts,amssymb}
\usepackage{cmll}
\usepackage{tensor}
\usepackage{booktabs}
\usepackage{tabularx}
\usepackage{makecell}

\usepackage{cancel}

\usepackage[font=small,skip=5pt]{caption}

\newcolumntype{Z}{ >{\centering\arraybackslash}X }

\newcommand{\crff}{\,\overline{\!\times\!}{}^{\,*}}

\usepackage{scalerel}
\newcommand{\timesM}{{\tilde{\smash[t]{\times}}}}
\newcommand{\timesfM}{\timesM {}^*}
\newcommand{\crffM}{\scaleobj{.87}{\,\tilde{\bar{\smash[t]{\!\times\!}}{}}}{}^{\,*}}

\newcommand{\timesf}{{\times}^*}

\newcommand{\mysp}{.7ex}

\newcommand{\M}{\mathcal{M}}
\newcommand{\I}{\mathcal{I}}
\newcommand{\IC}[1]{\I_{#1}^{C}}
\newcommand{\BC}[1]{\B_{#1}^{C}}

\newcommand{\T}{^\top}
\newcommand{\Tten}{^{\widetilde{\top}}} 
\newcommand{\Rten}{^{\widetilde{\mathrm{R}}}} 
\newcommand{\RTten}{^{\widetilde{\mathrm{R}},\widetilde{\top}}} 

\newcommand{\J}{\boldsymbol{J}{}}
\newcommand{\Jc}{\J_{c}}
\newcommand{\Jdc}{\dot{\J_{c}}}

\newcommand{\w}{\rmvec{w}}

\newcommand{\R}{\mathbb{R}}
\newcommand{\C}{\vC}

\newcommand{\rmvec}[1]{\boldsymbol{#1}}
\newcommand{\greekvec}[1]{\boldsymbol{#1}}

\usepackage{xcolor}

\newcommand{\red}[1]{{\color{red}#1}}

\newcommand{\blue}[1]{{\color{blue}#1}}

\newcommand{\B}{\boldsymbol{B}{}}
\renewcommand{\C}{\boldsymbol{C}}
\renewcommand{\M}{\boldsymbol{M}}

\newcommand{\A}{\boldsymbol{A}{}}

\renewcommand{\I}{\boldsymbol{I}{}}
\newcommand{\g}{\rmvec{g}}
\newcommand{\q}{\rmvec{q}}

\newcommand{\U}{\rmvec{U}}
\newcommand{\V}{\rmvec{V}}
\newcommand{\F}{\rmvec{F}}
\newcommand{\W}{\rmvec{W}}

\renewcommand{\b}{\rmvec{b}}
\renewcommand{\u}{\rmvec{u}}

\newcommand{\m}{\rmvec{m}}

\newcommand{\f}{\rmvec{f}}

\renewcommand{\v}{\rmvec{v}}
\newcommand{\vlam}[1]{\rmvec{v}_{\lambda(#1)}}
\newcommand{\alam}[1]{\rmvec{a}_{\lambda(#1)}}

\newcommand{\vJ}[1]{\v_{J_{#1}}}

\renewcommand{\t}{\rmvec{t}}

\renewcommand{\a}{\rmvec{a}}
\newcommand{\ag}[1]{\a_{g_{#1}}}

\newcommand{\qd}{\dot{\q}}
\newcommand{\qdd}{\ddot{\q}}

\newcommand{\psibar}{\greekvec{\psi}}
\newcommand{\psibardot}{\,\dot{\!\psibar}}
\newcommand{\Psibar}{\greekvec{\Psi}}
\newcommand{\Psibardot}{\,\dot{\!\Psibar}}
\newcommand{\Psibarddot}{\,\ddot{\!\Psibar}}
\newcommand{\psibarddot}{\,\ddot{\!\psibar}}

\newcommand{\phibar}{\boldsymbol{s}}
\newcommand{\Phibar}{\boldsymbol{S}}
\newcommand{\phibardot}{\,\dot{\!\phibar}}
\newcommand{\Phibardot}{\,\dot{\!\Phibar}}

\newcommand{\taubar}{\greekvec{\tau}}
\newcommand{\gammabar}{\greekvec{\gamma}}

\newcommand{\xibar}{\greekvec{\xi}}

\newcommand{\subtree}{\nu}
\newcommand{\subtreeb}{\overline{\nu}}

\newcommand{\dtaudqSO}[3]{\frac{\partial ^{2} \taubar_{#1}}{\partial \q_{#2} \partial \q_{#3}}}

\usepackage{setspace}

\newcommand{\Ff}{\F}
\newcommand{\Uu}{\U}
\newcommand{\Uv}{\V}

\newcommand{\Bmat}[2]{\B(#1,#2)}
\newcommand{\Bten}[2]{\mathcal{B}(#1,#2)}
\newcommand{\K}{\boldsymbol{K}{}}
\renewcommand{\S}{\mathbf{S}\T}
\newcommand{\lambdah}{\hat{\lambda}}

\begin{document}
%
\title{Details of Partial Derivatives of Rigid-Body Inverse Dynamics}
%
%
%
\author{ Shubham Singh$^{1}$, Ryan P. Russell$^{2}$ and Patrick M.~Wensing$^{3}$
\thanks{$^{1}$Graduate Research Assistant, Aerospace Engineering, The University of Texas at Austin, TX-78751, USA. \href{mailto:singh281@utexas.edu}{singh281@utexas.edu}}
       
     \thanks{$^{2}$Professor, Aerospace Engineering, The University of Texas at Austin, TX-78751, USA. \href{mailto:ryan.russell@utexas.edu}{ryan.russell@utexas.edu} 
           }%
       \thanks{$^{3}$Associate Professor, Aerospace \& Mechanical Engineering, University of Notre Dame, IN-46556, USA. \href{mailto:pwensing@nd.edu}{pwensing@nd.edu}
       }%
    
}

%



\maketitle

\begin{abstract}
The details of second-order partial derivatives of rigid-body Inverse/Forward dynamics are provided. Several properties and identities using Spatial Vector Algebra are listed, along with their detailed derivations. The expressions build upon previous work by the author on first-order partial derivatives of inverse dynamics. The first/second-order derivatives are also extended for systems with external forces. Finally, the KKT Forward dynamics and Impact dynamics derivatives are derived.
\end{abstract}

%
\IEEEpeerreviewmaketitle
\section{Notations, Symbols and Acronyms}
\label{notation}

Cartesian 3D  vectors are denoted by lower-case letters with an overhead bar $(\bar{v})$. Spatial vectors are denoted by lower-case bold letters ($\a$). Matrices are denoted by upper-case bold letters ($\M$), while third-order tensors are denote by upper-case calligraphic letters ($\mathcal{A}$).\\ \\ 
Acronyms:
\begin{enumerate}
    \item RHS : Right-hand-side
    \item w.r.t : with respect to
    \item FO : first-order
    \item SO : second-order
    \item SVA : Spatial Vector Algebra
\end{enumerate}
\vspace{5ex}
Section~\ref{sva_props} is an introduction on Spatial Vector Algebra (SVA)~\cite{Featherstone08}, also covered in Ref.~\cite[App A]{arxivss}. Section~\ref{sma_props} provides an extension of SVA for spatial motion matrices, and spatial cross-product operators associated with them. Then, the properties of spatial matrices (numbered K1-K16) are listed and derived in detail. Then, Sections~\ref{SO_q_sec} and \ref{SO_qd_sec} provide the derivations of SO partial derivatives of ID w.r.t $\q$ and $\qd$ respectively. Section~\ref{MSO_sec} provides the SO cross-derivatives w.r.t $\q$ and $\qd$, while Section~\ref{MFO_sec} gives the FO partial derivatives of $\M(\q)$ w.r.t $\q$. In the end, Section~\ref{eff_impl_sec} provides the matrix and vector form of expressions used in the Algorithm developed. Section~\ref{sec_SO_eqn} proves a SO relation between ID and FD derivatives. Section~\ref{sec_ID_FD_contacts} extends the ID FO/SO derivatives for systems with external forces. Section~\ref{sec_KKT_derivs} amd \ref{sec_impact} gives the FO/SO derivatives of the KKT Forward dynamics and impact dynamics respectively.

\section{Spatial Vector Algebra Identities and Properties}
\label{sva_props}
\subsection{Spatial Vector Algebra}
\label{sva_intro}
A body $k$ with spatial velocity ${}^{k}\v_{k} \in M^{6}$ in the body frame is decomposed in its angular and linear components as:

  \begin{align}
    {}^{k}\v_{k} &= \begin{bmatrix}
           {}^{k}\bar{\omega}_{k} \\
           {}^{k}\bar{v}_{k}
         \end{bmatrix}
  \end{align}
where ${}^{k}\bar{\omega}_{k} \in \mathbb{R}^{3}$ is the angular velocity of the body in a coordinate frame fixed to the body, while ${}^{k}\bar{v}_{k} \in \mathbb{R}^{3}$ is the linear velocity of the body-fixed point at the origin of the body frame. Spatial vectors can also be expressed in the ground frame. For example, the spatial velocity of the body $k$ in the ground frame is denoted as ${}^{0}\v_{k}$. In this case, the linear velocity is associated with the body-fixed point on body $k$ that is coincident with the origin of the ground frame. The net spatial force ${}^{0}\f_{k} \in F^{6}$ defined in Eq.~\ref{fvec_defn} on the body can be calculated from the spatial equation of motion (Eq.\ref{sp_eqn_of_motion}):
  \begin{align}
    {}^{0}\f_{k} &= \begin{bmatrix}
           {}^{0}\bar{n}_{k} \\
           {}^{0}\bar{f}_{k}
         \end{bmatrix}
         \label{fvec_defn} \\
       {}^{0}\f_{k} &= {}^{0}\I_{k}{}^{0}\a_{k} + {}^{0}\v_{k} \times^*{}^{0}\I_{k} {}^{0}\v_{k}
       \label{sp_eqn_of_motion}
\end{align}
where ${}^{0}\bar{n}_{k} \in \mathbb{R}^{3}$ is the net moment on the body about the origin of the ground frame, $ {}^{0}\bar{f}_{k} \in \mathbb{R}^{3}$ is the net linear force on body, ${}^{0}\I_{k}$ is the spatial inertia of the body $k$ that maps motion vectors to force vectors, and ${}^{0}\a_{k} \in M^{6}$ is the spatial acceleration of the body. The transformation matrix ${}^{i}\boldsymbol{X}_{j}$ is used to transform vectors in frame $j$ to frame $i$ is defined as:
\begin{align}
    {}^{i}\boldsymbol{X}_{j} = 
    \begin{bmatrix}
        {}^{i}\boldsymbol{R}_{j} & \bf{0}  \\
       -{}^{i}\boldsymbol{R}_{j} (\bar{p}_{i/j} \times) &  {}^{i}\boldsymbol{R}_{j} 
    \end{bmatrix}
\end{align}
where ${}^{i}\boldsymbol{R}_{j} \in \mathbb{R}^{3 \times 3}$ is the rotation matrix from frame $j$ to frame $i$, $\bar{p}_{i/j} \in \mathbb{R}^{3}$ is the Cartesian vector from origin of frame $j$ to $i$, and $\bf{0}$ is the 3 $\times$ 3 zero matrix. $\bar{p} \times$ is the 3D vector cross product on the elements of $\bar{p}$, defined as:
\begin{align}
    \bar{p} \times =   \begin{bmatrix}
        0 & -p_{z}  & p_{y}\\
        p_{z} & 0  & -p_{x}\\
        -p_{y} & p_{x}  & 0
    \end{bmatrix}
\end{align}

\noindent The spatial transformation matrix ${}^{0}\boldsymbol{X}_{k}$ can be used to obtain spatial velocity vector ${}^{0}\v_{k}$ from the vector ${}^{k}\v_{k}$ as:

\begin{equation}
    {}^{0}\v_{k} = {}^{0}\boldsymbol{X}_{k} {}^{k}\v_{k}
\end{equation}

\noindent A spatial cross-product operator between two motion vectors ($\v$,$\u$), written as $(\v \times) \u$, is given by (Eq.~\ref{cross_defn}). This operation can be understood as providing the time rate of change of $\u$, when $\u$ is moving with a spatial velocity $\v$. A spatial cross-product between a motion and a force vector is written as $(\v \times^*) \f$, and defined in Eq.~\ref{cross_f_defn}. 
\begin{align}
    \v \times = \begin{bmatrix}
        \bar{\omega} \times & \bf{0}  \\
       \bar{v} \times &  \bar{\omega} \times
    \end{bmatrix}
    \label{cross_defn}
\end{align}

\begin{align}
    \v \times^* = \begin{bmatrix}
        \bar{\omega} \times & \bar{v} \times  \\
        \bf{0} &  \bar{\omega} \times
    \end{bmatrix}
    \label{cross_f_defn}
\end{align}

\noindent An operator $\crff$ (Eq.~\ref{crff_oper}) is  defined by swapping the order of the cross product, such that $(\f \crff) \v = (\v \times ^*) \f $~\cite{echeandia2021numerical}. 

\begin{align}
    \f \crff = \begin{bmatrix}
        -\bar{n} \times & -\bar{f} \times  \\
        -\bar{f} \times & \bf{0}
    \end{bmatrix}
    \label{crff_oper}
\end{align}

\noindent Hence, the three spatial vector cross-products can be considered as binary operators using the above definitions to map between motion and force vector spaces as~\cite{Featherstone08}:  
\begin{equation}
\begin{aligned}
&    \times : M^{6} \times M^{6} \xrightarrow[]{} M^{6} \\
&    \times^{*} : M^{6} \times F^{6} \xrightarrow[]{} F^{6} \\
&    \crff : F^{6} \times M^{6} \xrightarrow[]{} F^{6}
\end{aligned}
\end{equation}

\subsection{Kinematics}
\noindent The spatial velocity of a body $i$ in a connectivity tree in given as:

\begin{equation}
    \v_{i} = \sum_{l \preceq i}\vJ{l}
\end{equation}
where $\vJ{l}$ is the spatial joint velocity, given as $\vJ{l} = \Phibar_{l}\qd_{l}$. The quantity $\Phibar_{l}$ is the joint motion sub-space matrix for a multi-DoF joint $l$ that precedes joint $i$ (i.e., $l\preceq i$) in the kinematic tree. The kinematic quantities $\a_{i}$, $\gammabar_{i}$, and $\xibar_{i}$ are defined in Ref.~\cite[Sec III]{singh2022efficient}:
\begin{equation}
\begin{aligned}
 & \a_{i} = \textstyle\sum_{l \preceq i} \big( \Phibar_{l} \qdd_{l} + \v_{l} \times \Phibar_{l} \qd_{l} \big)+\a_{0}\,. \\
 & \gammabar_{i} = \textstyle \sum_{l \preceq i} \Phibar_{l} \qdd_{l} \\
 & \xibar_{i} = \textstyle \sum_{l \preceq i} \v_{l} \times \Phibar_{l} \qd_{l}  
\end{aligned}
\end{equation}

\noindent Two more spatial quantities $\Psibardot_{i}$, and $\Psibarddot_{i}$ defined in Ref.~\cite[Sec IV]{singh2022efficient} are:
\begin{align}
\begin{split}
  \Psibardot_{j} = & \v_{\lambda(j)} \times \Phibar_j \\
 \Psibarddot_{j} = & \a_{\lambda(j)} \times \Phibar_{j} + \v_{\lambda(j)} \times \Psibardot_{j}
\end{split}
\label{psidot_psidotdot_defn}
\end{align}
A body-level Coriolis matrix $\B_{k}$ is defined~\cite{singh2022efficient} as:
 
\begin{equation}
    \B_{k} = \frac{1}{2} \big(\big(\v_{k}\times^* \big)\I_{k} - \I_{k} \big(\v_{k} \times   \big) + \big(\I_{k} \v_{k} \big) \crff    \big)
    \label{bl_term_defn}
\end{equation}

If in Eq.~\ref{bl_term_defn}, instead of $\v_{k}$, some other spatial quantity $\m$ is used, then the body-Coriolis matrix is denoted with a square bracket as $\Bmat{\I_{k}}{\m}$, and defined as:

\begin{equation}
    \Bmat{\I_{k}}{\m} = \frac{1}{2} \big(\big(\m \times^* \big)\I_{k} - \I_{k} \big(\m \times   \big) + \big(\I_{k} \m \big) \crff    \big)
    \label{bkm_term_defn}
\end{equation}

Composite term for $\Bmat{\I_{k}}{\m}$ is $\BC{k}[\m] = \sum_{l \succeq k} \Bmat{\I_l}{\m}$. 

\subsection{Dynamics}
\noindent The generalized force vector on a joint $i$ can be calculated as:

\begin{equation}
    \taubar_{i} = \Phibar_{i}\T\f{}_{i}^{C}
\end{equation}
where, $\f{}_{i}^{C} = \sum_{k\succeq i} \f_k$ is the net spatial force transmitted across joint $i$. The first-order partial derivatives of $\taubar_{i}$ with respect to joint configuration $(\q)$ and velocity ($\qd$) for the case $j \preceq i$ were derived in Ref.~\cite[Sec.~IV]{singh2022efficient} as:
\begin{align}
     \frac{\partial {\taubar_{i}}}{\partial \q_{j}} &=  { \Phibar_{i}\T \big[ 2  \B_{i}^{C} \big] \Psibardot{}_{j} + \Phibar_{i}\T  \I_{i}^{C} \Psibarddot{}_{j} } 
     \label{dtau_dq_1} \\
    \frac{\partial {\taubar_{j}}}{\partial {\q_{i}}} &=  {\Phibar_{j}\T [ 2 \B_{i}^{C} \Psibardot_{i} +  \I_{i}^{C}  \Psibarddot_{i}+(\f_{i}^{C} )\crff \Phibar_{i}  ] (j \neq i)} 
     \label{dtau_dq_2} \\
     \frac{\partial {\taubar_{i}}}{\partial {\dot{\q}_{j}}} &= {\Phibar_{i}\T \Big[2 \B_{i}^{C} \Phibar_{j} +  \I_{i}^{C} (    \Psibardot_{j} + \Phibardot_{j} ) \Big]  }
         \label{dtau_dqd_1} \\
    \frac{\partial {\taubar_{j}}}{\partial {\dot{\q}_{i}}}  &= {\Phibar_{j}\T \Big[2 \B_{i}^{C} \Phibar_{i} +  \I_{i}^{C} (    \Psibardot_{i} + \Phibardot_{i} ) \Big]  (j \neq i)}
         \label{dtau_dqd_2}
\end{align}
where, the quantities $\IC{i}= \sum_{k\succeq i} \I_k$, $\BC{i}= \sum_{k\succeq i} \B_k$ are the quantities defined for an entire sub-tree.

\subsection{Properties of Spatial Vectors}
\label{prop_vec}
\noindent Assuming $\u,\v,\m \in M^{6}$, and $\f \in F^{6}$, many spatial vector properties~\cite{Featherstone08} are utilized herein:

\begin{enumerate}[label=P{{\arabic*}}.]
    \item $ \u \times \v = -\v\times \u$
    \item $(\v \times \m) \times = (\v \times )( \m \times) - (\m \times)(\v \times)$ 
    \item $(\v \times \m) \times^{*} = (\v \times^{*} )( \m \times^*) - (\m \times^*)(\v \times^*)$ 
    \item $(\v \times^{*} \f) \crff = (\v \times^{*} )(\f \crff) - (\f \crff)(\v \times)$ 
    \item $(\u \times \v)\T\f = -\v^T (\u \times^* \f)  $
    \item $(\u \times^* \f)^T \v = -\f^T(\u \times \v)$
    \item $\u^T(\v \times^*)\f = \f^T(\u \times \v)$
    \item $(\u \times \v)^T  = -\v\T(\u \times^*) $
    \item $(\u \times^* \f) ^T = -\f ^T \u \times$
    \item $\u \times \v \times \m = \u \times (\v \times \m)$ 
\end{enumerate}

\subsection{Identities by Perturbing All DoFs of a multi-DoF Joint}

\noindent Ref.~\cite{singh2022efficient} provides the following identities with their derivations in Ref.~\cite{arxivss}. These give partial derivatives of some kinematic and dynamic identities by perturbing the full configuration ($\q$) or velocity vector ($\qd$)  of a multi-DoF joint.

\begin{enumerate}[label=J{{\arabic*}}]

         \item\label{J1}$\frac{\partial \Phibar_{i}}{\partial q_{j,p}}=
        \begin{cases}
          \phibar_{j,p} \times \Phibar_{i} , &~~~~~~~~~~~~~~~~~~~~ \text{if}\ j \preceq i \\
          0, &~~~~~~~~~~~~~~~~~~~~ \text{otherwise}
        \end{cases}$

       \item \label{J2}$ \frac{\partial (\v_{i} \times^{*} \f)}{\partial \q_{j}}=
        \begin{cases}
         \f \crff \big(\v_{\lambda(j)} - \v_{i})  \times \Phibar_{j} \big) ,  & \text{if}\ j \preceq i \\
          0,  & \text{otherwise}
        \end{cases}$
   
      \item \label{J3}$ \frac{\partial (\Phibar_{i}\T \f)}{\partial \q_{j}}=
    \begin{cases}
      -\Phibar_{i}\T (\f \crff  \Phibar_{j}) , &~~~~~~~~~~~~ \text{if}\ j \preceq i \\
      0, &~~~~~~~~~~~~ \text{otherwise}
    \end{cases}$

     \item\label{J4}$\frac{\partial (\I_{i} \a)}{\partial \q_{j}} =
    \begin{cases}
     (\I_{i} \a) \crff \Phibar_{j} +\I_{i}(\a \times \Phibar_{j})   , & \text{if}\ j \preceq i \\
      0, & \text{otherwise}
    \end{cases}$

     \item \label{J5}$\frac{\partial (\I_{i} \v_{i})}{\partial \q_{j}}  =
    \begin{cases}
    (\I_{i} \v_{i}) \crff \Phibar_{j} + \I_{i}(\Psibardot_{j})  , &~~~ \text{if}\ j \preceq i \\
      0, &~~~ \text{otherwise}
    \end{cases}$

     \item \label{J6}$\frac{\partial \xibar_{i}}{\partial \q_{j}}=
    \begin{cases}
       (\v_{\lambda(j)} - \v_{i} )  \times  \Psibardot_{j} + \\
       ~~( \xibar_{\lambda(j)} - \xibar_{i}  ) \times  \Phibar_{j}   , &~~~~~~~~~~ \text{if}\ j \preceq i \\
      0, &~~~~~~~~~~ \text{otherwise}
    \end{cases}$

       \item \label{J7}$\frac{\partial \gammabar_{i}}{\partial \q_{j}}=
    \begin{cases}
       (\gammabar_{\lambda(j)} - \gammabar_{i})  \times \Phibar_{j} , &~~~~~~~~~~~~ \text{if}\ j \preceq i \\
      0, &~~~~~~~~~~~~ \text{otherwise}
    \end{cases}$
    
        \item \label{J8}$\frac{\partial \v_{i}}{\partial \qd_{j}}=
    \begin{cases}
          \Phibar_{j}, &~~~~~~~~~~~~~~~~~~~~~~~~~~~~~~ \text{if}\ j \preceq i \\
      0, &~~~~~~~~~~~~~~~~~~~~~~~~~~~~~~ \text{otherwise}
    \end{cases}$  
    
     \item \label{J9}$\frac{\partial \xibar_{i}}{\partial \qd_{j}}=
    \begin{cases}
    \Psibardot_{j} + \Phibardot_{j}-\v_{i}  \times \Phibar_{j}, &~~~~~~~~~~ \text{if}\ j \preceq i \\
      0, &~~~~~~~~~~ \text{otherwise}
    \end{cases}$
    
\end{enumerate}

\section{Tensorial Identities and Properties}
\label{sma_props}
The motion space $M^{6}$~\cite{Featherstone08} is extended to a space of spatial-motion matrices $M^{6\times n}$, where each column of such a matrix is a usual spatial-motion vector.
For any $\U \in M^{6 \times n}$ a new spatial cross-product operator $\U \timesM$ is considered,
and defined by carrying out the usual spatial cross-product operator ($\times$) on each column of $\U$. The result is a third-order tensor in $\R^{6 \times 6 \times n}$, where each $6 \times 6$ matrix in the 1-2 dimension is a result of the spatial cross-product operator on a column of $\U$. 

Given two spatial motion matrices, $\V \in M^{6 \times n_v}$ and $\U \in M^{6 \times n_u}$, we can now define a cross-product operation between them as $(\V \timesM) \U \in M^{6 \times n_u \times n_v}$ via a tensor-matrix product. Such an operation, denoted as $\mathcal{Z} = \mathcal{A}B$ is defined as:
\begin{equation}
    \mathcal{Z}_{i,j,k} = \sum_{\ell}\mathcal{A}_{i,\ell,k}\\B_{\ell,j}
\end{equation}
for any tensor $\mathcal{A}$ and suitably sized matrix $\mathbf{B}$. Thus, the $k$-th page, $j$-th column of $(\V \timesM) \U$ gives the cross product of the $k$-th column of $\V$ with the $j$-th column of $\U$.

In a similar manner, consider a spatial force matrix $\mathbf{F}\in F^{6 \times n_f}$. Defining $(\V\timesfM)$ in an analogous manner to in allows taking a cross-product-like operation $\V\timesfM \mathbf{F}$. Again, analogously, we consider a third operator $(\mathbf{F}\crffM)$ that provides  $\V\timesfM \mathbf{F} = \mathbf{F}\crffM \V$. In each case, the tilde indicates the spatial-matrix extension of the usual spatial-vector cross products.

For later use, the product of a matrix $\B \in \R^{n_{1} \times n_{2}}$, and a tensor $\mathcal{A}\in \R^{n_{2} \times n_{3} \times n_{4}}$, likewise results in another tensor, denoted as $\mathcal{Y} = \B\mathcal{A}$, and defined as:
\begin{equation}
    \mathcal{Y}_{i,j,k} = \sum_{\ell}\B_{i,\ell}\mathcal{A}_{\ell,j,k}
\end{equation}

Two types of tensor rotations are defined for this paper:
\begin{enumerate}
    \item  $\mathcal{A}\Tten$: Transpose along the 1-2 dimension. This operation can also be understood as the usual matrix transpose of each matrix  moving along the third dimension in the tensor. If $ \mathcal{A}\Tten = \mathcal{B}$, then $\mathcal{A}_{i,j,k} = \mathcal{B}_{j,i,k}$.
    \item  $\mathcal{A}\Rten$: Rotation of elements along the 2-3 dimension. If $ \mathcal{A}\Rten = \mathcal{B}$, then $\mathcal{A}_{i,j,k} = \mathcal{B}_{i,k,j}$.
\end{enumerate}
Another rotation is defined as the combination of $(\Rten)$ followed by $(\Tten)$. For example, if $\mathcal{A}\RTten = \mathcal{B}$, then $\mathcal{A}_{i,j,k} = \mathcal{B}_{k,i,j}$.

\subsection{Properties of Spatial Matrices}
\label{prop_mat}
Some additional properties using the operators defined above are as follows. These properties are a natural extension of the spatial vector properties from Sec.~\ref{prop_vec}. Assuming $\v \in M^{6}$,$\f \in F^{6}$, $\U \in M^{6 \times n}$, $\W \in M^{6 \times p}$, $\Ff \in F^{6 \times m} $, $\Uv \in M^{6 \times l}$, $\A \in \R^{n_{1} \times n_{2}}$, $\I, \B \in \R^{6 \times 6}$, $\mathcal{Y} \in \R^{n_{2} \times n_{3} \times n_{4}}$, $\mathcal{B} \in \R^{6\times 6 \times n_{5}}$:

\begin{enumerate}[label=M{{\arabic*}}]
 \setlength\itemsep{.15em}
      \item \label{m1} $\U \timesfM = -(\U \timesM)\Tten$ 
    \item \label{m2} $-\Uv\T (\Uu \timesfM) = (\Uu \timesM \Uv)\Tten $
    \item \label{m3} $-\Uv\T (\Uu \timesfM) \Ff= (\Uu \timesM \Uv)\Tten \Ff$
    \item \label{m4} $ (\U \timesM \v)\Rten = - \v \times \U $ 
    \item \label{m5} $\U \timesfM \Ff = ( \Ff \crffM \U)\Rten$ 
    \item \label{m6} $\Ff \crffM \U = (\U \timesfM \Ff)\Rten $ 
    \item \label{m7} $(\lambda \U) \timesM = \lambda(\U \timesM)$
    \item \label{m8} $\Uu \timesM \Uv = - (\Uv \timesM \Uu)\Rten $ 
    \item \label{m9} $(\v \times \U) \timesM = \v \times \U \timesM - \U \timesM \v \times$ 
    \item \label{m10} $(\v \times \U) \timesfM = \v \times^* \U \timesfM - \U \timesfM \v \times^*$
    \item \label{m11} $((\U \timesfM \f)\Rten) \crffM = \U \timesfM \f \crff - \f \crff \U \timesM$
    \item \label{m12} $(\U \timesfM \Ff)\Tten = -\Ff\T (\U \timesM)$
    \item \label{m13} $\Uv\T (\Uu \timesfM \Ff) = (\Uv \timesM \Uu)\RTten \Ff=  (\Ff\T (\Uv \timesM \Uu)\Rten)\Tten $
    \item \label{m14} $ \v \times^* \Ff = (\Ff \crffM \v)\Rten$
    \item \label{m15} $ \f \crff \U = (\U \timesfM \f)\Rten$
     \item \label{m16} $\Uv\T (\Uu \timesfM \Ff)\Rten = \big[(\Uv \timesM \Uu)\RTten \Ff\big]\Rten$
          \item \label{m17} $\Uv\T (\Uu \timesfM \Ff)\Rten =-[\Uu\T (\Uv \timesfM \Ff)\Rten]\Tten$
      \item \label{m18} $\I  (\Uu \timesfM \Ff)\Rten = [ \I (\Uu \timesfM \Ff)]\Rten$ 
        \item \label{m19} $\I  (\Uu \timesM \Uv)\Rten = [ \I (\Uu \timesM \Uv)]\Rten$ 
    \item \label{m20} $(\A\mathcal{Y})\Tten = \mathcal{Y}\Tten \A\T$
            \item \label{m21} $\Ff\T(\U \timesM \V) = -[\V\T(\Ff\crffM\U)\Rten]\Tten$
       \item \label{m22} $\Bmat{\I}{\v}\T\w = -\Bmat{\I}{\w}\T\v$
    \item \label{m23}  $\Bmat{\I}{\v}\w= \Bmat{\I}{\w}\v-\I(\v\times \w)$
    \item \label{m24} $\u\T[\Bmat{\I}{\v}\w] = -\v\T[\Bmat{\I}{\u}\w]$
       \item \label{m25} $\Bten{\I}{\V}\Tten\W = -[\Bten{\I}{\W}\Tten\V]\Rten$
          \item \label{m26} $\Bten{\I}{\V}\W = [\Bten{\I}{\W}\V]\Rten - \I(\V\timesM)\W$
          \item \label{m27} $\U\T[\Bten{\I}{\V}\W]\Rten = -\Big[\V\T[\Bten{\I}{\U}\W]\Rten\Big]\Tten$
\end{enumerate}

\vspace{5ex}

\noindent Using the properties and spatial matrix operators defined above, some more identities are derived as follows:

\hfill 

\begin{enumerate}[label=K{\arabic*}]

        \item\label{dphii} $\frac{\partial \Phibar_{i}}{\partial \q_{j}}=
    \begin{cases}
          \Phibar_{j}   \timesM \Phibar_{i}, &~~~~~~~~~~~~~~~~~~~~~~~~~~~~~~~~~~~~~~~~~~~ \text{if}\ j \preceq i \\
      0,  &~~~~~~~~~~~~~~~~~~~~~~~~~~~~~~~~~~~~~~~~~~~ \text{otherwise}
    \end{cases}$

        \item \label{dphii_dot} $\frac{\partial \Phibardot_{i}}{\partial \q_{j}}=
    \begin{cases}
          \Psibardot_{j}   \timesM \Phibar_{i}+ \Phibar_{j} \timesM \Phibardot_{i}, &~~~~~~~~~~~~~~~~~~~~~~~~~~~~~~ \text{if}\ j \preceq i \\
      0, &~~~~~~~~~~~~~~~~~~~~~~~~~~~~~~ \text{otherwise}
    \end{cases}$
    
        \item \label{vji_phii} $\frac{\partial (\vJ{i} \times \Phibar_{i})}{\partial \q_{j}}=
    \begin{cases}
          \Phibar_{j}   \timesM (\vJ{i} \times \Phibar_{i}), &~~~~~~~~~~~~~~~~~~~~~~~~~~ \text{if}\ j \preceq i \\
      0, &~~~~~~~~~~~~~~~~~~~~~~~~~~ \text{otherwise}
    \end{cases}$
    
       \item \label{psii_dot} $\frac{\partial \Psibardot_{i}}{\partial \q_{j}}=
    \begin{cases}
          \Psibardot_{j}   \timesM  \Phibar_{i} + \Phibar_{j} \timesM \Psibardot_{i}, &~~~~~~~~~~~~~~~~~~~~~~~~~~~~~~ \text{if}\ j \preceq i \\
      0, &~~~~~~~~~~~~~~~~~~~~~~~~~~~~~~ \text{otherwise}
    \end{cases}$  
    
    \item \label{Ii} $\frac{\partial \I_{i}}{\partial \q_{j}}=
    \begin{cases}
      \Phibar_{j} \timesfM \I_{i}- \I_{i}(\Phibar_{j} \timesM),  &~~~~~~~~~~~~~~~~~~~~~~~~~~~~ \text{if}\ j \preceq i \\
      0, &~~~~~~~~~~~~~~~~~~~~~~~~~~~~ \text{otherwise}
    \end{cases}$ 
    
    \item \label{Iic} $\frac{\partial \I_{i}^{C}}{\partial \q_{j}}=
    \begin{cases}
      \Phibar_{j} \timesfM \I_{i}^{C}- \I_{i}^{C}(\Phibar_{j} \timesM), &~~~~~~~~~~~~~~~~~~~~~~~~~ \text{if}\ j \preceq i \\
       \Phibar_{j} \timesfM \I_{j}^{C}- \I_{j}^{C}(\Phibar_{j} \timesM), &~~~~~~~~~~~~~~~~~~~~~~~~~ \text{if}\ j \succ i 
      \\ 0 &~~~~~~~~~~~~~~~~~~~~~~~~~ \text{otherwise}
    \end{cases}$
    
  \item \label{ai} $ \frac{\partial \a_{i}}{\partial \q_{j}}=
    \begin{cases}
      \Psibarddot_{j} - \v_{i} \times \Psibardot_{j} - \a_{i} \times \Phibar_{j} ,   &~~~~~~~~~~~~~~~~~~~~~ \text{if}\ j \preceq i \\
      0,  &~~~~~~~~~~~~~~~~~~~~~ \text{otherwise}
    \end{cases}$
    
     \item \label{Iiai} $ \frac{\partial (\I_{i}\a_{i})}{\partial \q_{j}}=
    \begin{cases}
      (\I_{i} \a_{i})\crff \Phibar_{j}+ \I_{i} \Psibarddot_{j} - \I_{i}( \v_{i} \times \Psibardot_{j}) ,   &~~~~~~ \text{if}\ j \preceq i \\
      0,  &~~~~~~ \text{otherwise}
    \end{cases}$
    
       \item \label{Psiddoti} $\frac{\partial \Psibarddot_{i}}{\partial \q_{j}}=
    \begin{cases}
          \Psibarddot_{j}   \timesM  \Phibar_{i} + 2\Psibardot_{j} \timesM \Psibardot_{i}+\Phibar_{j} \timesM \Psibarddot_{i}, &~~~~~~~~~~~~~~~ \text{if}\ j \preceq i \\
      0, &~~~~~~~~~~~~~~~ \text{otherwise}
    \end{cases}$

     \item \label{Bic} $\frac{\partial \B_{i}^{C}}{\partial \q_{j}}=
    \begin{cases}
      \Bten{\I_{i}^{C}}{\Psibardot_{j}}+\Phibar_{j} \timesfM \B_{i}^{C}- \B_{i}^{C}(\Phibar_{j} \timesM), &~~~~~~~~~ \text{if}\ j \preceq i \\
          \Bten{\I_{j}^{C}}{\Psibardot_{j}}+\Phibar_{j} \timesfM \B_{j}^{C}- \B_{j}^{C}(\Phibar_{j} \timesM), &~~~~~~~~~ \text{if}\ j \succ i 
      \\ 0 &~~~~~~~~~ \text{otherwise}
    \end{cases}$

   \item \label{fi} $ \frac{\partial \f_{i}}{\partial \q_{j}}=
    \begin{cases}
      \I_{i} \Psibarddot_{j} + \f_{i}\crff  \Phibar_{j}  + 2 \B_{i} \Psibardot_{j} ,   &~~~~~~~~~~~~~~~~~~~ \text{if}\ j \preceq i \\
      0,  &~~~~~~~~~~~~~~~~~~~ \text{otherwise}
    \end{cases}$  
    
   \item \label{fic}$\frac{\partial \f_{i}^{C}}{\partial \q_{j}}=
    \begin{cases}
       \I_{i}^{C} \Psibarddot_{j} +  \f_{i}^{C} \crff \Phibar_{j} + 2 \B_{i}^{C} \Psibardot_{j} ,   &~~~~~~~~~~~~~~~~~~~ \text{if}\ j \preceq i \\
       \I_{j}^{C} \Psibarddot_{j} +   \f_{j}^{C}\crff\Phibar_{j} + 2 \B_{j}^{C} \Psibardot_{j}, &~~~~~~~~~~~~~~~~~~~ \text{otherwise}
    \end{cases}$
    
     \item \label{PhiT} $\frac{\partial \Phibar_{i} \T}{\partial \q_{j}}=
    \begin{cases}
         -\Phibar_{i} ^{T}\Phibar_{j} \timesfM, &~~~~~~~~~~~~~~~~~~~~~~~~~~~~~~~~~~~~~~~ \text{if}\ j \preceq i \\
      0, &~~~~~~~~~~~~~~~~~~~~~~~~~~~~~~~~~~~~~~ \text{otherwise}
    \end{cases}$
    
      \item \label{Phidot_qdj}$\frac{\partial \Phibardot_{i}}{\partial \qd_{j}}=
    \begin{cases}
         \Phibar_{j} \timesM \Phibar_{i}, &~~~~~~~~~~~~~~~~~~~~~~~~~~~~~~~~~~~~~~~~~~~ \text{if}\ j \preceq i \\
      0, &~~~~~~~~~~~~~~~~~~~~~~~~~~~~~~~~~~~~~~~~~~~ \text{otherwise}
    \end{cases}$
    
      \item \label{Psidot_qdj} $\frac{\partial \Psibardot_{i}}{\partial \qd_{j}}=
    \begin{cases}
         \Phibar_{j} \timesM \Phibar_{i}, &~~~~~~~~~~~~~~~~~~~~~~~~~~~~~~~~~~~~~~~~~~{\text{if}\ j \prec i} \\
      0,&~~~~~~~~~~~~~~~~~~~~~~~~~~~~~~~~~~~~~~~~~~ \text{otherwise}
    \end{cases}$
    
     \item \label{Bic_qdj} $\frac{\partial \B_{i}^{C}}{\partial \qd_{j}}=
    \begin{cases}
     \Bten{\I_{i}^{C}}{\Phibar_{j}}, &~~~~~~~~~~~~~~~~~~~~~~~~~~~~~~~~~~~~~~~~~ \text{if}\ j \preceq i \\
   \Bten{\I_{j}^{C}}{\Phibar_{j}}, &~~~~~~~~~~~~~~~~~~~~~~~~~~~~~~~~~~~~~~~~~ \text{if}\ j \succ i 
      \\ 0 &~~~~~~~~~~~~~~~~~~~~~~~~~~~~~~~~~~~~~~~~~ \text{otherwise}
    \end{cases}$    
    
\end{enumerate}
\vspace{5ex}

\noindent The identities above are derived in detail as follows.
\begin{enumerate}
    \item[\ref{dphii}]  From identity J1, the directional derivative of joint motion subspace matrix $\Phibar_{i}$ along the $p^{th}$ free-mode of joint $j$ is:
    
    \begin{equation}
         \frac{\partial \Phibar_{i}}{\partial q_{j,p}}=\phibar_{j,p} \times \Phibar_{i}
    \end{equation}

    Collectively for each dimension $p$ for joint $j$:
    
    \begin{equation}
     \frac{\partial \Phibar_{i}}{\partial \q_{j}}=\Phibar_{j} \timesM \Phibar_{i}    
    \end{equation}

    \item[\ref{dphii_dot}] For the partial derivative of $\Phibardot_{i}$ w.r.t $\q_{j}$, we use the definition of $\Phibardot_{i}$ and the product rule for $j \preceq i$ as:
    
    \begin{equation}
        \frac{\partial \Phibardot_{i}}{\partial \q_{j}} = \frac{\partial \v_{i}}{\partial \q_{j}} \timesM \Phibar_{i} + \v_{i} \times \frac{\partial \Phibar_{i}}{\partial \q_{j}}
    \end{equation}
    
The directional partial derivative of the spatial velocity of a body $i$, $\v_{i}$ with respect to $q_{j,p}$, where $j \preceq i$ is given as:

\begin{equation}
    \frac{\partial \v_{i}}{\partial q_{j,p}} = \sum_{l \succeq j} \frac{\partial \Phibar_{l}}{\partial q_{j,p}}  \qd_{l}
\end{equation}

Using J1:
\begin{equation}
    \frac{\partial \v_{i}}{\partial q_{j,p}} = \sum_{l \succeq j} \phibar_{j,p} \times \Phibar_{l} \qd_{l}
\label{dv_eq2}
\end{equation}

Using P1:
\begin{equation}
    \frac{\partial \v_{i}}{\partial q_{j,p}} = - \sum_{l \succeq j} \Phibar_{l} \qd_{l}  \times \phibar_{j,p}
\label{dv_eq3}
\end{equation}

Eq~\ref{dv_eq3} is written for all DoFs collectively as:

\begin{equation}
    \frac{\partial \v_{i}}{\partial \q_{j}} = - \sum_{l \succeq j} \Phibar_{l} \qd_{l}  \times \Phibar_{j}
\label{dv_eq4}
\end{equation}

Using the definition of $\v_{l}$: 

\begin{equation}
    \frac{\partial \v_{i}}{\partial \q_{j}} = (\v_{\lambda(j)} - \v_{i})  \times \Phibar_{j}
\label{J10_eqn}
\end{equation}

Using Eq.~\ref{J10_eqn}, and \ref{dphii}:

     \begin{equation}
        \frac{\partial \Phibardot_{i}}{\partial \q_{j}} = ((\v_{\lambda(j)} - \v_{i})  \times \Phibar_{j}) \timesM \Phibar_{i} + \v_{i} \times (\Phibar_{j} \timesM \Phibar_{i}    )
    \end{equation}

Expanding terms, and using the definition of $\Psibardot_{j}$ (Eq.~\ref{psidot_psidotdot_defn}), 

 \begin{equation}
    \frac{\partial \Phibardot_{i}}{\partial \q_{j}} = (\Psibardot_{j} - \v_{i}   \times \Phibar_{j}) \timesM \Phibar_{i} + \v_{i} \times (\Phibar_{j} \timesM \Phibar_{i}    )
\end{equation}

Using \ref{m9}, and property P10:

\begin{equation}
     \frac{\partial \Phibardot_{i}}{\partial \q_{j}} = \Psibardot_{j} \timesM \Phibar_{i}  - \v_{i}   \times \Phibar_{j} \timesM \Phibar_{i} + \Phibar_{j} \timesM \v_{i} \times \Phibar_{i} +  \v_{i} \times \Phibar_{j} \timesM \Phibar_{i}  
\end{equation}

Cancelling terms and using the definition of $\Phibardot_{i}$:

\begin{equation}
     \frac{\partial \Phibardot_{i}}{\partial \q_{j}} = \Psibardot_{j} \timesM \Phibar_{i}  + \Phibar_{j} \timesM  \Phibardot_{i}   
\end{equation}

\item[\ref{vji_phii}] Using the product rule:

\begin{equation}
    \frac{\partial (\vJ{i} \times \Phibar_{i})}{\partial \q_{j}} = \left( \frac{\partial \vJ{i}}{\partial \q_{j}} \right) \timesM \Phibar_{i} + \vJ{i} \times \frac{\partial \Phibar_{i}}{\partial \q_{j}}
    \label{vji_phii_eq1}
\end{equation}

The partial derivative of the joint velocity $\vJ{i}$ with respect to $\q_{j}$ for $j \preceq i$ is now calculated. The directional derivative of $\vJ{i}$ along the $p^{th}$ free-mode of joint $j$ is:

\begin{equation}
    \frac{\partial \vJ{i}}{\partial q_{j,p}} = \frac{\partial (\Phibar_{i} \qd_{i})}{\partial q_{j,p}}  
    \label{dvj_eq1}
\end{equation}

Using J1:

\begin{equation}
    \frac{\partial \vJ{i}}{\partial q_{j,p}} = \phibar_{j,p} \times \Phibar_{i} \qd_{i}
    \label{dvj_eq2}
\end{equation}

Using property P1 (Sec. \ref{prop_vec}):

\begin{equation}
    \frac{\partial \vJ{i}}{\partial q_{j,p}} = - \vJ{i} \times \phibar_{j,p}
    \label{dvj_eq3}
\end{equation}

Collectively, for all DoFs of joint $j$:

\begin{equation}
    \frac{\partial \vJ{i}}{\partial \q_{j}} = - \vJ{i} \times \Phibar_{j}
    \label{dvj_eq4}
\end{equation}

Using Eq.~\ref{dvj_eq4} and \ref{dphii} in Eq.~\ref{vji_phii_eq1}:

\begin{equation}
    \frac{\partial (\vJ{i} \times \Phibar_{i})}{\partial \q_{j}} = -( \vJ{i} \times \Phibar_{j}) \timesM \Phibar_{i} + \vJ{i} \times  \Phibar_{j}   \timesM \Phibar_{i}
\end{equation}

Using property \ref{m9}:

\begin{equation}
    \frac{\partial (\vJ{i} \times \Phibar_{i})}{\partial \q_{j}} = - \vJ{i} \times \Phibar_{j} \timesM \Phibar_{i} + \Phibar_{j} \timesM \vJ{i} \times \Phibar_{i} + \vJ{i} \times  \Phibar_{j}   \timesM \Phibar_{i}
\end{equation}

Cancelling terms:

\begin{equation}
    \frac{\partial (\vJ{i} \times \Phibar_{i})}{\partial \q_{j}} =  \Phibar_{j} \timesM ( \vJ{i} \times \Phibar_{i} )
\end{equation}

\item [\ref{psii_dot}] Using the definition of $\Psibardot_{i}$ (Eq.~\ref{psidot_psidotdot_defn}):

\begin{equation}
    \frac{\partial \Psibardot_{i}}{\partial \q_{j}} = \frac{\partial (\Phibardot_{i}-\vJ{i} \times \Phibar_{i})}{\partial \q_{j}}
\end{equation}

Using identities \ref{dphii_dot} and \ref{vji_phii} for $j \preceq i$:

\begin{equation}
    \frac{\partial \Psibardot_{i}}{\partial \q_{j}} =   \Psibardot_{j}   \timesM \Phibar_{i}+ \Phibar_{j} \timesM \Phibardot_{i} -  \Phibar_{j}   \timesM (\vJ{i} \times \Phibar_{i})
\end{equation}

Combining the second and third terms on the RHS, and using the definition of $\Psibardot_{i}$ (Eq.~\ref{psidot_psidotdot_defn}):

\begin{equation}
    \frac{\partial \Psibardot_{i}}{\partial \q_{j}} =   \Psibardot_{j}   \timesM \Phibar_{i}+ \Phibar_{j} \timesM \Psibardot_{i} 
\end{equation}

\item [\ref{Ii}]

The directional partial derivative of $\I_{i}$ w.r.t to the $p^{th}$ free-mode of a joint $j$ is given by~\cite{Featherstone08}:

\begin{equation}
    \frac{\partial \I_{i}}{\partial q_{j,p}}=  \phibar_{j,p} \times^* \I_{i}- \I_{i}(\phibar_{j,p} \times)
\end{equation}

Collectively, for all free modes $p$ of the joint $j$, the partial derivative of $\I_{i}$ wrt $\q_{j}$ can be written as:

\begin{equation}
    \frac{\partial \I_{i}}{\partial \q_{j}} = \Phibar_{j} \timesfM \I_{i}- \I_{i}(\Phibar_{j} \timesM)
\end{equation}

\item[\ref{Iic}] Consider first the case when $i \succeq j$. Using \ref{Ii}, we have:
\begin{equation}
     \frac{\partial \I_{i}}{\partial \q_{j}}=   \Phibar_{j} \timesfM \I_{i}- \I_{i}(\Phibar_{j} \timesM)
     \label{diff_Ii_qj}
\end{equation}

Taking the partial derivative of $\I_{i}^{C}$:
\begin{equation}
    \frac{\partial \I_{i}^{C}}{\partial \q_{j}}= \frac{\partial \sum_{k \succeq i} \I_{k} }{\partial \q_{j}}
\end{equation}

\begin{equation}
    \frac{\partial \I_{i}^{C}}{\partial \q_{j}}= \sum_{k \succeq i} \frac{\partial  \I_{k} }{\partial \q_{j}} 
\end{equation}

Using Eq.~\ref{diff_Ii_qj}:
\begin{equation}
    \frac{\partial \I_{i}^{C}}{\partial \q_{j}}= \sum_{k \succeq i} \big( \Phibar_{j} \timesfM \I_{k}- \I_{k}(\Phibar_{j} \timesM) \big)
\end{equation}

Summing over $k$:
\begin{equation}
    \frac{\partial \I_{i}^{C}}{\partial \q_{j}}=  \Phibar_{j} \timesfM \I_{i}^{C}- \I_{i}^{C}(\Phibar_{j} \timesM) 
\end{equation}

For the case $j \succ i$:
\begin{equation}
    \frac{\partial \I_{i}^{C}}{\partial \q_{j}}= \sum_{k \succeq j} \big( \Phibar_{j} \timesfM \I_{k}- \I_{k}(\Phibar_{j} \timesM) \big)
\end{equation}

Summing over $k$:
\begin{equation}
    \frac{\partial \I_{i}^{C}}{\partial \q_{j}}=  \Phibar_{j} \timesfM \I_{j}^{C}- \I_{j}^{C}(\Phibar_{j} \timesM) 
\end{equation}

\item [\ref{ai}] Using the definition of $\a_{i} = \gammabar_{i} + \xibar_{i} + \ag{i}$ (See Ref.~\cite{singh2022efficient}), and taking the partial derivative wrt $\q_{j}$:

\begin{equation}
    \frac{\partial \a_{i}}{\partial \q_{j}} =  \frac{\partial (\gammabar_{i} + \xibar_{i} + \ag{i})}{\partial \q_{j}}
\end{equation}

The quantity $\ag{i}$ is constant, it results into a zero partial derivative. Hence, 

\begin{equation}
    \frac{\partial \a_{i}}{\partial \q_{j}} =  \frac{\partial (\gammabar_{i}  )}{\partial \q_{j}} +\frac{\partial ( \xibar_{i} )}{\partial \q_{j}}
\end{equation}

Using J6, and J7, for $j \preceq i$:

\begin{equation}
    \frac{\partial \a_{i}}{\partial \q_{j}} =  (\v_{\lambda(j)} - \v_{i} )  \times  \Psibardot_{j} + ( \xibar_{\lambda(j)} - \xibar_{i}) \times  \Phibar_{j} +(\gammabar_{\lambda(j)} - \gammabar_{i})  \times \Phibar_{j} 
\end{equation}

Combining second and the third terms:

\begin{equation}
    \frac{\partial \a_{i}}{\partial \q_{j}} =  (\v_{\lambda(j)} - \v_{i} )  \times  \Psibardot_{j} + ( \xibar_{\lambda(j)} +\gammabar_{\lambda(j)} - (\xibar_{i}+\gammabar_{i})) \times  \Phibar_{j} 
\end{equation}

Since $\ag{i} =\ag{\lambda(j)} = \ag{}$, adding $\ag{\lambda(j)}$ and subtracting $\ag{i}$:

\begin{equation}
    \frac{\partial \a_{i}}{\partial \q_{j}} =  (\v_{\lambda(j)} - \v_{i} )  \times  \Psibardot_{j} + ( \xibar_{\lambda(j)} +\gammabar_{\lambda(j)}+\ag{\lambda(j)} - (\xibar_{i}+\gammabar_{i}+\ag{i})) \times  \Phibar_{j} 
\end{equation}

Using the definition of $\a_{\lambda(j)}$ and $\a_{i}$:

\begin{equation}
    \frac{\partial \a_{i}}{\partial \q_{j}} =  (\v_{\lambda(j)} - \v_{i} )  \times  \Psibardot_{j} + ( \a_{\lambda(j)} - \a_{i}) \times  \Phibar_{j} 
\end{equation}

Using the definition of $\Psibarddot_{j}$ (Eq.~\ref{psidot_psidotdot_defn}), and simplifying:

\begin{equation}
    \frac{\partial \a_{i}}{\partial \q_{j}} =  \Psibarddot_{j} - \v_{i} \times \Psibardot_{j} - \a_{i} \times \Phibar_{j}
\end{equation}

\item [\ref{Iiai}]

Using the product rule of differentiation to take the partial derivative of $\I_{i} \a_{i}$ wrt $\q_{j}$ as:

\begin{equation}
    \frac{\partial (\I_{i}\a_{i})}{\partial \q_{j}} =  \Bigg( \frac{\partial \I_{i}}{\partial \q_{j}} \Bigg) \a_{i} +  \I_{i}\Bigg( \frac{\partial \a_{i}}{\partial \q_{j}} \Bigg) 
\end{equation}

Using \ref{Ii} and \ref{ai} as:

\begin{equation}
    \frac{\partial (\I_{i}\a_{i})}{\partial \q_{j}} =  \big(  \Phibar_{j} \timesfM \I_{i}- \I_{i}(\Phibar_{j} \timesM) \big) \a_{i} +  \I_{i}\big( \Psibarddot_{j} - \v_{i} \times \Psibardot_{j} - \a_{i} \times \Phibar_{j} \big) 
\end{equation}

Expanding:

\begin{equation}
    \frac{\partial (\I_{i}\a_{i})}{\partial \q_{j}} =   \Phibar_{j} \timesfM \I_{i} \a_{i}- \I_{i}(\Phibar_{j} \timesM) \a_{i} +  \I_{i} \Psibarddot_{j} -  \I_{i}\v_{i} \times \Psibardot_{j} -  \I_{i}\a_{i} \times \Phibar_{j}
\end{equation}

Using the property \ref{m4}  in the second term on RHS:

\begin{equation}
    \frac{\partial (\I_{i}\a_{i})}{\partial \q_{j}} =   \Phibar_{j} \timesfM \I_{i} \a_{i}+ \I_{i}(\a_{i} \times \Phibar_{j})\Rten +  \I_{i} \Psibarddot_{j} -  \I_{i}\v_{i} \times \Psibardot_{j} -  \I_{i}\a_{i} \times \Phibar_{j}
\end{equation}
The terms $(\a_{i} \times \Phibar_{j})\Rten$ and $(\a_{i} \times \Phibar_{j})$ are matrices but in different dimensions. Hence these can be cancelled:

\begin{equation}
    \frac{\partial (\I_{i}\a_{i})}{\partial \q_{j}} =   \Phibar_{j} \timesfM \big(\I_{i} \a_{i} \big)+  \I_{i} \Psibarddot_{j} -  \I_{i}\big( \v_{i} \times \Psibardot_{j} \big) 
\end{equation}

\item [\ref{Psiddoti}] Taking the partial derivative of $\Psibarddot_{i}$ (Eq.~\ref{psidot_psidotdot_defn}) and using the product rule results in:

\begin{equation}
  \frac{\partial \Psibarddot_{i}}{\partial \q_{j}} =   \Bigg( \frac{\partial \alam{i}}{\partial \q_{j}} \Bigg) \timesM \Phibar_{i} + \alam{i} \times  \Bigg( \frac{\partial \Phibar_{i}}{\partial \q_{j}} \Bigg) + \Bigg( \frac{\partial \vlam{i}}{\partial \q_{j}} \Bigg) \timesM \Psibardot_{i} + \vlam{i} \times  \Bigg( \frac{\partial \Psibardot_{i}}{\partial \q_{j}} \Bigg) 
\end{equation}

Using \ref{ai} for the first term, \ref{dphii} for the second term, Eq.~\ref{J10_eqn} for the third term, and \ref{psii_dot} for the last term results in:

\begin{equation}
\begin{aligned}
  \frac{\partial \Psibarddot_{i}}{\partial \q_{j}} =   \big( \Psibarddot_{j} - \vlam{i} \times \Psibardot_{j} - \alam{i} \times \Phibar_{j} \big) \timesM \Phibar_{i} + \alam{i} \times  \big( \Phibar_{j}   \timesM \Phibar_{i} \big) + \\
  \big(  (\vlam{j} - \vlam{i} )  \times \Phibar_{j}\big) \timesM \Psibardot_{i} + \vlam{i} \times  \big( \Psibardot_{j}   \timesM  \Phibar_{i} + \Phibar_{j} \timesM \Psibardot_{i} \big) 
  \end{aligned}
\end{equation}

Expanding terms using the property \ref{m9} repeatedly, and using the definition of $\Psibardot_{j}$ (Eq.~\ref{psidot_psidotdot_defn}):

\begin{equation}
\begin{aligned}
  \frac{\partial \Psibarddot_{i}}{\partial \q_{j}} &=    \Psibarddot_{j}  \timesM \Phibar_{i} - \vlam{i} \times \Psibardot_{j} \timesM  \Phibar_{i}+ \Psibardot_{j} \timesM \vlam{i} \times \Phibar_{i}-  \alam{i} \times \Phibar_{j}  \timesM \Phibar_{i} +  \Phibar_{j}  \timesM \alam{i} \times \Phibar_{i} \\&~~
  + \alam{i} \times  \Phibar_{j}   \timesM \Phibar_{i}  + 
 \Psibardot_{j} \timesM \Psibardot_{i} - \vlam{i}  \times \Phibar_{j} \timesM \Psibardot_{i} + \Phibar_{j} \timesM  \vlam{i}  \times \Psibardot_{i}  \\&~~~~
 + \vlam{i} \times   \Psibardot_{j}   \timesM  \Phibar_{i} +  \vlam{i} \times \Phibar_{j} \timesM \Psibardot_{i}
  \end{aligned}
\end{equation}

Cancelling terms:

\begin{equation}
\begin{aligned}
  \frac{\partial \Psibarddot_{i}}{\partial \q_{j}} &=    \Psibarddot_{j}  \timesM \Phibar_{i}  + \Psibardot_{j} \timesM \vlam{i} \times \Phibar_{i} +  \Phibar_{j}  \timesM \alam{i} \times \Phibar_{i} \\&~~
 +  \Psibardot_{j} \timesM \Psibardot_{i}  + \Phibar_{j} \timesM  \vlam{i}  \times \Psibardot_{i}
  \end{aligned}
\end{equation}

Using the definition of $\Psibardot_{i}$ (Eq.~\ref{psidot_psidotdot_defn}) and collecting terms:

\begin{equation}
\begin{aligned}
  \frac{\partial \Psibarddot_{i}}{\partial \q_{j}} &=    \Psibarddot_{j}  \timesM \Phibar_{i}  +2 \Psibardot_{j} \timesM \Psibardot_{i} +  \Phibar_{j}  \timesM \big( \alam{i} \times \Phibar_{i} +  \vlam{i}  \times \Psibardot_{i} \big)
  \end{aligned}
\end{equation}

Using the definition of $\Psibarddot_{i}$ (Eq.~\ref{psidot_psidotdot_defn}) results in:

\begin{equation}
\begin{aligned}
  \frac{\partial \Psibarddot_{i}}{\partial \q_{j}} &=    \Psibarddot_{j}  \timesM \Phibar_{i}  +2 \Psibardot_{j} \timesM \Psibardot_{i} +  \Phibar_{j}  \timesM \Psibarddot_{i}
  \end{aligned}
\end{equation}

\item [\ref{Bic}] We consider first the case when $i\succeq j$. Using the definition of $\B_{i}$ (\eqref{bkm_term_defn}), we know:
where $\B{}_i^C = \sum_{k\succeq i} \B_{k}$.

For the case, $j \preceq i$, taking partial derivative wrt $\q_{j}$

\begin{equation}
   \frac{\partial \B{}_i^C}{\partial \q_{j}}  = \frac{1}{2} \sum_{k\succeq i} \left(  \frac{\partial \big(\v_{k}\times^* \big)\I_{k}}{\partial \q_{j}}\ - \frac{\partial \I_{k} \big(\v_{k} \times   \big)}{\partial \q_{j}}  + \frac{\partial \big(\I_{k} \v_{k} \big) \crff}{\partial \q_{j}}\right)
\end{equation}

Using the product rule of differentiation:

\begin{equation}
\begin{aligned}
   \frac{\partial \B{}_i^C}{\partial \q_{j}} = \frac{1}{2} \sum_{k\succeq i}  \left(\frac{\partial \v_{k} } {\partial \q_{j}} \right)\timesfM \I_{k} + \v_{k} \times^* \frac{\partial \I_{k} } {\partial \q_{j}}- \left(\frac{\partial \I_{k} } {\partial \q_{j}} \right)\v_{k} \times-\I_{k} \left(\frac{\partial \v_{k} } {\partial \q_{j}} \right) \timesM  + \left( \left(\frac{\partial \I_{k} } {\partial \q_{j}} \v_{k} + \I_{k} \frac{\partial \v_{k} } {\partial \q_{j}} \right) \crffM \right)
   \end{aligned}
\end{equation}

Using Eq.~\ref{J10_eqn} and \ref{Ii}:

\begin{equation}
    \begin{aligned}
       \frac{\partial \B{}_i^C}{\partial \q_{j}}  &= \frac{1}{2} \sum_{k\succeq i} \Big(  \big(  (\vlam{j} - \v_{k} )  \times \Phibar_{j} \big)\timesfM \I_{k} + \v_{k} \times^* ( \Phibar_{j} \timesfM \I_{k}- \I_{k}(\Phibar_{j} \timesM))- \\&~~~
       (  \Phibar_{j} \timesfM \I_{k}- \I_{k}(\Phibar_{j} \timesM) )\v_{k} \times-\I_{k} (  (\vlam{j} - \v_{k} )  \times \Phibar_{j} ) \timesM  + \\ &~~~ ( (\Phibar_{j} \timesfM \I_{k}- \I_{k}(\Phibar_{j} \timesM) ) \v_{k}  + \I_{k} (  (\vlam{j} - \v_{k} )  \times \Phibar_{j} ) ) \crffM \Big)
       \end{aligned}
\end{equation}

Using the definition of $\Psibardot_{j}$ (Eq.~\ref{psidot_psidotdot_defn}), and expanding terms:
\begin{equation}
    \begin{aligned}
       \frac{\partial \B{}_i^C}{\partial \q_{j}}  &= \frac{1}{2} \sum_{k\succeq i} \big(  (  \Psibardot_{j} - \v_{k}   \times \Phibar_{j} )\timesfM \I_{k} + \v_{k} \times^*  \Phibar_{j} \timesfM \I_{k}- \v_{k} \times^* \I_{k}(\Phibar_{j} \timesM)- \\&~~~
         \Phibar_{j} \timesfM \I_{k} \v_{k} \times+ \I_{k}(\Phibar_{j} \timesM) \v_{k} \times -\I_{k} \big(   \Psibardot_{j} - \v_{k}   \times \Phibar_{j} \big) \timesM  + \\ &~~~
         \big( \Phibar_{j} \timesfM \I_{k} \v_{k} - \I_{k}(\Phibar_{j} \timesM)  \v_{k} + \I_{k}  \Psibardot_{j} - \I_{k} \v_{k}   \times \Phibar_{j}\big)  \crffM \big)
    \end{aligned}
\end{equation}
using properties \ref{m9} and \ref{m10} to expand terms:
\begin{equation}
    \begin{aligned}
       \frac{\partial \B{}_i^C}{\partial \q_{j}}  &= \frac{1}{2} \sum_{k\succeq i} \big(   \Psibardot_{j}\timesfM \I_{k} - \v_{k}   \times^* \Phibar_{j} \timesfM \I_{k} + \Phibar_{j} \timesfM \v_{k}   \times^* \I_{k} + \v_{k} \times^*  \Phibar_{j} \timesfM \I_{k}- \v_{k} \times^* \I_{k}(\Phibar_{j} \timesM)- \\&~~~
         \Phibar_{j} \timesfM \I_{k} \v_{k} \times+ \I_{k}(\Phibar_{j} \timesM) \v_{k} \times -\I_{k} \Psibardot_{j}\timesM  + \I_{k}\v_{k} \times \Phibar_{j}  \timesM - \I_{k}\Phibar_{j}  \timesM \v_{k} \times + \\ &~~~
         \big( \Phibar_{j} \timesfM \I_{k} \v_{k} - \I_{k}(\Phibar_{j} \timesM)  \v_{k} + \I_{k}  \Psibardot_{j} - \I_{k} \v_{k}   \times \Phibar_{j}\big)  \crffM \big)
    \end{aligned}
\end{equation}

Cancelling terms, and using property \ref{m4} results in simplifications.

\begin{equation}
    \begin{aligned}
       \frac{\partial \B{}_i^C}{\partial \q_{j}}  &= \frac{1}{2} \sum_{k \succeq i} \Big(    \Psibardot_{j}\timesfM \I_{k}  + \Phibar_{j} \timesfM \v_{k}   \times^* \I_{k} - \v_{k} \times^* \I_{k}(\Phibar_{j} \timesM)- \\&~~~
         \Phibar_{j} \timesfM \I_{k} \v_{k} \times -\I_{k} \Psibardot_{j}\timesM  + \I_{k}\v_{k} \times \Phibar_{j}  \timesM  + \\ &~~~
         \Big( \Phibar_{j} \timesfM \I_{k} \v_{k} + \I_{k}(\v_{k} \timesM\Phibar_{j})\Rten  + \I_{k}  \Psibardot_{j} - \I_{k} \v_{k}   \times \Phibar_{j}\Big)  \crffM \Big)
    \end{aligned}
\end{equation}

The terms $(\v_{k} \timesM\Phibar_{j})\Rten$ and $(\v_{k} \timesM\Phibar_{j})$  are both matrices but in different dimensions. Hence these two can be cancelled. The term $\Phibar_{j} \timesfM \I_{k} \v_{k} $ is replaced with $(\Phibar_{j} \timesfM \I_{k} \v_{k} )\Rten$ since the resulting quantity is a matrix. Now using $\ref{m11}$ on it:
\begin{equation}
    \begin{aligned}
       \frac{\partial \B{}_i^C}{\partial \q_{j}}  &= \frac{1}{2} \sum_{k \succeq i} \Bigg[    \Psibardot_{j}\timesfM \I_{k} + \Phibar_{j} \timesfM \v_{k}   \times^* \I_{k} -  \v_{k} \times^* \I_{k}(\Phibar_{j} \timesM)- \\&~~~
         \Phibar_{j} \timesfM \I_{k} \v_{k} \times -\I_{k} \Psibardot_{j}\timesM  + \I_{k}\v_{k} \times \Phibar_{j}  \timesM + \\ &~~~
          \Phibar_{j} \timesfM \big(\I_{k} \v_{k}\big)\crffM -  \big(\I_{k} \v_{k}\big)\crffM  \Phibar_{j} \timesM + \big(\I_{k}  \Psibardot_{j} \big)  \crffM \Bigg]
    \end{aligned}
\end{equation}
Collecting the $\Psibardot_{j}$ terms:
\begin{equation}
    \begin{aligned}
       \frac{\partial \B{}_i^C}{\partial \q_{j}}  &= \frac{1}{2} \sum_{k\succeq i} \big(   \Psibardot_{j}\timesfM \I_{k} -\I_{k} \Psibardot_{j}\timesM+ \big(\I_{k}  \Psibardot_{j} \big)  \crffM \\&~~~
       + \Phibar_{j} \timesfM \big(\v_{k}   \times^* \I_{k} -   \I_{k} \v_{k} \times +\big(\I_{k} \v_{k}\big)\crffM \big) - \big(  \v_{k} \times^* \I_{k} - \I_{k}\v_{k} \times +\big(\I_{k} \v_{k}\big)\crffM  \big) \Phibar_{j}  \timesM \big)
    \end{aligned}
\end{equation}

Using the definition of $\B_{i}^C$:
\begin{equation}
       \frac{\partial \B{}_i^C}{\partial \q_{j}}  = \Bten{\I_{i}^{C}}{\Psibardot_{j}} + \Phibar_{j} \timesfM \B_{i}^C - \B_{i}^C\Phibar_{j}  \timesM 
\end{equation}

where $\Bten{\I_{i}^{C}}{\Psibardot_{j}} = \sum_{k\succeq i} \Bten{\I_{k}}{\Psibardot_j}$, with $\Bten{\I_{k}}{\Psibardot_j}$:

\begin{equation}
    \Bten{\I_{k}}{\Psibardot_j} = \frac{1}{2} \big(\big(\Psibardot_{j}\timesfM \big)\I_{k} - \I_{k} \big(\Psibardot_{j} \timesM   \big) + \big(\I_{k} \Psibardot_{j} \big) \crffM    \big)
    \label{bl_psijdot_term_defn}
\end{equation}
For the case, $j \succ i$, taking partial derivative wrt $\q_{j}$

\begin{equation}
   \frac{\partial \B{}_i^C}{\partial \q_{j}}  = \frac{1}{2} \sum_{k\succeq j} \left(  \frac{\partial \big(\v_{k}\times^* \big)\I_{k}}{\partial \q_{j}}\ - \frac{\partial \I_{k} \big(\v_{k} \times   \big)}{\partial \q_{j}}  + \frac{\partial \big(\I_{k} \v_{k} \big) \crff}{\partial \q_{j}}  \right)
\end{equation}

Similar steps for this case results in:

\begin{equation}
       \frac{\partial \B{}_i^C}{\partial \q_{j}}  = \Bten{\I_{j}^{C}}{\Psibardot_{j}}  + \Phibar_{j} \timesfM \B_{j}^C - \B_{j}^C\Phibar_{j}  \timesM 
\end{equation}

\item [\ref{fi}]

Using the definition of $\f_{i}$ (Eq.~\ref{sp_eqn_of_motion}) to take the partial derivative as:

\begin{equation}
    \frac{\partial \f_{i}}{ \partial \q_{j}} =  \Bigg( \frac{\partial \I_{i} \a_{i}} {\partial \q_{j}} \Bigg)   + \Bigg( \frac{\partial \v_{i} } {\partial \q_{j}}  \Bigg) \timesfM \I_{i} \v_{i} + \v_{i} \times^*  \Bigg( \frac{\partial \I_{i} \v_{i}  } {\partial \q_{j}}    \Bigg)
\end{equation}

Using J5, Eq.~\ref{J10_eqn} and \ref{Iiai}:

\begin{equation}
    \begin{aligned}
        \frac{\partial \f_{i}}{ \partial \q_{j}} &=  \big(\I_{i} \a_{i} \big)\crff \Phibar_{j}+    \I_{i} \Psibarddot_{j} -  \I_{i}\big( \v_{i} \times \Psibardot_{j} \big)   + \big(     (\vlam{j} - \v_{i} )  \times \Phibar_{j} \big) \timesfM\I_{i} \v_{i} + \\&~~~~
        \v_{i} \times^*  \left( (\I_{i} \v_{i}) \crff \Phibar_{j} + \I_{i}(\Psibardot_{j})   \right)
    \end{aligned}
\end{equation}

Using the definition of $\Psibardot_{j}$ (Eq.~\ref{psidot_psidotdot_defn}), and expanding terms:

\begin{equation}
    \begin{aligned}
        \frac{\partial \f_{i}}{ \partial \q_{j}} &= \big(\I_{i} \a_{i} \big)\crff \Phibar_{j}+    \I_{i} \Psibarddot_{j} -  \I_{i}\big( \v_{i} \times \Psibardot_{j} \big)   + \big(     \Psibardot_{j} - \v_{i}  \times \Phibar_{j} \big) \timesfM\I_{i} \v_{i} + \\&~~~
        \v_{i} \times^*  (\I_{i} \v_{i}) \crff \Phibar_{j} +  \v_{i} \times^*\I_{i}(\Psibardot_{j})   
    \end{aligned}
\end{equation}

Using the property \ref{m10} to expand terms:

\begin{equation}
    \begin{aligned}
        \frac{\partial \f_{i}}{ \partial \q_{j}} &= \big(\I_{i} \a_{i} \big)\crff \Phibar_{j}+   \I_{i} \Psibarddot_{j} -  \I_{i}\big( \v_{i} \times \Psibardot_{j} \big)   + \Psibardot_{j} \timesfM\I_{i} \v_{i} - \v_{i}  \times^* \Phibar_{j}  \timesfM  \I_{i} \v_{i} + \Phibar_{j}  \timesfM \v_{i}  \times^* \I_{i} \v_{i} + \\&~~~
        \v_{i} \times^*  (\I_{i} \v_{i}) \crff \Phibar_{j} +  \v_{i} \times^*\I_{i}(\Psibardot_{j})   
    \end{aligned}
\end{equation}

Using property \ref{m15} so that $(\I_{i} \v_{i}) \crff \Phibar_{j} = (\Phibar_{j} \timesfM (\I_{i} \v_{i}))\Rten $ to get:

\begin{equation}
    \begin{aligned}
        \frac{\partial \f_{i}}{ \partial \q_{j}} &=  \big(\I_{i} \a_{i} \big)\crff \Phibar_{j}+  \I_{i} \Psibarddot_{j} -  \I_{i}\big( \v_{i} \times \Psibardot_{j} \big)   + \Psibardot_{j} \timesfM\I_{i} \v_{i} - \v_{i}  \times^* \Phibar_{j}  \timesfM  \I_{i} \v_{i} + \Phibar_{j}  \timesfM \v_{i}  \times^* \I_{i} \v_{i} + \\&~~~
        \v_{i} \times^* (\Phibar_{j} \timesfM (\I_{i} \v_{i}))\Rten  +  \v_{i} \times^*\I_{i}(\Psibardot_{j})   
    \end{aligned}
\end{equation}
The term $ (\Phibar_{j} \timesfM (\I_{i} \v_{i}))$ and $ (\Phibar_{j} \timesfM (\I_{i} \v_{i}))\Rten$ are equivalent since both result into a matrix. Hence these can be cancelled, resulting into:

\begin{equation}
    \begin{aligned}
        \frac{\partial \f_{i}}{ \partial \q_{j}} &= \big(\I_{i} \a_{i} \big)\crff \Phibar_{j}+    \I_{i} \Psibarddot_{j} -  \I_{i}\big( \v_{i} \times \Psibardot_{j} \big)   + \Psibardot_{j} \timesfM\I_{i} \v_{i} + \Phibar_{j}  \timesfM \v_{i}  \times^* \I_{i} \v_{i} +  \v_{i} \times^*\I_{i}\Psibardot_{j})  
    \end{aligned}
\end{equation}

Using \ref{m15} again for the first and fourth term on RHS, followed by flattening the terms:

\begin{equation}
    \begin{aligned}
        \frac{\partial \f_{i}}{ \partial \q_{j}} &=  \Phibar_{j} \timesfM \big(\I_{i} \a_{i} +\v_{i}  \times^* \I_{i} \v_{i}  \big)+  \I_{i} \Psibarddot_{j} -  \I_{i}\big( \v_{i} \times \Psibardot_{j} \big)   +(\I_{i} \v_{i}) \crff \Psibardot_{j} +   \v_{i} \times^*\I_{i}(\Psibardot_{j})   
    \end{aligned}
\end{equation}

Using the definition of $\f_{i}$, and \ref{m5} for the first term:

\begin{equation}
    \begin{aligned}
        \frac{\partial \f_{i}}{ \partial \q_{j}} &=  ( \f_{i} \crff\Phibar_{j})\Rten+  \I_{i} \Psibarddot_{j} +2 \Bigg(-\frac{1}{2} \I_{i}\big( \v_{i} \times  \big)   +\frac{1}{2}(\I_{i} \v_{i}) \crff +   \frac{1}{2}\v_{i} \times^*\I_{i}  \Bigg)\Psibardot_{j} 
    \end{aligned}
\end{equation}

Using the definition of $\Bmat{I_i}{\v_i}$ and flattening the first term:

\begin{equation}
    \begin{aligned}
        \frac{\partial \f_{i}}{ \partial \q_{j}} &=    \I_{i} \Psibarddot_{j}+ \f_{i} \crff\Phibar_{j} +2 \B_{i}\Psibardot_{j}
    \end{aligned}
\end{equation}

\item [\ref{fic}]

This identity follows directly from \ref{fi}. Using the definition of $\f{}_{i}^{C} = \sum_{k\succeq i} \f_{k}$ and taking the partial derivative for the case $j \preceq i$:

\begin{equation}
      \frac{\partial \f_{i}^{C}}{ \partial \q_{j}} = \sum_{k\succeq i} \Bigg(   \frac{\partial \f_{k}}{ \partial \q_{j}}  \Bigg)
\end{equation}

Using \ref{fi}:

\begin{equation}
      \frac{\partial \f_{i}^{C}}{ \partial \q_{j}} = \sum_{k\succeq i} \left( \I_{k} \Psibarddot_{j}+  \f_{k}\crff\Phibar_{j} +2 \B_{k}\Psibardot_{j}  \right)
\end{equation}

Summing over the index $k$:

\begin{equation}
      \frac{\partial \f_{i}^{C}}{ \partial \q_{j}} =  \I_{i}^{C} \Psibarddot_{j}+ \f_{i}^{C} \crff \Phibar_{j}+2 \B_{i}^{C}\Psibardot_{j}  
\end{equation}

For the case $ j \succ i$, we follow similar steps to get:

\begin{equation}
      \frac{\partial \f_{i}^{C}}{ \partial \q_{j}} =  \I_{j}^{C} \Psibarddot_{j}+ \f_{j}^{C}\crff \Phibar_{j} +2 \B_{j}^{C}\Psibardot_{j}  
\end{equation}

\item [\ref{PhiT}]

Using \ref{dphii} to take the partial derivative of $\Phibar_{i}\T $:

\begin{equation}
    \frac{\partial \Phibar_{i}\T }{\partial \q_{j}} = \big(  \Phibar_{j}   \timesM \Phibar_{i} \big)\Tten
\end{equation}

Using the property \ref{m1} to get:

\begin{equation}
    \frac{\partial \Phibar_{i}\T }{\partial \q_{j}} = -\Phibar_{i} \T \Phibar_{j} \timesfM
\end{equation}

\item [\ref{Phidot_qdj}]

Using the definition of $\Phibardot_{i} = \v_{i} \timesM \Phibar_{i}$, and taking the partial derivative w.r.t $\qd_{j}$ as:

\begin{equation}
    \frac{\partial \Phibardot_{i}}{\partial \qd_{j}} = \frac{(\partial \v_{i} \timesM \Phibar_{i})}{\partial \qd_{j}}
\end{equation}

Using identity J8 results for $j \preceq i$ in:

\begin{equation}
     \frac{\partial \Phibardot_{i}}{\partial \qd_{j}} = \Phibar_{j} \timesM \Phibar_{i}
\end{equation}

\item [\ref{Psidot_qdj}]

Using the definition of $\Psibardot_{i} = (\v_{i} - \vJ{i}) \times \Phibar_{i}$:

\begin{equation}
    \frac{\partial \Psibardot_{i}}{\partial \qd_{j}} = \left( \frac{\partial \v_{i}}{\partial \qd_{j}} - \frac{\partial \vJ{i}}{\partial \qd_{j}} \right) \timesM \Phibar_{i}
\end{equation}

Using J8 for the first term for $j \preceq i$, and the definition of $\vJ{i} = \Phibar_{i} \qd_{i}$ for the second term:

\begin{equation}
    \frac{\partial \Psibardot_{i}}{\partial \qd_{j}} = \Bigg( \frac{\partial \v_{i}}{\partial \qd_{j}} - \frac{\partial \vJ{i}}{\partial \qd_{j}} \Bigg) \timesM \Phibar_{i}
\end{equation}

For $j = i$, the expression results in 0 since
\[
 \frac{\partial \v_{j}}{\partial \qd_{j}} = \frac{\partial \vJ{j}}{\partial \qd_{j}} = \Phibar_j
\]
But for the case $ j \prec i$, the expression is:

\begin{equation}
    \frac{\partial \Psibardot_{i}}{\partial \qd_{j}} =  \Phibar_{j} \timesM \Phibar_{i}
\end{equation}

\item [\ref{Bic_qdj}]
Using the definition of $\B_{i}^{C}$ (Eq.~\ref{bl_term_defn}) and taking the partial derivative w.r.t $\qd_{j}$ for the case $j \preceq i$:

\begin{equation}
    \frac{\partial \B_{i}^{C}}{\partial \qd_{j}}  = \frac{1}{2} \sum_{l\succeq i} \Bigg( \frac{ \partial \big(\v_{l}\times^* \big)\I_{l}}{\partial \qd_{j}} - \frac{ \partial \I_{l} \big(\v_{l} \times   \big)}{\partial \qd_{j}} + \frac{\partial \big(\I_{l} \v_{l} \big) \crff}{\partial \qd_{k}}    \Bigg)
\end{equation}

Using J8:

\begin{equation}
    \frac{\partial \B_{i}^{C}}{\partial \qd_{j}}  = \frac{1}{2} \sum_{l \succeq i} \big( \Phibar_{j}\timesfM \I_{l} -  \I_{l} \big(\Phibar_{j} \timesM \big) +  \big(\I_{l} \Phibar_{j} \big) \crffM   \big)
\end{equation}

Summing over the index $l$ results in:

\begin{equation}
    \frac{\partial \B_{i}^{C}}{\partial \qd_{j}}  = \Bten{\I_{i}^{C}}{\Phibar_{j}}
\end{equation}

where $\Bten{\I_{i}^{C}}{\Phibar_{j}} = \sum_{l\succeq i} \Bten{I_{l}}{\Phibar_j} $, and $\mathcal{B}_l [\Phibar_{j}]$ is given as:

\begin{equation}
    \Bten{I_{l}}{\Phibar_j} = \frac{1}{2} \big(\big(\Phibar_{j}\timesfM \big)\I_{l} - \I_{l} \big(\Phibar_{j} \timesM   \big) + \big(\I_{l} \Phibar_{j} \big) \crffM    \big)
    \label{Bi_phi_defn}
\end{equation}
For the case with $j \succ i$, we follow the similar procedure to get:
\begin{equation}
    \frac{\partial \B_{i}^{C}}{\partial \qd_{j}}  = \Bten{\I_{j}^{C}}{\Phibar_{j}}
\end{equation}

\pagebreak

\end{enumerate}

\pagebreak

\section{Second Order Partial Derivatives of ID wrt $\q$}
\label{SO_q_sec}
\noindent We take the partial derivative of Eq.~\ref{dtau_dq_1} and Eq.~\ref{dtau_dq_2} w.r.t.  $\q_{k}$ where the index $k$ results in three cases $k \preceq j \preceq i$ (A), $j \prec k \preceq i$ (B) and $j \preceq i \prec k$ (C). We individually solve for the three cases as follows:

\subsection{$k \preceq j \preceq i$ } 
\label{SO_q_case_A}
\subsubsection{1A}
\label{SO_q_case_1A}
Taking the partial derivative of Eq.~\ref{dtau_dq_1} wrt $\q_{k}$:

\begin{equation}
    \frac{\partial^2 \taubar_{i}}{\partial \q_{j} \partial \q_{k}} = \frac{\partial \big( \Phibar_{i} \T \big[ 2  \B_{i}^{C} \big]\Psibardot{}_{j}  \big)}{\partial \q_{k}} +\frac{\partial \big(\Phibar_{i} \T \I_{i}^{C} \Psibarddot{}_{j} \big)}{\partial \q_{k}}
    \label{SO_q_case1_eqn1}
\end{equation}

Considering the first term in Eq.~\ref{SO_q_case1_eqn1} and using the product rule of differentiation:

\begin{equation}
    \frac{\partial \big( \Phibar_{i} \T \big[ 2  \B_{i}^{C} \big]\Psibardot{}_{j} \big)}{\partial \q_{k}} = 2 \Bigg( \frac{\partial  \Phibar_{i} \T   }{\partial \q_{k}}\Bigg) \B_{i}^{C} \Psibardot{}_{j}+ 2\Phibar_{i} \T \Bigg(  \frac{\partial \B_{i}^{C}}{\partial \q_{k}}\Psibardot{}_{j} + \B_{i}^{C}\frac{\partial \Psibardot{}_{j}}{\partial \q_{k}}\Bigg)
\end{equation}

Using \ref{psii_dot}, \ref{Bic}, and \ref{PhiT}:

\begin{equation}
    \begin{aligned}
        \frac{\partial \big( \Phibar_{i} \T \big[ 2  \B_{i}^{C} \big]\Psibardot{}_{j} \big)}{\partial \q_{k}} &= 2 \big( -\Phibar_{i} \T \Phibar_{k} \timesfM \big) \B_{i}^{C} \Psibardot{}_{j}+ 
        2\Phibar_{i} \T \big( \Bten{\I_{i}^{C}}{\Psibardot_{k}}+\Phibar_{k} \timesfM \B_{i}^{C}- \B_{i}^{C}(\Phibar_{k} \timesM) \big) \Psibardot{}_{j} + \\&~~~~~~~~~
         2\Phibar_{i} \T \B_{i}^{C}\big(   \Psibardot_{k}   \timesM  \Phibar_{j} + \Phibar_{k} \timesM \Psibardot_{j} \big)
    \end{aligned}
\end{equation}

Expanding terms:

\begin{equation}
    \begin{aligned}
        \frac{\partial \big( \Phibar_{i} \T \big[ 2  \B_{i}^{C} \big]\Psibardot{}_{j} \big)}{\partial \q_{k}} &= -2  \Phibar_{i} \T \Phibar_{k} \timesfM \B_{i}^{C} \Psibardot{}_{j}+ 
        2\Phibar_{i} \T \big( \Bten{\I_{i}^{C}}{\Psibardot_{k}}- \B_{i}^{C}(\Phibar_{k} \timesM) \big) \Psibardot{}_{j} +2  \Phibar_{i} \T \Phibar_{k} \timesfM \B_{i}^{C} \Psibardot{}_{j} \\&~~~~~~~~~
         2\Phibar_{i} \T \B_{i}^{C}\big(   \Psibardot_{k}   \timesM  \Phibar_{j} + \Phibar_{k} \timesM \Psibardot_{j} \big)
    \end{aligned}
\end{equation}

Cancelling terms lead to:

\begin{equation}
        \frac{\partial \big( \Phibar_{i} \T \big[ 2  \B_{i}^{C} \big]\Psibardot{}_{j} \big)}{\partial \q_{k}} = 
        2\Phibar_{i} \T \big( \Bten{\I_{i}^{C}}{\Psibardot_{k}} \Psibardot{}_{j} +  \B_{i}^{C}  \Psibardot_{k} \timesM \Phibar_{j}  \big)
\label{SO_q_case1_eqn2}        
\end{equation}

Now, considering the second term in Eq.~\ref{SO_q_case1_eqn1}, and using the product rule:

\begin{equation}
\frac{\partial \big(\Phibar_{i} \T \I_{i}^{C} \Psibarddot{}_{j} \big)}{\partial \q_{k}} =   \Bigg( \frac{\partial  \Phibar_{i} \T   }{\partial \q_{k}}\Bigg) \I_{i}^{C} \Psibarddot{}_{j} + \Phibar_{i} \T  \Bigg(\frac{\partial \I_{i}^{C}}{\partial \q_{k}}\Psibarddot{}_{j} + \I_{i}^{C}\frac{\partial \Psibarddot{}_{j}}{\partial \q_{k}}  \Bigg)
\end{equation}

Using the identities \ref{Iic}, \ref{Psiddoti}, and \ref{PhiT}:

\begin{equation}
    \begin{aligned}
    \frac{\partial \big(\Phibar_{i} \T \I_{i}^{C} \Psibarddot{}_{j} \big)}{\partial \q_{k}} &=  \big( -\Phibar_{i} \T \Phibar_{k} \timesfM \big) \I_{i}^{C} \Psibarddot{}_{j} + 
    \Phibar_{i} \T \big(  \Phibar_{k} \timesfM \I_{i}^{C}- \I_{i}^{C}(\Phibar_{k} \timesM)   \big)\Psibarddot{}_{j} + \\&~~~~~~
    \Phibar_{i} \T \I_{i}^{C} \big(   \Psibarddot_{k}   \timesM  \Phibar_{j} + 2\Psibardot_{k} \timesM \Psibardot_{j}+\Phibar_{k} \timesM \Psibarddot_{j}\big)
    \end{aligned}
\end{equation}

Cancelling terms lead to:

\begin{equation}
    \begin{aligned}
    \frac{\partial \big(\Phibar_{i} \T \I_{i}^{C} \Psibarddot{}_{j} \big)}{\partial \q_{k}} &= 
    \Phibar_{i} \T \I_{i}^{C}  \Psibarddot_{k}   \timesM  \Phibar_{j} + 2\Phibar_{i} \T \I_{i}^{C} \Psibardot_{k} \timesM \Psibardot_{j}
    \end{aligned}
    \label{SO_q_case1_eqn3}
\end{equation}

Adding Eq.~\ref{SO_q_case1_eqn2} and \ref{SO_q_case1_eqn3}:

\begin{equation}
     \frac{\partial^2 \taubar_{i}}{\partial \q_{j} \partial \q_{k}} =  2\Phibar_{i} \T \big( \Bten{\I_{i}^{C}}{\Psibardot_{k}} \Psibardot{}_{j} +  \B_{i}^{C}  \Psibardot_{k} \timesM \Phibar_{j}  \big)+ \Phibar_{i} \T \I_{i}^{C}  \Psibarddot_{k}   \timesM  \Phibar_{j} + 2\Phibar_{i} \T \I_{i}^{C} \Psibardot_{k} \timesM \Psibardot_{j}
\end{equation}
Re-arranging common terms and using the property \ref{m8}:
\begin{equation}
       \frac{\partial^2 \taubar_{i}}{\partial \q_{j} \partial \q_{k}} =  2\Phibar_{i} \T \Bten{\I_{i}^{C}}{\Psibardot_{k}} \Psibardot{}_{j} -   2 \Phibar_{i}\T \B_{i}^{C} \big(  \Phibar_{j} \timesM \Psibardot_{k} \big) \Rten -   2 \Phibar_{i} \T \I_{i}^{C} \big(\Psibardot_{j} \timesM  \Psibardot_{k}\big)\Rten  - \Phibar_{i} \T \I_{i}^{C}   \big(\Phibar_{j} \timesM  \Psibarddot_{k} \big)\Rten 
     \label{SO_q_case1_eqn4}
\end{equation}
Using \ref{m26} for the first-term and \ref{m8} for the second term
\begin{equation}
     \frac{\partial^2 \taubar_{i}}{\partial \q_{j} \partial \q_{k}} =  2\Phibar_{i} \T (\Bten{\I_{i}^{C}}{\Psibardot_{j}} \Psibardot_{k})\Rten -2\Phibar_{i} \T\I_{i}^{C}(\Psibardot_{k}\timesM \Psibardot_{j}) -   2 \Phibar_{i}\T \B_{i}^{C} \big(  \Phibar_{j} \timesM \Psibardot_{k} \big) \Rten -   2 \Phibar_{i} \T \I_{i}^{C} \big(\Psibardot_{j} \timesM  \Psibardot_{k}\big)\Rten  - \Phibar_{i} \T \I_{i}^{C}   \big(\Phibar_{j} \timesM  \Psibarddot_{k} \big)\Rten  
\end{equation}
Cancelling terms and using \ref{m27} for the first-term
\begin{equation}
     \frac{\partial^2 \taubar_{i}}{\partial \q_{j} \partial \q_{k}} =  -2[\Psibardot_{j} \T (\Bten{\I_{i}^{C}}{\Phibar_{i}} \Psibardot_{k})\Rten]\Tten  -   2 \Phibar_{i}\T \B_{i}^{C} \big(  \Phibar_{j} \timesM \Psibardot_{k} \big) \Rten   - \Phibar_{i} \T \I_{i}^{C}   \big(\Phibar_{j} \timesM  \Psibarddot_{k} \big)\Rten  
\end{equation}
Using \ref{m8} followed by \ref{m21}  for the second and the third terms:
\begin{equation}
\begin{aligned}
     \frac{\partial^2 \taubar_{i}}{\partial \q_{j} \partial \q_{k}} &=  -\Big[\Psibardot_{j}\T [2\Bten{\I_{i}^{C}}{\Phibar_{i}}\Psibardot_{k}]\Rten \Big]\Tten - 2\big[\Phibar_{j}\T ((\B_{i}^{C,T} \Phibar_{i}) \crffM\Psibardot_{k} )\Rten\big] \Tten   - \Big[  \Phibar_{j}\T ((\I_{i}^{C}\Phibar_{i} ) \crffM\Psibarddot_{k} )\Rten \Big]\Tten  , (k \preceq j \preceq i)
     \end{aligned}
     \label{tau_SO_q_1A}
\end{equation}

\subsubsection{2A}
The expression for $ \frac{\partial^2 \taubar_{i}}{\partial \q_{k} \partial \q_{j}}$ for the condition $(k \preceq j \preceq i)$ can be calculated using $\frac{\partial^2 \taubar_{i}}{\partial \q_{j} \partial \q_{k}}$ (Eq.~\ref{tau_SO_q_1A}). A 3D rotation occurs as a result of the symmetry along the 2-3 dimension. 

\begin{equation}
  \frac{\partial^2 \taubar_{i}}{\partial \q_{k} \partial \q_{j}} =  \Bigg[\frac{\partial^2 \taubar_{i}}{\partial \q_{j} \partial \q_{k}} \Bigg]\Rten, (k \preceq j \prec i)
      \label{tau_SO_q_2A}
\end{equation}

\subsection{$j \prec k \preceq i$}
\label{SO_q_case_B}
\subsubsection{1B}
For this case, since $j \prec i$, we take the partial derivative of Eq.~\ref{dtau_dq_2} wrt $\q_{k}$ and use the product rule as:
\begin{equation}
     \frac{\partial^2 \taubar_{j}}{\partial \q_{i} \partial \q_{k}} = \Phibar_{j}\T \Bigg( 2 \frac{\partial \B_{i}^{C}}{\partial \q_{k}}\Psibardot_{i} + 2\B_{i}^{C}\frac{\partial \Psibardot_{i}}{\partial \q_{k}} + \frac{\partial \I_{i}^{C} }{\partial \q_{k}}\Psibarddot_{i}+ \I_{i}^{C}\frac{\partial \Psibarddot_{i} }{\partial \q_{k}}+  \bigg(\frac{\partial \f_{i}^{C}}{\partial \q_{k}} \bigg)\crffM \Phibar_{i}+ (\f_{i}^{C} )\crff\frac{\partial \Phibar_{i}}{\partial \q_{k}}\Bigg)
     \label{SO_q_case3_eqn1}
\end{equation}
Considering the first two terms in Eq.~\ref{SO_q_case3_eqn1}, and using the identities \ref{psii_dot} and \ref{Bic}:
\begin{equation}
    \begin{aligned}
        \Phibar_{j}\T \Bigg( 2 \frac{\partial \B_{i}^{C}}{\partial \q_{k}}\Psibardot_{i} + 2\B_{i}^{C}\frac{\partial \Psibardot_{i}}{\partial \q_{k}}  \Bigg) &= 2\Phibar_{j}\T \big(\Bten{\I_{i}^{C}}{\Psibardot_{k}}+\Phibar_{k} \timesfM \B_{i}^{C}- \B_{i}^{C}(\Phibar_{k} \timesM)  \big)\Psibardot_{i} + \\&~~~~
        2\Phibar_{j}\T \B_{i}^{C}\big( \Psibardot_{k}   \timesM  \Phibar_{i} + \Phibar_{k} \timesM \Psibardot_{i}  \big)
    \end{aligned}
\end{equation}
Cancelling terms result in:
\begin{equation}
        \Phibar_{j}\T \Bigg( 2 \frac{\partial \B_{i}^{C}}{\partial \q_{k}}\Psibardot_{i} + 2\B_{i}^{C}\frac{\partial \Psibardot_{i}}{\partial \q_{k}}  \Bigg) = 2\Phibar_{j}\T \left( \big(\Bten{\I_{i}^{C}}{\Psibardot_{k}}+\Phibar_{k} \timesfM \B_{i}^{C} \big)\Psibardot_{i} + \B_{i}^{C} \Psibardot_{k}   \timesM  \Phibar_{i}   \right)
     \label{SO_q_case3_eqn2}
\end{equation}
Considering the next two terms in Eq.~\ref{SO_q_case3_eqn1} and using the identities \ref{Iic} and \ref{Psiddoti} as:
\begin{equation}
       \Phibar_{j}\T \Bigg( \frac{\partial \I_{i}^{C} }{\partial \q_{k}}\Psibarddot_{i}+ \I_{i}^{C}\frac{\partial \Psibarddot_{i} }{\partial \q_{k}} \Bigg) = \Phibar_{j}\T \left( \big( \Phibar_{k} \timesfM \I_{i}^{C}- \I_{i}^{C}(\Phibar_{k} \timesM) \big) \Psibarddot_{i} +  \I_{i}^{C} \big(    \Psibarddot_{k}   \timesM  \Phibar_{i} + 2\Psibardot_{k} \timesM \Psibardot_{i}+\Phibar_{k} \timesM \Psibarddot_{i} \big) \right)
\end{equation}

Cancelling terms result in:

\begin{equation}
       \Phibar_{j}\T \Bigg( \frac{\partial \I_{i}^{C} }{\partial \q_{k}}\Psibarddot_{i}+ \I_{i}^{C}\frac{\partial \Psibarddot_{i} }{\partial \q_{k}} \Bigg) = \Phibar_{j}\T \left( \Phibar_{k} \timesfM \I_{i}^{C} \Psibarddot_{i} +  \I_{i}^{C} \big(    \Psibarddot_{k}   \timesM  \Phibar_{i} + 2\Psibardot_{k} \timesM \Psibardot_{i} \big) \right)
     \label{SO_q_case3_eqn3}
\end{equation}
Finally, considering the last two terms in Eq.~\ref{SO_q_case3_eqn1} and using the identities \ref{dphii} and \ref{fic} as:
\begin{equation}
  \Phibar_{j}\T \left( \left(\frac{\partial \f_{i}^{C}}{\partial \q_{k}} \right)\crffM \Phibar_{i} + (\f_{i}^{C} )\crff\frac{\partial \Phibar_{i}}{\partial \q_{k}} \right)= \Phibar_{j}\T  \big(  ( \I_{i}^{C} \Psibarddot_{k} + \Phibar_{k} \timesfM \f_{i}^{C} + 2 \B_{i}^{C} \Psibardot_{k})\crffM \Phibar_{i} + (\f_{i}^{C})\crff (\Phibar_{k}   \timesM \Phibar_{i}) \big)
\end{equation}
Flattening the term $\Phibar_{k} \timesfM \f_{i}^{C}$ to $(\Phibar_{k} \timesfM \f_{i}^{C})\Rten$, since it results into a matrix. Now, using the property \ref{m11} and expanding terms we get:
\begin{equation}
    \begin{aligned}
           \Phibar_{j}\T \left( \left(\frac{\partial \f_{i}^{C}}{\partial \q_{k}} \right)\crffM \Phibar_{i} + (\f_{i}^{C} )\crff\frac{\partial \Phibar_{i}}{\partial \q_{k}} \right) &= \Phibar_{j}\T  \big(  ( \I_{i}^{C} \Psibarddot_{k} + 2 \B_{i}^{C} \Psibardot_{k})\crffM \Phibar_{i} + \Phibar_{k} \timesfM \f_{i}^{C} \crff \Phibar_{i}- \f_{i}^{C} \crff \Phibar_{k} \timesM \Phibar_{i}+ \\&~~~~~~~
          (\f_{i}^{C})\crff (\Phibar_{k}   \timesM \Phibar_{i}) \big)
      \end{aligned}
\end{equation}
Cancelling terms lead to:
\begin{equation}
           \Phibar_{j}\T \left( \left(\frac{\partial \f_{i}^{C}}{\partial \q_{k}} \right)\crffM \Phibar_{i} + (\f_{i}^{C} )\crff\frac{\partial \Phibar_{i}}{\partial \q_{k}} \right) = \Phibar_{j}\T  \big(  ( \I_{i}^{C} \Psibarddot_{k} + 2 \B_{i}^{C} \Psibardot_{k})\crffM \Phibar_{i} + \Phibar_{k} \timesfM \f_{i}^{C} \crff \Phibar_{i} \big)
     \label{SO_q_case3_eqn4}
\end{equation}

Adding the terms in Eq.~\ref{SO_q_case3_eqn2},~\ref{SO_q_case3_eqn3}, and \ref{SO_q_case3_eqn4} to get:

\begin{equation}
    \begin{aligned}
         \frac{\partial^2 \taubar_{j}}{\partial \q_{i} \partial \q_{k}} &= \Phibar_{j}\T \Big( 2\big(\Bten{\I_{i}^{C}}{\Psibardot_{k}}+\Phibar_{k} \timesfM \B_{i}^{C} \big)\Psibardot_{i} + 2\B_{i}^{C} \Psibardot_{k}   \timesM  \Phibar_{i} +  \Phibar_{k} \timesfM \I_{i}^{C} \Psibarddot_{i} + \\&~~~~~ \I_{i}^{C} \big(    \Psibarddot_{k}   \timesM  \Phibar_{i} + 2\Psibardot_{k} \timesM \Psibardot_{i} \big) +  ( \I_{i}^{C} \Psibarddot_{k} + 2 \B_{i}^{C} \Psibardot_{k})\crffM \Phibar_{i} + \Phibar_{k} \timesfM \f_{i}^{C} \crff \Phibar_{i}\Big)
    \end{aligned}
\end{equation}

Re-arranging terms:

\begin{equation}
    \begin{aligned}
         \frac{\partial^2 \taubar_{j}}{\partial \q_{i} \partial \q_{k}} &= \Phibar_{j}\T \Big( 2\big(\Bten{\I_{i}^{C}}{\Psibardot_{k}}+\Phibar_{k} \timesfM \B_{i}^{C} \big)\Psibardot_{i} + 2\B_{i}^{C} (\Psibardot_{k}   \timesM  \Phibar_{i}) +2 (\B_{i}^{C} \Psibardot_{k}) \crffM \Phibar_{i} \\&~~~~~
         + \Phibar_{k} \timesfM \I_{i}^{C} \Psibarddot_{i}  + \I_{i}^{C} \big( 2\Psibardot_{k} \timesM \Psibardot_{i} \big) + 
          ( \I_{i}^{C} \Psibarddot_{k} )\crffM \Phibar_{i} +\I_{i}^{C}     \Psibarddot_{k}   \timesM  \Phibar_{i}+ \Phibar_{k} \timesfM \f_{i}^{C} \crff \Phibar_{i}\Big)
    \end{aligned}
\end{equation}

Using the property \ref{m6} and \ref{m8}, and re-arranging terms:

\begin{equation}
    \begin{aligned}
         \frac{\partial^2 \taubar_{j}}{\partial \q_{i} \partial \q_{k}} &= \Phibar_{j}\T \Big( 2\big(\Bten{\I_{i}^{C}}{\Psibardot_{k}}+\Phibar_{k} \timesfM \B_{i}^{C} \big)\Psibardot_{i}  - 2\B_{i}^{C}\big(\Phibar_{i}  \timesM   \Psibardot_{k}\big)\Rten +2 \big(  \Phibar_{i} \timesfM \B_{i}^{C}\Psibardot_{k}\big)\Rten\\&~~~~~
        +  \Phibar_{k} \timesfM \I_{i}^{C} \Psibarddot_{i}  + \I_{i}^{C} \big( 2\Psibardot_{k} \timesM \Psibardot_{i} \big) + 
         \big( \Phibar_{i} \timesfM \I_{i}^{C}\Psibarddot_{k} \big) \Rten    -\I_{i}^{C}  \big( \Phibar_{i}\timesM \Psibarddot_{k}\big)\Rten + \Phibar_{k} \timesfM \f_{i}^{C} \crff \Phibar_{i}\Big)
    \end{aligned}
\end{equation}
Collecting terms with commonality and using the property \ref{m8} :
\begin{equation}
    \begin{aligned}
         \frac{\partial^2 \taubar_{j}}{\partial \q_{i} \partial \q_{k}} &= \Phibar_{j}\T \Big( 2\Bten{\I_{i}^{C}}{\Psibardot_{k}}\Psibardot_{i} +\Phibar_{k} \timesfM \big( 2\B_{i}^{C}  \Psibardot_{i} +\I_{i}^{C} \Psibarddot_{i}+\f_{i}^{C} \crff \Phibar_{i}\big) \\&~~~~
         +2 \big(  \Phibar_{i} \timesfM \B_{i}^{C}\Psibardot_{k}\big)\Rten  -2 \B_{i}^{C}\big( \Phibar_{i}  \timesM \Psibardot_{k} \big)\Rten -2\I_{i}^{C} \big( \Psibardot_{i}\timesM \Psibardot_{k} \big)\Rten  + 
          \big( \Phibar_{i} \timesfM \I_{i}^{C}\Psibarddot_{k} \big) \Rten    -\I_{i}^{C}  \big( \Phibar_{i}\timesM \Psibarddot_{k}\big)\Rten   \Big)
    \end{aligned}
\end{equation}
Using property \ref{m18} on the last term:
\begin{equation}
    \begin{aligned}
         \frac{\partial^2 \taubar_{j}}{\partial \q_{i} \partial \q_{k}} &= \Phibar_{j}\T \Big( 2\Bten{\I_{i}^{C}}{\Psibardot_{k}}\Psibardot_{i} +\Phibar_{k} \timesfM \big( 2\B_{i}^{C}  \Psibardot_{i} +\I_{i}^{C} \Psibarddot_{i}+\f_{i}^{C} \crff \Phibar_{i}\big) 
         +2 \big(  \Phibar_{i} \timesfM \B_{i}^{C}\Psibardot_{k}\big)\Rten  \\&~~~~-2 \big( \B_{i}^{C}(\Phibar_{i}  \timesM \Psibardot_{k}) \big)\Rten -2\big( \I_{i}^{C} (\Psibardot_{i}\timesM \Psibardot_{k}) \big)\Rten  + 
          \big( \Phibar_{i} \timesfM \I_{i}^{C}\Psibarddot_{k} \big) \Rten    - \big( \I_{i}^{C} (\Phibar_{i}\timesM \Psibarddot_{k})\big)\Rten   \Big)
    \end{aligned}
\end{equation}
We switch the index $k$ and $j$ to get the case $k \prec j \preceq i$, and the expression changes to:
\begin{equation}
    \begin{aligned}
         \frac{\partial^2 \taubar_{k}}{\partial \q_{i} \partial \q_{j}} &= \Phibar_{k}\T \Big( 2\Bten{\I_{i}^{C}}{\Psibardot_{j}}\Psibardot_{i} +\Phibar_{j} \timesfM \big( 2\B_{i}^{C}  \Psibardot_{i} +\I_{i}^{C} \Psibarddot_{i}+\f_{i}^{C} \crff \Phibar_{i}\big) 
       +2 \big(  \Phibar_{i} \timesfM \B_{i}^{C}\Psibardot_{j}\big)\Rten \\&~~~~ -2 \big( \B_{i}^{C}(\Phibar_{i}  \timesM \Psibardot_{j}) \big)\Rten -2\big( \I_{i}^{C} (\Psibardot_{i}\timesM \Psibardot_{j}) \big)\Rten  + 
          \big( \Phibar_{i} \timesfM \I_{i}^{C}\Psibarddot_{j} \big) \Rten    - \big( \I_{i}^{C} (\Phibar_{i}\timesM \Psibarddot_{j})\big)\Rten   \Big)
    \end{aligned}
\end{equation}

Using \ref{m26} for the first term
\begin{equation}
    \begin{aligned}
         \frac{\partial^2 \taubar_{k}}{\partial \q_{i} \partial \q_{j}} &= \Phibar_{k}\T \Big( 2(\Bten{\I_{i}^{C}}{\Psibardot_{i}}\Psibardot_{j})\Rten-2\IC{i}(\Psibardot_{j}\timesM\Psibardot_{i}) +\Phibar_{j} \timesfM \big( 2\B_{i}^{C}  \Psibardot_{i} +\I_{i}^{C} \Psibarddot_{i}+\f_{i}^{C} \crff \Phibar_{i}\big) \\&~~~~
       +2 \big(  \Phibar_{i} \timesfM \B_{i}^{C}\Psibardot_{j}\big)\Rten  -2 \big( \B_{i}^{C}(\Phibar_{i}  \timesM \Psibardot_{j}) \big)\Rten -2\big( \I_{i}^{C} (\Psibardot_{i}\timesM \Psibardot_{j}) \big)\Rten  + \\&~~~~~~~
          \big( \Phibar_{i} \timesfM \I_{i}^{C}\Psibarddot_{j} \big) \Rten    - \big( \I_{i}^{C} (\Phibar_{i}\timesM \Psibarddot_{j})\big)\Rten   \Big)
    \end{aligned}
\end{equation}
Using \ref{m8} and \ref{m19} for the second term and cancelling,
\begin{equation}
    \begin{aligned}
         \frac{\partial^2 \taubar_{k}}{\partial \q_{i} \partial \q_{j}} &= \Phibar_{k}\T \Big( 2(\Bten{\I_{i}^{C}}{\Psibardot_{i}}\Psibardot_{j})\Rten+\Phibar_{j} \timesfM \big( 2\B_{i}^{C}  \Psibardot_{i} +\I_{i}^{C} \Psibarddot_{i}+\f_{i}^{C} \crff \Phibar_{i}\big) 
       +2 \big(  \Phibar_{i} \timesfM \B_{i}^{C}\Psibardot_{j}\big)\Rten  -2 \big( \B_{i}^{C}(\Phibar_{i}  \timesM \Psibardot_{j}) \big)\Rten   + \\&~~~~~~~
          \big( \Phibar_{i} \timesfM \I_{i}^{C}\Psibarddot_{j} \big) \Rten    - \big( \I_{i}^{C} (\Phibar_{i}\timesM \Psibarddot_{j})\big)\Rten   \Big)
    \end{aligned}
\end{equation}
Collecting terms results in the final expression:
\begin{equation}
    \begin{aligned}
         \frac{\partial^2 \taubar_{k}}{\partial \q_{i} \partial \q_{j}} &= \Phibar_{k}\T \Big( \big[2(\Bten{\I_{i}^{C}}{\Psibardot_{i}}+ \Phibar_{i} \timesfM \B_{i}^{C} - \B_{i}^{C}\Phibar_{i}  \timesM)\Psibardot_{j} + \big( \Phibar_{i} \timesfM \I_{i}^{C}   -\I_{i}^{C}   \Phibar_{i}\timesM \big)\Psibarddot_{j} \big]\Rten +\\&~~~~
          \Phibar_{j} \timesfM \big( 2\B_{i}^{C}  \Psibardot_{i} +\I_{i}^{C} \Psibarddot_{i}+\f_{i}^{C} \crff \Phibar_{i}\big) \Big)  , (k \prec j \preceq i)
    \end{aligned}
    \label{tau_SO_q_1B}
\end{equation}

\subsubsection{2B}
The expression for $\frac{\partial^2 \taubar_{k}}{\partial \q_{j} \partial \q_{i}} $ for the condition $(k \prec j \preceq i)$ can be calculated using $\frac{\partial^2 \taubar_{k}}{\partial \q_{i} \partial \q_{j}}$ (Eq.~\ref{tau_SO_q_1B}). A 3D rotation occurs as a result of the symmetry along the 2-3 dimension. Since the previous case (Eq.~\ref{tau_SO_q_1B}) covers the condition $j=i$ , this case is solved for a stricter condition $(k \prec j \prec i)$.

\begin{equation}
     \frac{\partial^2 \taubar_{k}}{\partial \q_{j} \partial \q_{i}} =  \Bigg[\frac{\partial^2 \taubar_{k}}{\partial \q_{i} \partial \q_{j}} \Bigg]\Rten, (k \prec j \prec i)
      \label{tau_SO_q_2B}
\end{equation}

\subsection{$j \preceq i \prec k$ }
\label{SO_q_case_C}

\subsubsection{1C}

For this case we take the partial derivatives of Eq.~\ref{dtau_dq_1} and apply the product rule of differentiation to get:

\begin{equation}
  \frac{\partial^2 \taubar_{i}}{\partial \q_{j} \partial \q_{k}} = 2\Phibar_{i} \T \Bigg(  \frac{\partial \B_{i}^{C}}{\partial \q_{k}}\Psibardot{}_{j} \Bigg) +  \Phibar_{i} \T  \Bigg(\frac{\partial \I_{i}^{C}}{\partial \q_{k}}\Psibarddot{}_{j} \Bigg)
\end{equation}

Applying identities \ref{Iic} and \ref{Bic} to get:

\begin{equation}
\begin{aligned}
  \frac{\partial^2 \taubar_{i}}{\partial \q_{j} \partial \q_{k}} &= 2\Phibar_{i} \T \big(   \Bten{\I_{k}^{C}}{\Psibardot_{k}}+\Phibar_{k} \timesfM \B_{k}^{C}- \B_{k}^{C}(\Phibar_{k} \timesM) \big)\Psibardot_{j} +  \Phibar_{i} \T  \big( \Phibar_{k} \timesfM \I_{k}^{C}- \I_{k}^{C}(\Phibar_{k} \timesM)\big) \Psibarddot_{j} 
  \end{aligned}
  \label{SO_q_case2_eqn1}
\end{equation}

 First switching the indices $k$ and $j$, followed by $j$ and $i$ to get the case $k \preceq j \prec i$, Eq.~\ref{SO_q_case2_eqn1} now converts to:

\begin{equation}
  \frac{\partial^2 \taubar_{j}}{\partial \q_{k} \partial \q_{i}} = 2\Phibar_{j} \T \big(   \Bten{\I_{i}^{C}}{\Psibardot_{i}}+\Phibar_{i} \timesfM \B_{i}^{C}- \B_{i}^{C}(\Phibar_{i} \timesM) \big)\Psibardot_{k} + 
  \Phibar_{j} \T  \big( \Phibar_{i} \timesfM \I_{i}^{C}- \I_{i}^{C}(\Phibar_{i} \timesM)\big) \Psibarddot_{k} , (k \preceq j \prec i)
  \label{tau_SO_q_1C}
\end{equation}

\subsubsection{2C}
To exploit the block Hessian symmetry, $\frac{\partial^2 \taubar_{j}}{\partial \q_{i} \partial \q_{k}}$ is calculated using $ \frac{\partial^2 \taubar_{j}}{\partial \q_{k} \partial \q_{i}}$ (Eq.~\ref{tau_SO_q_1C}). A 3D rotation along the 2-3 dimension occurs as a result of it.
\begin{equation}
    \frac{\partial^2 \taubar_{j}}{\partial \q_{i} \partial \q_{k}} =  \Bigg[  \frac{\partial^2 \taubar_{j}}{\partial \q_{k} \partial \q_{i}} \Bigg]\Rten, (k \preceq j \prec i)
      \label{tau_SO_q_2C}
\end{equation}

\pagebreak

\section{Second Order Partial Derivatives of ID wrt $\qd$}
\label{SO_qd_sec}
Now we take the partial derivatives of Eq.~\ref{dtau_dqd_1} and Eq.~\ref{dtau_dqd_2} wrt $\qd_{k}$ to get the second-order partial derivative of $\taubar_{i}$ w.r.t $\qd$. We consider 5 cases as follows:

\subsection{$k \prec j \preceq i$}
\label{SO_qd_case1}

\subsubsection{1A}
We take the partial derivative of Eq.~\ref{dtau_dqd_1} wrt $\qd_{k}$ using the product rule of differentiation as:

\begin{equation}
    \frac{\partial^2 \taubar_{i}}{\partial \qd_{j} \partial \qd_{k}} = \Phibar_{i}\T \left[2 \frac{\partial \B_{i}^{C}}{\partial \qd_{k}} \Phibar_{j} +  \I_{i}^{C} \left(   \frac{\partial \Psibardot_{j}}{\partial \qd_{k}}  + \frac{\partial \Phibardot_{j} }{\partial \qd_{k}} \right) \right]
\end{equation}

Using the identities \ref{Phidot_qdj}-\ref{Bic_qdj} we get:

\begin{equation}
    \frac{\partial^2 \taubar_{i}}{\partial \qd_{j} \partial \qd_{k}} = \Phibar_{i}\T \left[2 \Bten{\I_{i}^C}{\Phibar_{k}} \Phibar_{j} +  \I_{i}^{C} \left(   \Phibar_{k} \timesM \Phibar_{j}  + \Phibar_{k} \timesM \Phibar_{j} \right) \right]
\end{equation}

Upon simplifying:

\begin{equation}
    \frac{\partial^2 \taubar_{i}}{\partial \qd_{j} \partial \qd_{k}} = 2 \Phibar_{i}\T \left[ \Bten{\I_{i}^C}{\Phibar_{k}} \Phibar_{j} +  \I_{i}^{C} \left(   \Phibar_{k} \timesM \Phibar_{j}   \right) \right]
\end{equation}
Using \ref{m26}

\begin{equation}
    \frac{\partial^2 \taubar_{i}}{\partial \qd_{j} \partial \qd_{k}} = 2 \Phibar_{i}\T \Big[ (\Bten{\I_{i}^{C}}{\Phibar_{j}}  \Phibar_{k})\Rten -\I_{i}^{C} (   \Phibar_{k} \timesM \Phibar_{j}   )+  \I_{i}^{C} (   \Phibar_{k} \timesM \Phibar_{j}   ) \Big]
\end{equation}
Using \ref{m27}, and cancelling terms, the final expression is:
\begin{equation}
    \frac{\partial^2 \taubar_{i}}{\partial \qd_{j} \partial \qd_{k}} =-\left[\Phibar_{j}\T (2\Bten{\I_{i}^C}{\Phibar_{i}} \Phibar_{k})\Rten \right]\Tten, (k \prec j \preceq i)
    \label{SO_qd_case_1A}
\end{equation}

\subsubsection{2A}
Exploiting the block Hessian symmetry, $ \frac{\partial^2 \taubar_{i}}{\partial \qd_{k} \partial \qd_{j}}$ is calculated using  $\frac{\partial^2 \taubar_{i}}{\partial \qd_{j} \partial \qd_{k}}$ (Eq.~\ref{SO_qd_case_1A}). Due to the symmetry between the 2-3 dimensions, a 3D rotation occurs as:

\begin{equation}
    \frac{\partial^2 \taubar_{i}}{\partial \qd_{k} \partial \qd_{j}} = \left[ \frac{\partial^2 \taubar_{i}}{\partial \qd_{j} \partial \qd_{k}} \right]\Rten, (k \prec j \preceq i)
        \label{SO_qd_case_2A}
\end{equation}

\subsection{$k = j \preceq i$}

This case is similar to 1A except for the term $\frac{\partial \Psibardot_{j}}{\partial \qd_{k}}$, which results into 0, based on the identity \ref{Psidot_qdj}. Hence the expression for this case results in:

\begin{equation}
    \frac{\partial^2 \taubar_{i}}{\partial \qd_{j} \partial \qd_{k}} =  \Phibar_{i}\T \left[ 2\Bten{\I_{i}^C}{\Phibar_{k}} \Phibar_{j} +  \I_{i}^{C} \left(   \Phibar_{k} \timesM \Phibar_{j}   \right) \right]
\end{equation}

Using \ref{m27}

\begin{equation}
    \frac{\partial^2 \taubar_{i}}{\partial \qd_{j} \partial \qd_{k}} =  \Phibar_{i}\T \Big[ 2(\Bten{\I_{i}^{C}}{\Phibar_{j}} \Phibar_{k})\Rten -2\I_{i}^{C}(\Phibar_{k}\timesM\Phibar_{j})  +  \I_{i}^{C} (   \Phibar_{k} \timesM \Phibar_{j} ) \Big]
\end{equation}

simplifying,

\begin{equation}
    \frac{\partial^2 \taubar_{i}}{\partial \qd_{j} \partial \qd_{k}} =  \Phibar_{i}\T \big[ 2(\Bten{\I_{i}^{C}}{\Phibar_{j}} \Phibar_{k})\Rten\big] - \Phibar_{i}\T\I_{i}^{C}(\Phibar_{k}\timesM\Phibar_{j}) 
\end{equation}

using \ref{m27} for the first term, and \ref{m8} for the second-term

\begin{equation}
    \frac{\partial^2 \taubar_{i}}{\partial \qd_{j} \partial \qd_{k}} =  -\Phibar_{j}\T \big[ 2(\Bten{\I_{i}^{C}}{\Phibar_{i}} \Phibar_{k})\Rten\big]\Tten + \Phibar_{i}\T\I_{i}^{C}(\Phibar_{j}\timesM\Phibar_{k})\Rten 
\end{equation}

Using \ref{m13} for the second term

\begin{equation}
    \frac{\partial^2 \taubar_{i}}{\partial \qd_{j} \partial \qd_{k}} =  -\Phibar_{j}\T \big[ 2(\Bten{\I_{i}^{C}}{\Phibar_{i}} \Phibar_{k})\Rten\big]\Tten + [\Phibar_{j}\T(\Phibar_{k}\timesfM\I_{i}^{C}\Phibar_{i})]\Tten 
\end{equation}

Using \ref{m5} for the second term

\begin{equation}
    \frac{\partial^2 \taubar_{i}}{\partial \qd_{j} \partial \qd_{k}} =  -\Phibar_{j}\T \big[ 2(\Bten{\I_{i}^{C}}{\Phibar_{i}} \Phibar_{k})\Rten\big]\Tten + [\Phibar_{j}\T(\I_{i}^{C}\Phibar_{i}\crffM\Phibar_{k})\Rten]\Tten 
\end{equation}
Expanding the first-term by using the definition of $\Bten{I_i^{C}}{\Phibar_i}$, and simplifying
Final expression can be written as:

\begin{equation}
    \frac{\partial^2 \taubar_{i}}{\partial \qd_{j} \partial \qd_{k}} =   -\left[ \Phibar_{j}\T (  (\Phibar_{i} \timesfM \I_{i}^{C}   - \I_{i}^{C}\Phibar_{i} \timesM ) \Phibar_{k} )\Rten\right]\Tten , (k = j \preceq i)
            \label{SO_qd_case_B}
\end{equation}

\subsection{$j \prec k \prec i$}
\label{SO_qd_case4}
\subsubsection{1C}

For this case, since $j \prec i$, we take the partial derivative of Eq.~\ref{dtau_dqd_2} w.r.t $\qd_{k}$, noting that $\frac{\partial \Phibar_{i}}{\partial \qd_{k}} = \frac{\partial \IC{i}}{\partial \qd_{k}}=0$ as:

\begin{equation}
    \frac{\partial^2 \taubar_{j}}{\partial \qd_{i} \partial \qd_{k}} = \Phibar_{j}\T \left[2 \frac{\partial \B_{i}^{C}}{\partial \qd_{k}} \Phibar_{i} +  \I_{i}^{C} \left(   \frac{\partial \Psibardot_{i}}{\partial \qd_{k}}  + \frac{\partial \Phibardot_{i} }{\partial \qd_{k}} \right) \right]
\end{equation}

Using identities \ref{Phidot_qdj}-\ref{Bic_qdj} results in:

\begin{equation}
    \frac{\partial^2 \taubar_{j}}{\partial \qd_{i} \partial \qd_{k}} = \Phibar_{j}\T \left[2 \Bten{\I_{i}^C}{\Phibar_{k}} \Phibar_{i} +  \I_{i}^{C} \left(   \Phibar_{k} \timesM \Phibar_{i}  + \Phibar_{k} \timesM \Phibar_{i} \right) \right]
\end{equation}

Upon simplifying:

\begin{equation}
    \frac{\partial^2 \taubar_{j}}{\partial \qd_{i} \partial \qd_{k}} = 2 \Phibar_{j}\T \left[\Bten{\I_{i}^C}{\Phibar_{k}} \Phibar_{i} +  \I_{i}^{C} \left(   \Phibar_{k} \timesM \Phibar_{i} \right) \right]
\end{equation}

Switching the indices $k$ and $j$ to get the case $k \prec j \prec i$ for the term $ \frac{\partial^2 \taubar_{k}}{\partial \qd_{i} \partial \qd_{j}} $ as:

\begin{equation}
  \frac{\partial^2 \taubar_{k}}{\partial \qd_{i} \partial \qd_{j}} = 2 \Phibar_{k}\T \left[\Bten{\I_{i}^{C}}{\Phibar_{j}} \Phibar_{i} +  \I_{i}^{C} \left(   \Phibar_{j} \timesM \Phibar_{i} \right) \right]  
\end{equation}

Using \ref{m26}, 

\begin{equation}
  \frac{\partial^2 \taubar_{k}}{\partial \qd_{i} \partial \qd_{j}} = 2 \Phibar_{k}\T \Big[ \big[\Bten{\I_{i}^{C}}{\Phibar_{i}} \Phibar_{j}\big]\Rten -\I_{i}^{C} (   \Phibar_{j} \timesM \Phibar_{i})+  \I_{i}^{C} (   \Phibar_{j} \timesM \Phibar_{i} ) \Big]   
\end{equation}
Cancelling terms, 

\begin{equation}
  \frac{\partial^2 \taubar_{k}}{\partial \qd_{i} \partial \qd_{j}} =  \Phibar_{k}\T \left[2\Bten{\I_{i}^C}{\Phibar_{i}} \Phibar_{j} \right]\Rten, (k \prec j \prec i) 
     \label{SO_qd_case_1C}
\end{equation}

\subsubsection{2C}
To get the term $ \frac{\partial^2 \taubar_{k}}{\partial \qd_{j} \partial \qd_{i}}$ for the condition $k \prec j \prec i$, we use the block Hessian symmetry for the term $\frac{\partial^2 \taubar_{k}}{\partial \qd_{i} \partial \qd_{j}}$ (Eq.~\ref{SO_qd_case_1C}). A 3D rotation occurs due to symmetry between the 2-3 dimension.

\begin{equation}
     \frac{\partial^2 \taubar_{k}}{\partial \qd_{j} \partial \qd_{i}} = \left[  \frac{\partial^2 \taubar_{k}}{\partial \qd_{i} \partial \qd_{j}} \right]\Rten, (k \prec j \prec i)
  \label{SO_qd_case_2C}
\end{equation}

\subsection{$j \prec k = i$}
This case is similar to 1C except the expression for $\frac{\partial \Psibardot_{i}}{\partial \qd_{k}}$, which is 0 using the identity \ref{Psidot_qdj}.  Switching the indices $j$ and $k$ results into the case $k \prec j = i$, and the expression:

\begin{equation}
  \frac{\partial^2 \taubar_{k}}{\partial \qd_{i} \partial \qd_{j}} =  \Phibar_{k}\T \left[2\Bten{\I_{i}^{C}}{\Phibar_{j}} \Phibar_{i} +  \I_{i}^{C} \left(   \Phibar_{j} \timesM \Phibar_{i} \right) \right]  
\end{equation}

Using \ref{m27}

\begin{equation}
  \frac{\partial^2 \taubar_{k}}{\partial \qd_{i} \partial \qd_{j}} =  \Phibar_{k}\T \Big[2 \big(\Bten{\I_{i}^{C}}{\Phibar_{i}} \Phibar_{j}\big)\Rten-2 \I_{i}^{C} \big(   \Phibar_{j} \timesM \Phibar_{i}\big) +  \I_{i}^{C} \big(   \Phibar_{j} \timesM \Phibar_{i} \Big) \Big]  
\end{equation}

Simplifying

\begin{equation}
  \frac{\partial^2 \taubar_{k}}{\partial \qd_{i} \partial \qd_{j}} =  \Phibar_{k}\T \Big[2 \big(\Bten{\I_{i}^{C}}{\Phibar_{i}} \Phibar_{j}\big)\Rten- \I_{i}^{C} \big(   \Phibar_{j} \timesM \Phibar_{i}\big) \Big]  
\end{equation}
Using \ref{m8} and \ref{m19}

\begin{equation}
  \frac{\partial^2 \taubar_{k}}{\partial \qd_{i} \partial \qd_{j}} =  \Phibar_{k}\T \Big[2 \big(\Bten{\I_{i}^{C}}{\Phibar_{i}} \Phibar_{j}\big)\Rten+ \big( \I_{i}^{C}   \Phibar_{i} \timesM \Phibar_{j}\big)\Rten \Big]  
\end{equation}

Using the definition of $\Bten{\IC{i}}{\Phibar_{i}}$ and simplifying, the final eqn is:

The final equation is:

\begin{equation}
  \frac{\partial^2 \taubar_{k}}{\partial \qd_{i} \partial \qd_{j}} =  \Phibar_{k}\T \left[ \left((\I_{i}^{C}\Phibar_{i})\crffM   +\Phibar_{i} \timesfM \I_{i}^{C}\right) \Phibar_{j}  \right]\Rten , (k \prec j = i)
       \label{SO_qd_case_D}
\end{equation}

\subsection{$j \preceq i \prec k$}
\subsubsection{1E}

Taking the partial derivative of  $\frac{\partial \taubar_{i}}{\partial \qd_{j}}$ (Eq.~\ref{dtau_dqd_1}) wrt $\qd_{k}$ for this case results in:

\begin{equation}
    \frac{\partial^2 \taubar_{i}}{\partial \qd_{j} \partial \qd_{k}} = \Phibar_{i}\T \left[2 \frac{\partial \B_{i}^{C}}{\partial \qd_{k}} \Phibar_{j} +  \I_{i}^{C} \left(   \frac{\partial \Psibardot_{j}}{\partial \qd_{k}}  + \frac{\partial \Phibardot_{j} }{\partial \qd_{k}} \right) \right]
\end{equation}

Using the identities \ref{Phidot_qdj}-\ref{Bic_qdj} results in:

\begin{equation}
    \frac{\partial^2 \taubar_{i}}{\partial \qd_{j} \partial \qd_{k}} = \Phibar_{i}\T \left[2\Bten{\I_{i}^C}{\Phibar_k} \Phibar_{j}\right]
\end{equation}

Switching the indices $j$ and $k$, and then $i$ and $j$ to get the case $k \preceq j \prec i$ as:

\begin{equation}
     \frac{\partial^2 \taubar_{j}}{\partial \qd_{k} \partial \qd_{i}} = \Phibar_{j}\T \left[2\Bten{\I_{i}^C}{\Phibar_{i}} \Phibar_{k}\right], (k \preceq j \prec i)
          \label{SO_qd_case_1E}
\end{equation}

\subsubsection{2E}
Exploiting the block Hessian symmetry, we get the term $\frac{\partial^2 \taubar_{j}}{\partial \qd_{i} \partial \qd_{k}} $ for the case $k \preceq j \prec i$. A 3D rotation occurs as a result of the symmetry between 2-3 dimensions.

\begin{equation}
     \frac{\partial^2 \taubar_{j}}{\partial \qd_{i} \partial \qd_{k}} = \left[   \frac{\partial^2 \taubar_{j}}{\partial \qd_{k} \partial \qd_{i}}  \right]\Rten, (k \preceq j \prec i)
               \label{SO_qd_case_2E}
\end{equation}

\pagebreak

\section{Cross Second-Order Partial Derivatives of ID w.r.t. $\qd$ and $\q$}
\label{MSO_sec}
In this section, the second-order cross partial derivatives of ID w.r.t $\q$ and $\qd$ are derived. For simplifying the algebra, first-order partial derivative of $\frac{\partial \taubar}{\partial \qd}$ (Eq.~\ref{dtau_dqd_1},\ref{dtau_dqd_2}) w.r.t $\q$ are taken. Hence, 3 cases each for $\frac{\partial \taubar_{i}}{\partial \qd_{j}}$ and $\frac{\partial \taubar_{j}}{\partial \qd_{i}}$ w.r.t $\q_{k}$ are considered pertaining to $k \preceq j \preceq i$, $j \prec k \preceq i$, and $j \preceq i \prec k$ as follows:

\subsection{$k \preceq j \preceq i$, 1A, 2A}
\subsubsection{1A}

Since for this case $j \preceq i$, we first take the derivative of $\frac{\partial \taubar_{i}}{\partial \qd_{j}}$ (Eq.~\ref{dtau_dqd_1}) w.r.t $\q_{k}$ as:

\begin{equation}
     \frac{\partial^2 \taubar_{i}}{ \partial \qd_{j}\partial \q_{k}} =  \frac{\partial  \Phibar_{i} \T   }{\partial \q_{k}}(2 \B_{i}^{C} \Phibar_{j} +  \I_{i}^{C} (\Psibardot_{j} + \Phibardot_{j} ))+  \Phibar_{i}\T \Bigg(2 \frac{\partial \B_{i}^{C}}{\partial \q_{k}} \Phibar_{j} + 2\B_{i}^{C} \frac{\partial \Phibar_{j}}{\partial \q_{k}} + \frac{\partial \I_{i}^{C}}{\partial\q_{k}} (\Psibardot_{j} + \Phibardot_{j} )  + \I_{i}^{C} \bigg( \frac{\partial \Psibardot_{j}}{\partial \q_{k}}+\frac{\partial \Phibardot_{j}}{\partial \q_{k}} \bigg) \Bigg) 
      \label{MSO_case1_1}
\end{equation}

Considering only the first term $\frac{\partial  \Phibar_{i} \T   }{\partial \q_{k}}(2 \B_{i}^{C} \Phibar_{j} +  \I_{i}^{C} (\Psibardot_{j} + \Phibardot_{j} ))$ in Eq.~\ref{MSO_case1_1} and using the identity \ref{PhiT}:

\begin{equation}
     \frac{\partial  \Phibar_{i} \T   }{\partial \q_{k}}(2 \B_{i}^{C} \Phibar_{j} +  \I_{i}^{C} (\Psibardot_{j} + \Phibardot_{j} )) =   -\Phibar_{i} \T \Phibar_{k} \timesfM \big(2 \B_{i}^{C} \Phibar_{j} +  \I_{i}^{C} (\Psibardot_{j} + \Phibardot_{j} ) \big)
     \label{MSO_case1_2}
\end{equation}

Considering the next two terms in Eq.~\ref{MSO_case1_1} and using the identities \ref{dphii} and \ref{Bic} as:

\begin{equation}
    \begin{aligned}
     \Phibar_{i}\T \Bigg(2 \frac{\partial \B_{i}^{C}}{\partial \q_{k}} \Phibar_{j} + 2\B_{i}^{C} \frac{\partial \Phibar_{j}}{\partial \q_{k}} \Bigg) =     \Phibar_{i}\T \big(2 \big(   \Bten{\I_{i}^{C}}{\Psibardot_{k}}+\Phibar_{k} \timesfM \B_{i}^{C}- \B_{i}^{C}(\Phibar_{k} \timesM) \big) \Phibar_{j} + 2\B_{i}^{C} \big(  \Phibar_{k}   \timesM \Phibar_{j}\big) \big)
    \end{aligned} 
\end{equation}

Cancelling terms lead to:

\begin{equation}
    \begin{aligned}
     \Phibar_{i}\T \Bigg(2 \frac{\partial \B_{i}^{C}}{\partial \q_{k}} \Phibar_{j} + 2\B_{i}^{C} \frac{\partial \Phibar_{j}}{\partial \q_{k}} \Bigg) =     2\Phibar_{i}\T \big(   \Bten{\I_{i}^{C}}{\Psibardot_{k}}+\Phibar_{k} \timesfM \B_{i}^{C} \big)\Phibar_{j} 
    \end{aligned} 
     \label{MSO_case1_3}
\end{equation}

Now considering the next term in Eq.~\ref{MSO_case1_1} and using the identity \ref{Iic} as:

\begin{equation}
    \begin{aligned}
\Phibar_{i} \T  \frac{\partial \I_{i}^{C}}{\partial\q_{k}} (\Psibardot_{j} + \Phibardot_{j} )  =   \Phibar_{i} \T ( \Phibar_{k} \timesfM \I_{i}^{C}- \I_{i}^{C}(\Phibar_{k} \timesM) )  (\Psibardot_{j} + \Phibardot_{j} ) 
    \end{aligned} 
\end{equation}

Expanding:

\begin{equation}
    \begin{aligned}
  \Phibar_{i} \T \frac{\partial \I_{i}^{C}}{\partial\q_{k}} (\Psibardot_{j} + \Phibardot_{j} )  =    \Phibar_{i} \T \Phibar_{k} \timesfM \I_{i}^{C} \Psibardot_{j} - \Phibar_{i} \T\I_{i}^{C}(\Phibar_{k} \timesM) \Psibardot_{j} +  \Phibar_{i} \T \Phibar_{k} \timesfM \I_{i}^{C} \Phibardot_{j} -\Phibar_{i} \T \I_{i}^{C}(\Phibar_{k} \timesM) \Phibardot_{j}
    \end{aligned}
     \label{MSO_case1_4}
\end{equation}

Now considering the last term in Eq.~\ref{MSO_case1_1}, and using the identity \ref{dphii_dot} and \ref{psii_dot} as:

\begin{equation}
  \Phibar_{i} \T \I_{i}^{C} \bigg( \frac{\partial \Psibardot_{j}}{\partial \q_{k}}+\frac{\partial \Phibardot_{j}}{\partial \q_{k}} \bigg) =  \Phibar_{i} \T \I_{i}^{C} (\Psibardot_{k}   \timesM  \Phibar_{j} + \Phibar_{k} \timesM \Psibardot_{j}) + \Phibar_{i} \T\I_{i}^{C}( \Psibardot_{k}   \timesM \Phibar_{j}+ \Phibar_{k} \timesM \Phibardot_{j})
     \label{MSO_case1_5}
\end{equation}

Adding the terms in Eq.~\ref{MSO_case1_2},~\ref{MSO_case1_3},~\ref{MSO_case1_4},~\ref{MSO_case1_5} results in:

\begin{equation}
\begin{aligned}
      \frac{\partial^2 \taubar_{i}}{ \partial \qd_{j}\partial \q_{k}} &= -\Phibar_{i} \T \Phibar_{k} \timesfM \big(2 \B_{i}^{C} \Phibar_{j} +  \I_{i}^{C} (\Psibardot_{j} + \Phibardot_{j} ) \big) + 
       2\Phibar_{i}\T \big(   \Bten{\I_{i}^{C}}{\Psibardot_{k}}+\Phibar_{k} \timesfM \B_{i}^{C} \big)\Phibar_{j}  +\\&~~~
         \Phibar_{i} \T \Phibar_{k} \timesfM \I_{i}^{C} \Psibardot_{j} - \Phibar_{i} \T\I_{i}^{C}(\Phibar_{k} \timesM) \Psibardot_{j} +  \Phibar_{i} \T \Phibar_{k} \timesfM \I_{i}^{C} \Phibardot_{j} -\Phibar_{i} \T \I_{i}^{C}(\Phibar_{k} \timesM) \Phibardot_{j}+ \\&
          \Phibar_{i} \T \I_{i}^{C} (\Psibardot_{k}   \timesM  \Phibar_{j} + \Phibar_{k} \timesM \Psibardot_{j}) + \Phibar_{i} \T\I_{i}^{C}( \Psibardot_{k}   \timesM \Phibar_{j}+ \Phibar_{k} \timesM \Phibardot_{j})
      \end{aligned}
\end{equation}

Cancelling terms lead to:

\begin{equation}
    \frac{\partial^2 \taubar_{i}}{ \partial \qd_{j}\partial \q_{k}} =    2\Phibar_{i}\T \big(   \Bten{\I_{i}^{C}}{\Psibardot_{k}}\Phibar_{j}  + \I_{i}^{C} (\Psibardot_{k}   \timesM  \Phibar_{j} ) \big)
\end{equation}

Using \ref{m26}

\begin{equation}
    \frac{\partial^2 \taubar_{i}}{ \partial \qd_{j}\partial \q_{k}} =    2\Phibar_{i}\T \Big(   (\Bten{\I_{i}^{C}}{\Phibar_{j}}{\Psibardot_{k}})\Rten -\I_{i}^{C} (\Psibardot_{k}   \timesM  \Phibar_{j} )   + \I_{i}^{C} (\Psibardot_{k}   \timesM  \Phibar_{j} ) \Big) 
\end{equation}

Cancelling terms and using \ref{m27} for the first-term

\begin{equation}
    \frac{\partial^2 \taubar_{i}}{ \partial \qd_{j}\partial \q_{k}} =    -\big[ \Phibar_{j}\T \big( 2\Bten{\I_{i}^C}{\Phibar_{i}} \Psibardot_{k}\big) \Rten  \big] \Tten , (k \preceq j \preceq i)
    \label{MSO_eqn_1}
\end{equation}

\subsubsection{2A ($j \ne i$)}

For this case, we take the partial derivative of $\frac{\partial \taubar_{j}}{\partial \qd_{i}}$ (Eq.~\ref{dtau_dqd_2}) w.r.t $\q_{k}$. Due to $j \ne i$, the condition becomes $k \preceq j \prec i$.

\begin{equation}
    \frac{\partial^2 \taubar_{j}}{\partial \qd_{i} \partial \q_{k}} = \frac{\partial  \Phibar_{j} \T   }{\partial \q_{k}} \big(2 \B_{i}^{C} \Phibar_{i} +  \I_{i}^{C} (    \Psibardot_{i} + \Phibardot_{i} ) \big) +  \Phibar_{j} \T  \Bigg(2 \frac{\partial \B_{i}^{C}}{\partial \q_{k}} \Phibar_{i} + 2\B_{i}^{C} \frac{\partial \Phibar_{i}}{\partial \q_{k}} +  \frac{\partial \I_{i}^{C}}{\partial\q_{k}} (\Psibardot_{i} + \Phibardot_{i} )  + \I_{i}^{C} \bigg( \frac{\partial \Psibardot_{i}}{\partial \q_{k}}+\frac{\partial \Phibardot_{i}}{\partial \q_{k}} \bigg) \Bigg)
    \label{MSO_case2_1}
\end{equation}

Considering only the first term in Eq.~\ref{MSO_case2_1}, and using \ref{PhiT}:

\begin{equation}
     \frac{\partial  \Phibar_{j} \T   }{\partial \q_{k}}(2 \B_{i}^{C} \Phibar_{i} +  \I_{i}^{C} (\Psibardot_{i} + \Phibardot_{i} )) =   -\Phibar_{j} \T \Phibar_{k} \timesfM \big(2 \B_{i}^{C} \Phibar_{i} +  \I_{i}^{C} (\Psibardot_{i} + \Phibardot_{i} ) \big)
     \label{MSO_case2_2}
\end{equation}

Now, considering the next two terms in Eq.~\ref{MSO_case2_1} and using \ref{dphii} and \ref{Bic} as:

\begin{equation}
    \begin{aligned}
     \Phibar_{j}\T \Bigg(2 \frac{\partial \B_{i}^{C}}{\partial \q_{k}} \Phibar_{i} + 2\B_{i}^{C} \frac{\partial \Phibar_{i}}{\partial \q_{k}} \Bigg) =     \Phibar_{j}\T \big(2 \big(   \Bten{\I_{i}^{C}}{\Psibardot_{k}}+\Phibar_{k} \timesfM \B_{i}^{C}- \B_{i}^{C}(\Phibar_{k} \timesM) \big) \Phibar_{i} + 2\B_{i}^{C} \big(  \Phibar_{k}   \timesM \Phibar_{i}\big) \big)
    \end{aligned} 
\end{equation}

Cancelling terms lead to:

\begin{equation}
    \begin{aligned}
     \Phibar_{j}\T \Bigg(2 \frac{\partial \B_{i}^{C}}{\partial \q_{k}} \Phibar_{i} + 2\B_{i}^{C} \frac{\partial \Phibar_{i}}{\partial \q_{k}} \Bigg) =     2\Phibar_{j}\T \big(    \Bten{\I_{i}^{C}}{\Psibardot_{k}}+\Phibar_{k} \timesfM \B_{i}^{C}  \big)\Phibar_{i} 
    \end{aligned} 
     \label{MSO_case2_3}
\end{equation}

Considering the next term in Eq.~\ref{MSO_case2_1} and using the identity \ref{Iic} as:

\begin{equation}
    \Phibar_{j} \T  \frac{\partial \I_{i}^{C}}{\partial\q_{k}} (\Psibardot_{i} + \Phibardot_{i} )  =   \Phibar_{j} \T ( \Phibar_{k} \timesfM \I_{i}^{C}- \I_{i}^{C}(\Phibar_{k} \timesM) )  (\Psibardot_{i} + \Phibardot_{i} ) 
\end{equation}

Expanding the terms result into

\begin{equation}
    \begin{aligned}
      \Phibar_{j} \T \frac{\partial \I_{i}^{C}}{\partial\q_{k}} (\Psibardot_{i} + \Phibardot_{i} )  =    \Phibar_{j} \T \Phibar_{k} \timesfM \I_{i}^{C} \Psibardot_{i} - \Phibar_{j} \T\I_{i}^{C}(\Phibar_{k} \timesM) \Psibardot_{i} +  \Phibar_{j} \T \Phibar_{k} \timesfM \I_{i}^{C} \Phibardot_{i} -\Phibar_{j} \T \I_{i}^{C}(\Phibar_{k} \timesM) \Phibardot_{i}
    \end{aligned}
     \label{MSO_case2_4}
\end{equation}

Now considering the last term in Eq.~\ref{MSO_case2_1}, and using the identity \ref{dphii_dot} and \ref{psii_dot} as:

\begin{equation}
  \Phibar_{j} \T \I_{i}^{C} \bigg( \frac{\partial \Psibardot_{i}}{\partial \q_{k}}+\frac{\partial \Phibardot_{i}}{\partial \q_{k}} \bigg) =  \Phibar_{j} \T \I_{i}^{C} (\Psibardot_{k}   \timesM  \Phibar_{i} + \Phibar_{k} \timesM \Psibardot_{i}) + \Phibar_{j}\T \I_{i}^{C}( \Psibardot_{k}   \timesM \Phibar_{i}+ \Phibar_{k} \timesM \Phibardot_{i})
     \label{MSO_case2_5}
\end{equation}

Now adding the terms in Eq.~\ref{MSO_case2_2},~\ref{MSO_case2_3},~\ref{MSO_case2_4},~\ref{MSO_case2_5} results in:

\begin{equation}
    \begin{aligned}
        \frac{\partial^2 \taubar_{j}}{\partial \qd_{i} \partial \q_{k}} &= -\Phibar_{j} \T \Phibar_{k} \timesfM \big(2 \B_{i}^{C} \Phibar_{i} +  \I_{i}^{C} (\Psibardot_{i} + \Phibardot_{i} ) \big)+ 
          2\Phibar_{j}\T \big(    \Bten{\I_{i}^{C}}{\Psibardot_{k}}+\Phibar_{k} \timesfM \B_{i}^{C}  \big)\Phibar_{i} + \\&
        \Phibar_{j} \T \Phibar_{k} \timesfM \I_{i}^{C} \Psibardot_{i} - \Phibar_{j}\T \I_{i}^{C}(\Phibar_{k} \timesM) \Psibardot_{i} +  \Phibar_{j} \T \Phibar_{k} \timesfM \I_{i}^{C} \Phibardot_{i} -\Phibar_{j} \T \I_{i}^{C}(\Phibar_{k} \timesM) \Phibardot_{i}\\&
         +\Phibar_{j} \T \I_{i}^{C} (\Psibardot_{k}   \timesM  \Phibar_{i} + \Phibar_{k} \timesM \Psibardot_{i}) + \Phibar_{j}\T \I_{i}^{C}( \Psibardot_{k}   \timesM \Phibar_{i}+ \Phibar_{k} \timesM \Phibardot_{i})
    \end{aligned}
\end{equation}

Cancelling terms leads to:

\begin{equation}
    \frac{\partial^2 \taubar_{j}}{\partial \qd_{i} \partial \q_{k}} =    2\Phibar_{j}\T \big(   \Bten{\I_{i}^{C}}{\Psibardot_{k}}\Phibar_{i}  + \I_{i}^{C} (\Psibardot_{k}   \timesM  \Phibar_{i} ) \big)
\end{equation}

Using \ref{m26}

\begin{equation}
    \frac{\partial^2 \taubar_{j}}{\partial \qd_{i} \partial \q_{k}} =    2\Phibar_{j}\T \Big(   \big(\Bten{\I_{i}^{C}}{\Phibar_{i}}\Psibardot_{k}\big) \Rten -\I_{i}^{C} (\Psibardot_{k}   \timesM  \Phibar_{i} )   + \I_{i}^{C} (\Psibardot_{k}   \timesM  \Phibar_{i} ) \Big)
\end{equation}

Cancelling terms, the final expression is:

\begin{equation}
    \frac{\partial^2 \taubar_{j}}{\partial \qd_{i} \partial \q_{k}} =    \Phibar_{j}\T \Big[ 2\Bten{\I_{i}^C}{\Phibar_{i}}\Psibardot_{k}  \Big] \Rten  , (k \preceq j \prec i)
    \label{MSO_eqn_2}
\end{equation}

\subsection{$j \prec k \preceq i$, 1B, 2B}

\subsubsection{1B}
For this case, the partial derivative of Eq.~\ref{dtau_dqd_1} w.r.t $\q_{k}$ for the case $j \prec k \preceq i$ is taken as:

\begin{equation}
       \frac{\partial^2 \taubar_{i}}{\partial \qd_{j} \partial \q_{k}} = \frac{\partial  \Phibar_{i} \T   }{\partial \q_{k}}(2 \B_{i}^{C} \Phibar_{j} +  \I_{i}^{C} (\Psibardot_{j} + \Phibardot_{j} ))+\Phibar_{i}\T \Bigg(2 \frac{\partial \B_{i}^{C}}{\partial \q_{k}} \Phibar_{j}  + \frac{\partial \I_{i}^{C}}{\partial\q_{k}} (\Psibardot_{j} + \Phibardot_{j} )   \Bigg) 
\end{equation}

Using the identities \ref{Iic}, \ref{Bic}, and \ref{PhiT} as:

\begin{equation}
\begin{aligned}
      \frac{\partial^2 \taubar_{i}}{\partial \qd_{j} \partial \q_{k}} &=  -\Phibar_{i} \T \Phibar_{k} \timesfM \big(2 \B_{i}^{C} \Phibar_{j} +  \I_{i}^{C} (\Psibardot_{j} + \Phibardot_{j} ) \big)+  \Phibar_{i}\T \big(2 \big(   \Bten{\I_{i}^{C}}{\Psibardot_{k}}+\Phibar_{k} \timesfM \B_{i}^{C}- \B_{i}^{C}(\Phibar_{k} \timesM) \big) \Phibar_{j} \big) +\\&
       \Phibar_{i} \T ( \Phibar_{k} \timesfM \I_{i}^{C}- \I_{i}^{C}(\Phibar_{k} \timesM) )  (\Psibardot_{j} + \Phibardot_{j} ) 
      \end{aligned}
\end{equation}

Cancelling terms:

\begin{equation}
\begin{aligned}
      \frac{\partial^2 \taubar_{i}}{\partial \qd_{j} \partial \q_{k}} &=  \Phibar_{i}\T \big(2 \big(   \Bten{\I_{i}^{C}}{\Psibardot_{k}}- \B_{i}^{C}(\Phibar_{k} \timesM) \big) \Phibar_{j} \big)+ \Phibar_{i} \T ( - \I_{i}^{C}(\Phibar_{k} \timesM) )  (\Psibardot_{j} + \Phibardot_{j} ) 
      \end{aligned}
\end{equation}

Simplifying:

\begin{equation}
      \frac{\partial^2 \taubar_{i}}{\partial \qd_{j} \partial \q_{k}} =  \Phibar_{i}\T \big(2\Bten{\I_{i}^{C}}{\Psibardot_{k}}  \Phibar_{j} - 2\B_{i}^{C}(\Phibar_{k} \timesM)  \Phibar_{j}  - \I_{i}^{C}(\Phibar_{k} \timesM)   (\Psibardot_{j} + \Phibardot_{j} ) \big) 
\end{equation}

Switching the indices $j$ and $k$ to get the case $k \prec j \preceq i$, results in:

\begin{equation}
       \frac{\partial^2 \taubar_{i}}{\partial \qd_{k} \partial \q_{j}} =  \Phibar_{i}\T \big(2\Bten{\I_{i}^{C}}{\Psibardot_{j}}  \Phibar_{k} - 2\B_{i}^{C}(\Phibar_{j} \timesM)  \Phibar_{k}  - \I_{i}^{C}(\Phibar_{j} \timesM)   (\Psibardot_{k} + \Phibardot_{k} ) \big)
\end{equation}
Using \ref{m26} 

\begin{equation}
       \frac{\partial^2 \taubar_{i}}{\partial \qd_{k} \partial \q_{j}} =  \Phibar_{i}\T \Big(2(\Bten{\I_{i}^{C}}{\Phibar_{k}}\Psibardot_{j})\Rten -2\I_{i}^{C}(\Psibardot_{j}\timesM \Phibar_{k})  - 2\B_{i}^{C}(\Phibar_{j} \timesM)  \Phibar_{k}  - \I_{i}^{C}(\Phibar_{j} \timesM)   (\Psibardot_{k} + \Phibardot_{k} ) \Big)   
\end{equation}
Using \ref{m27} for the first term and simplifying
\begin{equation}
       \frac{\partial^2 \taubar_{i}}{\partial \qd_{k} \partial \q_{j}} =  -2[\Phibar_{k}\T (\Bten{\I_{i}^{C}}{\Phibar_{i}}\Psibardot_{j})\Rten]\Tten -2(\I_{i}^{C}\Phibar_{i})\T(\Psibardot_{j}\timesM \Phibar_{k})  - 2(\B_{i}^{C,T}\Phibar_{i})\T(\Phibar_{j} \timesM \Phibar_{k})  - (\I_{i}^{C}\Phibar_{i})\T (\Phibar_{j} \timesM)   (\Psibardot_{k} + \Phibardot_{k} )     
\end{equation}
Using \ref{m21} for the last three terms to  get:

\begin{equation}
    \begin{aligned}
       \frac{\partial^2 \taubar_{i}}{\partial \qd_{k} \partial \q_{j}} &=  -\big[\Phibar_{k}\T (2\Bten{\I_{i}^{C}}{\Phibar_{i}} \Psibardot_{j})\Rten \big]\Tten  +2\big[\Phibar_{k} \T ((\I_{i}^{C}\Phibar_{i})\crffM \Psibardot_{j}  )\Rten \big]\Tten   \\&~~~
       +\big[ \Phibar_{k}\T ((2\B_{i}^{C,T} \Phibar_{i})\crffM \Phibar_{j} )\Rten \big]\Tten  + \big[ (\Psibardot_{k} + \Phibardot_{k} )\T ((I_{i}^{C}\Phibar_{i}) \crffM \Phibar_{j} )\Rten  \big]\Tten  
  \end{aligned}
\end{equation}

Simplifying

\begin{equation}
    \begin{aligned}
       \frac{\partial^2 \taubar_{i}}{\partial \qd_{k} \partial \q_{j}} &=  \big[\Phibar_{k}\T (-2\Bten{\I_{i}^C}{\Phibar_{i}} \Psibardot_{j} + (2\B_{i}^{C\T} \Phibar_{i})\crffM \Phibar_{j}+2(\I_{i}^{C}\Phibar_{i})\crffM \Psibardot_{j}   )\Rten \\&~~~ +(\Psibardot_{k} + \Phibardot_{k} ) \T ((\I_{i}^{C}\Phibar_{i})\crffM \Phibar_{j}  )\Rten   
       \big]\Tten    ,(k \prec j \preceq i)
  \end{aligned}
      \label{MSO_eqn_3}
\end{equation}

\subsubsection{2B ($j \ne i$)}
\noindent In this case, we take the partial of Eq.~\ref{dtau_dqd_2} w.r.t $\q_{k}$. This case condition $j \prec k\preceq i$ satisfies the assumption $j \ne i$ for using Eq.~\ref{dtau_dqd_2}.

\begin{equation}
    \frac{\partial^2 \taubar_{j}}{\partial \qd_{i} \partial \q_{k}} =  \Phibar_{j} \T  \Bigg(2 \frac{\partial \B_{i}^{C}}{\partial \q_{k}} \Phibar_{i} + 2\B_{i}^{C} \frac{\partial \Phibar_{i}}{\partial \q_{k}} +  \frac{\partial \I_{i}^{C}}{\partial\q_{k}} (\Psibardot_{i} + \Phibardot_{i} )  + \I_{i}^{C} \bigg( \frac{\partial \Psibardot_{i}}{\partial \q_{k}}+\frac{\partial \Phibardot_{i}}{\partial \q_{k}} \bigg) \Bigg)
    \label{MSO_case4_1}
\end{equation}

Using the identities \ref{dphii}, \ref{dphii_dot}, \ref{psii_dot}, \ref{Iic}, and \ref{Bic} results in:

\begin{equation}
\begin{aligned}
    \frac{\partial^2 \taubar_{j}}{\partial \qd_{i} \partial \q_{k}} &=   \Phibar_{j}\T \big(2 \big(   \Bten{\I_{i}^{C}}{\Psibardot_{k}}+\Phibar_{k} \timesfM \B_{i}^{C}- \B_{i}^{C}(\Phibar_{k} \timesM) \big) \Phibar_{i} + \B_{i}^{C} \big(  \Phibar_{k}   \timesM \Phibar_{i}\big) \big)+\\&
        \Phibar_{j} \T \Phibar_{k} \timesfM \I_{i}^{C} \Psibardot_{i} - \Phibar_{j}\T \I_{i}^{C}(\Phibar_{k} \timesM) \Psibardot_{i} +  \Phibar_{j} \T \Phibar_{k} \timesfM \I_{i}^{C} \Phibardot_{i} -\Phibar_{j} \T \I_{i}^{C}(\Phibar_{k} \timesM) \Phibardot_{i}+\\&
         \Phibar_{j} \T \I_{i}^{C} (\Psibardot_{k}   \timesM  \Phibar_{i} + \Phibar_{k} \timesM \Psibardot_{i}) + \Phibar_{j}\T \I_{i}^{C}( \Psibardot_{k}   \timesM \Phibar_{i}+ \Phibar_{k} \timesM \Phibardot_{i})
    \end{aligned}
\end{equation}

Cancelling terms lead to:

\begin{equation}
\begin{aligned}
    \frac{\partial^2 \taubar_{j}}{\partial \qd_{i} \partial \q_{k}} &=   \Phibar_{j}\T \big(2 \big(   \Bten{\I_{i}^{C}}{\Psibardot_{k}}+\Phibar_{k} \timesfM \B_{i}^{C} \big) \Phibar_{i} \big)+
        \Phibar_{j} \T \Phibar_{k} \timesfM \I_{i}^{C} \Psibardot_{i}  +  \Phibar_{j} \T \Phibar_{k} \timesfM \I_{i}^{C} \Phibardot_{i}+ \\&
         \Phibar_{j} \T \I_{i}^{C} (\Psibardot_{k}   \timesM  \Phibar_{i} ) + \Phibar_{j}\T \I_{i}^{C}( \Psibardot_{k}   \timesM \Phibar_{i})
    \end{aligned}
\end{equation}

Collecting terms:

\begin{equation}
    \frac{\partial^2 \taubar_{j}}{\partial \qd_{i} \partial \q_{k}} =   2\Phibar_{j}\T \big(\Bten{\I_{i}^{C}}{\Psibardot_{k}}\Phibar_{i}+\Phibar_{k} \timesfM \B_{i}^{C}\Phibar_{i} +\I_{i}^{C}( \Psibardot_{k}   \timesM \Phibar_{i}) \big)+ \Phibar_{j} \T \Phibar_{k} \timesfM \I_{i}^{C} ( \Psibardot_{i}  + \Phibardot_{i})
\end{equation}

Switching indices $j$ and $k$ to get the case $k \prec j \preceq i$:

\begin{equation}
    \frac{\partial^2 \taubar_{k}}{\partial \qd_{i} \partial \q_{j}} =   2\Phibar_{k}\T \big(\Bten{\I_{i}^{C}}{\Psibardot_{j}}\Phibar_{i}+\Phibar_{j} \timesfM \B_{i}^{C}\Phibar_{i} +\I_{i}^{C}( \Psibardot_{j}   \timesM \Phibar_{i}) \big)+ \Phibar_{k} \T \Phibar_{j} \timesfM \I_{i}^{C} ( \Psibardot_{i}  + \Phibardot_{i})
\end{equation}
Using \ref{m26}

\begin{equation}
    \frac{\partial^2 \taubar_{k}}{\partial \qd_{i} \partial \q_{j}} =   2\Phibar_{k}\T \Big((\Bten{\I_{i}^{C}}{\Phibar_{i}}\Psibardot_{j})\Rten - \I_{i}^{C}(\Psibardot_{j}\timesM\Phibar_{i})+\Phibar_{j} \timesfM \B_{i}^{C}\Phibar_{i} +\I_{i}^{C}( \Psibardot_{j}   \timesM \Phibar_{i}) \Big)+ \Phibar_{k} \T \Phibar_{j} \timesfM \I_{i}^{C} ( \Psibardot_{i}  + \Phibardot_{i})
\end{equation}
Cancelling terms,

\begin{equation}
    \frac{\partial^2 \taubar_{k}}{\partial \qd_{i} \partial \q_{j}} =   2\Phibar_{k}\T \Big((\Bten{\I_{i}^{C}}{\Phibar_{i}}\Psibardot_{j})\Rten +\Phibar_{j} \timesfM \B_{i}^{C}\Phibar_{i} \Big)+ \Phibar_{k} \T \Phibar_{j} \timesfM \I_{i}^{C} ( \Psibardot_{i}  + \Phibardot_{i})
\end{equation}
Using \ref{m5} for the last two terms:
\begin{equation}
        \frac{\partial^2 \taubar_{k}}{\partial \qd_{i} \partial \q_{j}} =   \Phibar_{k}\T \Big[2\Bten{\I_{i}^{C}}{\Phibar_{i}}\Psibardot_{j} \Big]\Rten+\Phibar_{k}\T \Big[ \Big(2\B_{i}^{C}\Phibar_{i} +
  \I_{i}^{C} ( \Psibardot_{i}  + \Phibardot_{i}) \Big) \crffM \Phibar_{j} \Big]\Rten
\end{equation}

Simplifying:

\begin{equation}
    \begin{aligned}
        \frac{\partial^2 \taubar_{k}}{\partial \qd_{i} \partial \q_{j}} &=   \Phibar_{k}\T \Big[2\Bten{\I_{i}^C}{\Phibar_{i}}\Psibardot_{j}+   \big(2\B_{i}^{C}\Phibar_{i} +
  \I_{i}^{C} ( \Psibardot_{i}  + \Phibardot_{i}) \big) \crffM \Phibar_{j} \Big]\Rten, (k \prec j \preceq i) 
 \end{aligned}
 \label{MSO_eqn_4}
\end{equation}

\subsection{$j \preceq i \prec k$, 1C, 2C}

\subsubsection{1C}
\label{MSO_case_3_1C}
For this case, the partial derivative of $\frac{\partial \taubar_{i}}{\partial \qd_{j}}$ (Eq.~\ref{dtau_dqd_1}) w.r.t $\q_{k}$ is taken as:

\begin{equation}
     \frac{\partial^2 \taubar_{i}}{\partial \qd_{j} \partial \q_{k}} = \Phibar_{i}\T \Bigg(2 \frac{\partial \B_{i}^{C}}{\partial \q_{k}} \Phibar_{j} +  \frac{\partial \I_{i}^{C}}{\partial\q_{k}} (\Psibardot_{j} + \Phibardot_{j} )  \Bigg) 
\end{equation}

Using the identities \ref{Iic} and \ref{Bic} for the first and the second term respectively:

\begin{equation}
    \frac{\partial^2 \taubar_{i}}{\partial \qd_{j} \partial \q_{k}} = \Phibar_{i}\T \big( 2 \big(   \Bten{\I_{k}^{C}}{\Psibardot_{k}}+\Phibar_{k} \timesfM \B_{k}^{C}- \B_{k}^{C}(\Phibar_{k} \timesM) \big) \Phibar_{j} + (\Phibar_{k} \timesfM \I_{k}^{C}- \I_{k}^{C}(\Phibar_{k} \timesM) )  (\Psibardot_{j} + \Phibardot_{j} ) \big)
\end{equation}

First, switching the indices $j$ and $k$, and then $i$ and $j$ to get $k \preceq j \prec i$ to get:

\begin{equation}
      \frac{\partial^2 \taubar_{j}}{\partial \qd_{k} \partial \q_{i}} = \Phibar_{j}\T \Big( 2 \big(   \Bten{\I_{i}^{C}}{\Psibardot_{i}}+\Phibar_{i} \timesfM \B_{i}^{C}- \B_{i}^{C}(\Phibar_{i} \timesM) \big) \Phibar_{k} + (\Phibar_{i} \timesfM \I_{i}^{C}- \I_{i}^{C}(\Phibar_{i} \timesM) )  (\Psibardot_{k} + \Phibardot_{k} ) \Big), (k \preceq j \prec i)
            \label{MSO_eqn_5}
\end{equation}

\subsubsection{2C $(j \neq i)$ }
\noindent In this case, we take the partial derivative of $\frac{\partial \taubar_{j}}{\partial \qd_{i}}$ (Eq.~\ref{dtau_dqd_2}) w.r.t $\q_{k}$. Assumption for using Eq.~\ref{dtau_dqd_2} results in the condition $j \ne i$, which results in this case condition as $j \prec i \prec k$.

\begin{equation}
    \frac{\partial^2 \taubar_{j}}{\partial \qd_{i} \partial \q_{k}} =  \Phibar_{j} \T  \Bigg(2 \frac{\partial \B_{i}^{C}}{\partial \q_{k}} \Phibar_{i} + \frac{\partial \I_{i}^{C}}{\partial\q_{k}} (\Psibardot_{i} + \Phibardot_{i} )  \Bigg)
\end{equation}

Using identities \ref{Iic} and \ref{Bic} for the first and the second term as:

\begin{equation}
    \frac{\partial^2 \taubar_{j}}{\partial \qd_{i} \partial \q_{k}} =  \Phibar_{j}\T \big(2 \big(   \Bten{\I_{k}^{C}}{\Psibardot_{k}}+\Phibar_{k} \timesfM \B_{k}^{C}- \B_{k}^{C}(\Phibar_{k} \timesM) \big) \Phibar_{i} + ( \Phibar_{k} \timesfM \I_{k}^{C}- \I_{k}^{C}(\Phibar_{k} \timesM) )  (\Psibardot_{i} + \Phibardot_{i} )  \big)
\end{equation}

Like the previous case (1C), first switching the indices $j$ and $k$, and then $i$ and $j$ to get the case $k \prec j \prec i$ results in:

\begin{equation}
    \frac{\partial^2 \taubar_{k}}{\partial \qd_{j} \partial \q_{i}} =  \Phibar_{k}\T \Big(2 \big(   \Bten{\I_{i}^{C}}{\Psibardot_{i}}+\Phibar_{i} \timesfM \B_{i}^{C}- \B_{i}^{C}(\Phibar_{i} \timesM) \big) \Phibar_{j} + ( \Phibar_{i} \timesfM \I_{i}^{C}- \I_{i}^{C}(\Phibar_{i} \timesM) )  (\Psibardot_{j} + \Phibardot_{j} )  \Big), (k \prec j \prec i)
            \label{MSO_eqn_6}
\end{equation}

\pagebreak

\section{First Order Partial Derivative Of $\M(\q)$ w.r.t $\q$}
\label{MFO_sec}
Here, we present the cross second-order partial derivatives of ID w.r.t $\qdd$ and $\q$ that results in $\frac{\partial \M}{\partial \q}$. The lower-triangle of $\M(\q)$ for the case $j \preceq i$ is given as:

\begin{equation}
    \M_{ji} = \Phibar_{j}\T \IC{i} \Phibar_{i}
 \label{M_eqn}
\end{equation}

We take the partial derivative of $\M_{ji}$ w.r.t $\q_{k}$ by considering 3 cases pertaining to $k \preceq j \preceq i$, $j \prec k \preceq i$, and $j \preceq i \prec k$ as follows:

\subsection{$k \preceq j \preceq i$}
\subsubsection{1A}
For this case, we use the product chain rule to get:

\begin{equation}
    \frac{\partial \M_{ji}}{\partial \q_{k}} = \frac{\partial \Phibar_{j}\T }{\partial \q_{k}}\IC{i} \Phibar_{i} + \Phibar_{j}\T \left( \frac{\partial \IC{i} }{\partial \q_{k}}\Phibar_{i} +\IC{i}\frac{\partial \Phibar_{i}}{\partial \q_{k}} \right)
\end{equation}

Using identities \ref{dphii}, \ref{Iic} and \ref{PhiT}, we get:

\begin{equation}
     \frac{\partial \M_{ji}}{\partial \q_{k}} = -\Phibar_{j}\T \Phibar_{k} \timesfM \IC{i} \Phibar_{i} + \Phibar_{j}\T ( \Phibar_{k} \timesfM \I_{i}^{C}- \I_{i}^{C}(\Phibar_{k} \timesM))\Phibar_{i} + \Phibar_{j}\T \IC{i}\Phibar_{k} \timesM \Phibar_{i}
\end{equation}

Cancelling terms lead to:

\begin{equation}
     \frac{\partial \M_{ji}}{\partial \q_{k}}  = \mathbf{0}, (k \preceq j \preceq i)
     \label{M_FO_case_1A}
\end{equation}

\subsubsection{2A}

Due to symmetry in $\M(q)$, we get:
\begin{equation}
     \frac{\partial \M_{ij}}{\partial \q_{k}}  = \mathbf{0}, (k \preceq j \preceq i)
     \label{M_FO_case_2A}
\end{equation}

\subsection{$j \prec k \preceq i$ }
\subsubsection{1B}

For this case, we use the product rule of differentiation:

\begin{equation}
    \frac{\partial \M_{ji}}{\partial \q_{k}} =  \Phibar_{j}\T \left( \frac{\partial \IC{i} }{\partial \q_{k}}\Phibar_{i} +\IC{i}\frac{\partial \Phibar_{i}}{\partial \q_{k}} \right)
\end{equation}

Now using with identities \ref{Iic}, and \ref{dphii} as:

\begin{equation}
        \frac{\partial \M_{ji}}{\partial \q_{k}} =  \Phibar_{j}\T ( \Phibar_{k} \timesfM \I_{i}^{C}- \I_{i}^{C}(\Phibar_{k} \timesM))\Phibar_{i} + \Phibar_{j}\T \IC{i}\Phibar_{k} \timesM \Phibar_{i}
\end{equation}

Cancelling terms lead to:

\begin{equation}
        \frac{\partial \M_{ji}}{\partial \q_{k}} =  \Phibar_{j}\T ( \Phibar_{k} \timesfM \I_{i}^{C})\Phibar_{i}
\end{equation}
We switch the indices $k$ and $j$ to get the form $k \prec j \preceq i$:

\begin{equation}
       \frac{\partial \M_{ki}}{\partial \q_{j}} =  \Phibar_{k}\T ( \Phibar_{j} \timesfM \I_{i}^{C})\Phibar_{i} 
\end{equation}

using the property \ref{m5} :

\begin{equation}
       \frac{\partial \M_{ki}}{\partial \q_{j}} =  \Phibar_{k}\T ((\I_{i}^{C}\Phibar_{i}) \crffM \Phibar_{j})\Rten   ,  (k \prec j \preceq i)
     \label{M_FO_case_1B}
\end{equation}

\subsubsection{2B}
Using the symmetry for $\M(q)$, we get the expression for $\frac{\partial \M_{ik}}{\partial \q_{j}}$ as:

\begin{equation}
      \frac{\partial \M_{ik}}{\partial \q_{j}} = \left[\frac{\partial \M_{ki}}{\partial \q_{j}}  \right]\Tten ,  (k \prec j \preceq i)
     \label{M_FO_case_2B}
\end{equation}

\subsection{$j \preceq i \prec k$ }
\subsubsection{1C}

For this case we get:

\begin{equation}
        \frac{\partial \M_{ji}}{\partial \q_{k}} = \Phibar_{j}\T \left( \frac{\partial \IC{i} }{\partial \q_{k}} \right) \Phibar_{i}
\end{equation}

using identity \ref{Iic}, we get:

\begin{equation}
      \frac{\partial \M_{ji}}{\partial \q_{k}} = \Phibar_{j}\T ( \Phibar_{k} \timesfM \I_{k}^{C}- \I_{k}^{C}(\Phibar_{k} \timesM)) \Phibar_{i}
\end{equation}

Switching first the indices $k$ and $j$, and the indices $j$ and $i$ to get the case $k \preceq  j \prec i$ results in the term:

\begin{equation}
      \frac{\partial \M_{kj}}{\partial \q_{i}} = \Phibar_{k}\T ( \Phibar_{i} \timesfM \I_{i}^{C}- \I_{i}^{C}(\Phibar_{i} \timesM)) \Phibar_{j}, (k \preceq  j \prec i)
      \label{M_FO_case_1C}
\end{equation}

\subsubsection{2C}
Using the symmetry for $\M(q)$, we get the expression for $\frac{\partial \M_{jk}}{\partial \q_{i}} $ as:

\begin{equation}
     \frac{\partial \M_{jk}}{\partial \q_{i}} = \left[ \frac{\partial \M_{kj}}{\partial \q_{i}} \right] \Tten , (k \preceq  j \prec i)
           \label{M_FO_case_2C}
\end{equation}

\pagebreak

\section{Efficient Implementation}
\label{eff_impl_sec}
For implementation simplicity, the expressions for SO partials are reduced from tensor to matrix and vector form. Each of the expression listed in Sec.~\ref{SO_q_sec}, \ref{SO_qd_sec}, and \ref{MSO_sec} are written in a form to avoid tensor operations. For this purpose, they are evaluated only for one DoF ($p^{th}$ DoF of of joint $i$, $t^{th}$ DoF of joint $j$, and $r^{th}$ DoF of joint $k$) at a time. The spatial matrix operators $\timesM$, $\timesfM$, and $\crffM$ automatically reduce to spatial vector operators $\times$, $\timesf$, and $\crff$ respectively.
\begin{enumerate}
    \item [1.] SO Partials w.r.t $\q$

\begin{enumerate}
    \item [1A)] Considering Eq.~\ref{tau_SO_q_1A}, and evaluating it for only the $p^{th}$ DoF of joint $i$, $t^{th}$ DoF of joint $j$, and $r^{th}$ DoF of joint $k$ results into:
    
\begin{equation}
\begin{aligned}
     \frac{\partial^2 \taubar_{i}}{\partial \q_{j} \partial \q_{k}} &=  -\Big[\psibardot_{j,t}\T [2\Bmat{\I_i^{C}}{\phibar_{i,p}}\psibardot_{k,r}]\Rten + 2\phibar_{j,t}\T ((\B_{i}^{C\T} \phibar_{i,p}) \crffM\psibardot_{k,r} )\Rten    +  \phibar_{j,t}\T ((\I_{i}^{C}\phibar_{i,p} ) \crffM\psibarddot_{k,r} )\Rten \Big]\Tten  , (k \preceq j \preceq i)
     \end{aligned}
\end{equation}

where $\phibar_{i,p}$ is a column vector of $\Phibar_i$, $\phibar_{j,t}$ is the $t^{th}$ column vector of $\Phibar_{j}$, and $\psibarddot_{k,r}$ is the $r^{th}$ column vector of $\Psibarddot_{k}$. The tensor $\Bten{\I_{i}^C}{\Phibar_{i}}$ takes the vector $\phibar_{i,p}$ as an argument and reduces to a matrix. Since the terms inside the 2-3 3D rotation ($\Rten$) are vectors, the 3D 2-3 rotation is no longer needed and can be safely removed. The 3D 1-2 rotation ($\Tten$) falls out, since it operates on a scalar, resulting in:

\begin{equation}
\begin{aligned}
   \frac{\partial^2 \taubar_{i,p}}{\partial \q_{j,t} \partial \q_{k,r}} =  -\psibardot_{j,t}\T 2\Bmat{\I_i^{C}}{\phibar_{i,p}}\psibardot_{k,r}  - 2\phibar_{j,t}\T ((\B_{i}^{C\T} \phibar_{i,p}) \crff \psibardot_{k,r} )   -  \phibar_{j,t}\T ((\I_{i}^{C}\phibar_{i,p} ) \crff \psibarddot_{k,r} )  , (k \preceq j \preceq i)
     \end{aligned}
\end{equation}

 \item [2A)] Considering Eq.~\ref{tau_SO_q_2A} for only the $p^{th}$ DoF of joint $i$, $t^{th}$ DoF of joint $j$, and $r^{th}$ DoF of joint $k$ results in:
 
 \begin{equation}
      \frac{\partial^2 \taubar_{i,p}}{\partial \q_{k,r} \partial \q_{j,t}} =  \Bigg[\frac{\partial^2 \taubar_{i,p}}{\partial \q_{j,t} \partial \q_{k,r}} \Bigg]\Rten
\end{equation}

Since the quantity $\frac{\partial^2 \taubar_{i,p}}{\partial \q_{j,t} \partial \q_{k,r}} $ is a scalar, the 3D 2-3 rotation ($\Rten$) falls out:

\begin{equation}
 \frac{\partial^2 \taubar_{i,p}}{\partial \q_{k,r} \partial \q_{j,t}} =  \frac{\partial^2 \taubar_{i,p}}{\partial \q_{j,t} \partial \q_{k,r}}  , (k \preceq j \preceq i)
\end{equation}

\item [1B)] Eq.~\ref{tau_SO_q_1B} is evaluated for  $p^{th}$ DoF of joint $i$, $t^{th}$ DoF of joint $j$, and $r^{th}$ DoF of joint $k$. 
The 2-3 3D rotation ($\Rten$) on the tensorial terms can be removed due to the resulting expression being a scalar:

\begin{equation}
    \begin{aligned}
         \frac{\partial^2 \taubar_{k,r}}{\partial \q_{i,p} \partial \q_{j,t}} &= \phibar_{k,r}\T \Big( 2(\Bmat{\I_{i}^C}{\psibardot_{i,p}}+ {\phibar_{i,p} \timesf \B_{i}^{C} - \B_{i}^{C}\phibar_{i,p}  \times})\psibardot_{j,t} + {\big( \phibar_{i,p} \timesf \I_{i}^{C}   -\I_{i}^{C}   \phibar_{i,p}\times \big)}\psibarddot_{j,t}  +\\&~~~~
          \phibar_{j,t} \timesf \big( 2\B_{i}^{C}  \psibardot_{i,p} +\I_{i}^{C} \psibarddot_{i,p}+\f_{i}^{C} \crff \phibar_{i,p}\big) \Big)   , (k \prec j \preceq i) 
    \end{aligned}
\end{equation}

\item [2B)] Eq.~\ref{tau_SO_q_2B} is considered only for  $p^{th}$ DoF of joint $i$, $t^{th}$ DoF of joint $j$, and $r^{th}$ DoF of joint $k$ as:

\begin{equation}
     \frac{\partial^2 \taubar_{k,r}}{\partial \q_{j,t} \partial \q_{i,p}} =  \Bigg[\frac{\partial^2 \taubar_{k,r}}{\partial \q_{i,p} \partial \q_{j,t}} \Bigg]\Rten
\end{equation}

In this case, since the quantity $\frac{\partial^2 \taubar_{k,r}}{\partial \q_{i,p} \partial \q_{j,t}}$ is a scalar, the 3D 2-3 rotation has no affect on it.
\begin{equation}
     \frac{\partial^2 \taubar_{k,r}}{\partial \q_{j,t} \partial \q_{i,p}} =  \frac{\partial^2 \taubar_{k,t}}{\partial \q_{i,p} \partial \q_{j,t}}   , (k \prec j \prec i)
\end{equation}

\item [1C)]

Considering Eq.~\ref{tau_SO_q_1C} for the  $p^{th}$ DoF of joint $i$, $t^{th}$ DoF of joint $j$, and $r^{th}$ DoF of joint $k$ results into:

\begin{equation}
  \frac{\partial^2 \taubar_{j,t}}{\partial \q_{k,r} \partial \q_{i,p}} = 2\phibar_{j,t} \T \Big(   \Bmat{\I_{i}^C}{\psibardot_{i,p}}+{\phibar_{i,p} \timesf \B_{i}^{C}- \B_{i}^{C}\phibar_{i,p} \times } \Big)\psibardot_{k,r} + 
  \phibar_{j,t} \T  \Big( {\phibar_{i,p} \timesf \I_{i}^{C}- \I_{i}^{C}(\phibar_{i,p} \times)}\Big) \psibarddot_{k,r}   , (k \preceq j \prec i)
\end{equation}

\item [2C)] For this case, Eq.~\ref{tau_SO_q_2C} is considered for  $p^{th}$ DoF of joint $i$, $t^{th}$ DoF of joint $j$, and $r^{th}$ DoF of joint $k$ as:

\begin{equation}
     \frac{\partial^2 \taubar_{j,t}}{\partial \q_{i,p} \partial \q_{k,r}} =  \Bigg[  \frac{\partial^2 \taubar_{j,t}}{\partial \q_{k,r} \partial \q_{i,p}} \Bigg]\Rten
\end{equation}

Since the quantity $\frac{\partial^2 \taubar_{j,t}}{\partial \q_{k,r} \partial \q_{i,p}}$ is a scalar, the 3D 2-3 rotation doesn't affect it, resulting into:

\begin{equation}
     \frac{\partial^2 \taubar_{j,t}}{\partial \q_{i,p} \partial \q_{k,r}} =   \frac{\partial^2 \taubar_{j,t}}{\partial \q_{k,r} \partial \q_{i,p}}  , (k \preceq j \prec i)
\end{equation}

\end{enumerate}

    \item [2.] SO Partials w.r.t $\qd$ \\\\
    All of the cases in Sec.~\ref{SO_qd_sec} are considered for only the  $p^{th}$ DoF of joint $i$, $t^{th}$ DoF of joint $j$, and $r^{th}$ DoF of joint $k$. The 3D 2-3 tensor rotation ($\Rten$) is dropped in the expressions due to reduction to scalar form. The 3D 1-2 tensor rotation ($\Tten$) is also dropped due to reduction to the scalar form.
    
    \hfill
    \hfill

    \begin{enumerate}
    
        \item [1A)]  Eq.~\ref{SO_qd_case_1A} for $p^{th}$ DoF of joint $i$, $t^{th}$ DoF of joint $j$, and $r^{th}$ DoF of joint $k$ becomes:
        
\begin{equation}
    \frac{\partial^2 \taubar_{i,p}}{\partial \qd_{j,t} \partial \qd_{k,r}} =-\phibar_{j,t}\T \Big[2\Bmat{\I_i^{C}}{\phibar_{i,p}}\phibar_{k,r} \Big] , (k \prec j \preceq i)
\end{equation} 
        
        \item [2A)] Eq.~\ref{SO_qd_case_2A} for  $p^{th}$ DoF of joint $i$, $t^{th}$ DoF of joint $j$, and $r^{th}$ DoF of joint $k$ becomes:
        
        \begin{equation}
    \frac{\partial^2 \taubar_{i,p}}{\partial \qd_{k,r} \partial \qd_{j,t}} =  \frac{\partial^2 \taubar_{i,p}}{\partial \qd_{j,t} \partial \qd_{k,r}} , (k \prec j \preceq i)
    \end{equation}
    
      \item [B)] Eq.~\ref{SO_qd_case_B} becomes:
   
\begin{equation}
    \frac{\partial^2 \taubar_{i,p}}{\partial \qd_{j,t} \partial \qd_{k,r}} =  -\phibar_{j,t}\T \big[{\big( \phibar_{i,p} \timesf \I_{i}^{C}   -\I_{i}^{C}   \phibar_{i,p}\times \big)}\phibar_{k,r} \big] , (k = j \preceq i) 
\end{equation}

        \item [1C)] Eq.~\ref{SO_qd_case_1C} becomes:
        
    \begin{equation}
      \frac{\partial^2 \taubar_{k,r}}{\partial \qd_{i,p} \partial \qd_{j,t}} =  \phibar_{k,r}\T \Big[2\Bmat{\I_i^{C}}{\phibar_{i,p}}\phibar_{j,t} \Big], (k \prec j \prec i) 
    \end{equation}
                
        \item [2C)] Eq.~\ref{SO_qd_case_2C} becomes:
        
\begin{equation}
     \frac{\partial^2 \taubar_{k,r}}{\partial \qd_{j,t} \partial \qd_{i,p}} =  \frac{\partial^2 \taubar_{k,r}}{\partial \qd_{i,p} \partial \qd_{j,t}} , (k \prec j \prec i) 
\end{equation}

        \item [D)] Eq.~\ref{SO_qd_case_D} becomes:
        
  \begin{equation}
  \frac{\partial^2 \taubar_{k,r}}{\partial \qd_{i,p} \partial \qd_{j,t}} =  \phibar_{k,r}\T \Big[( \phibar_{i,p}\timesf \I_{i}^{C} + (\I_{i}^{C}\phibar_{i,p})\crff )\phibar_{j,t}  \Big]  , (k \prec j = i)
\end{equation}

        \item [1E)] Eq.~\ref{SO_qd_case_1E} becomes:
        
\begin{equation}
     \frac{\partial^2 \taubar_{j,t}}{\partial \qd_{k,r} \partial \qd_{i,p}} = \phibar_{j,t}\T \Big[2\Bmat{\I_i^{C}}{\phibar_{i,p}} \phibar_{k,r}\Big], (k \preceq j \prec i)
\end{equation}

        \item [2E)]Eq.~\ref{SO_qd_case_2E} becomes:
        
\begin{equation}
     \frac{\partial^2 \taubar_{j,t}}{\partial \qd_{i,p} \partial \qd_{k,r}} =   \frac{\partial^2 \taubar_{j,t}}{\partial \qd_{k,r} \partial \qd_{i,p}}, (k \preceq j \prec i)  
\end{equation}

         Here, the 2-3 rotation is dropped since the term $\frac{\partial^2 \taubar_{j,t}}{\partial \qd_{k,r} \partial \qd_{i,p}} $ is a scalar, and the 2-3 rotation has no effect on it.
    \end{enumerate}

   \vspace{5ex}
    \item [3.] Cross SO Partials w.r.t $\q$ and $\qd$: \\
    
    Similar to previous case, expressions listed in Sec.~\ref{MSO_sec} are converted to the matrix form by considering only the $p^{th}$ DoF of joint $i$, $t^{th}$ DoF of joint $j$, and $r^{th}$ DoF of joint $k$. This reduction results in dropping out of the tensor 2-3 rotation in all the cases.
    
    \begin{enumerate}
    
        \item [1A)] Eq.~\ref{MSO_eqn_1} becomes:
        
\begin{equation}
    \frac{\partial^2 \taubar_{i,p}}{ \partial \qd_{j,t}\partial \q_{k,r}} =    -\phibar_{j,t}\T \big( 2\Bmat{\I_i^{C}}{\phibar_{i,p}} \psibardot_{k,r}\big) , (k \preceq j \prec i)
\end{equation}
    
         \item [2A)] Eq.~\ref{MSO_eqn_2} becomes:
         
\begin{equation}
    \frac{\partial^2 \taubar_{j,t}}{\partial \qd_{i,p} \partial \q_{k,r}} =  \phibar_{j,t}\T \Big[ 2\Bmat{\I_i^{C}}{\phibar_{i,p}}\psibardot_{k,r}  \Big]  , (k \preceq j \prec i)
\end{equation}
        
         \item [1B)] Eq.~\ref{MSO_eqn_3} becomes:

\begin{equation}
    \begin{aligned}
       \frac{\partial^2 \taubar_{i,p}}{\partial \qd_{k,r} \partial \q_{j,t}} &=  \phibar_{k,r}\T \big(-2\Bmat{\I_i^{C}}{\phibar_{i,p}} \psibardot_{j,t}   + 2(\I_{i}^{C}\phibar_{i,p})\crff \psibardot_{j,t} + (2\B_{i}^{C\T} \phibar_{i,p})\crff \phibar_{j,t}) \big)   \\&~~~
        +  (\psibardot_{k,r} + \phibardot_{k,r} )\T (I_{i}^{C}\phibar_{i,p}) \crff \phibar_{j,t}  ,(k \prec j \preceq i)    
  \end{aligned}
\end{equation}
    
         \item [2B)] Eq.~\ref{MSO_eqn_4} becomes:
         
\begin{equation}
      \frac{\partial^2 \taubar_{k,r}}{\partial \qd_{i,p} \partial \q_{j,t}} =   \phibar_{k,r}\T \Big[2\Bmat{\I_i^{C}}{\phibar_{i,p}}\psibardot_{j,t} + \big(2\B_{i}^{C}\phibar_{i,p} +
  \I_{i}^{C} ( \psibardot_{i,p}  + \phibardot_{i,p}) \big) \crff \phibar_{j,t} \Big]  , (k \prec j \preceq i) 
\end{equation}
         
     \item [1C)] Eq.~\ref{MSO_eqn_5} becomes:
\begin{equation}
    \begin{aligned}
      \frac{\partial^2 \taubar_{j,t}}{\partial \qd_{k,r} \partial \q_{i,p}} &=  \phibar_{j,t}\T \Big( 2 \big(   \Bmat{\I_{i}^C}{\psibardot_{i,p}}+{\phibar_{i,p} \timesf \B_{i}^{C}- \B_{i}^{C}\phibar_{i,p} \times} \big) \phibar_{k,r} + (\phibar_{i,p} \timesf \I_{i}^{C}- \I_{i}^{C}(\phibar_{i,p} \times) )  (\psibardot_{k,r} + \phibardot_{k,r} ) \Big) \\&~~~~~~~~~~~~~~~~~~~~~~~~~~~~~~~ , (k \preceq j \prec i)
      \end{aligned}
\end{equation}

         \item [2C)]   Eq.~\ref{MSO_eqn_6} becomes:
         
\begin{equation}
\begin{aligned}
      \frac{\partial^2 \taubar_{k,r}}{\partial \qd_{j,t} \partial \q_{i,p}} &= \phibar_{k,r}\T \Big(2 \big(   \Bmat{\I_{i}^C}{\psibardot_{i,p}}+{\phibar_{i,p} \timesf \B_{i}^{C}- \B_{i}^{C}\phibar_{i,p} \times} \big) \phibar_{j,t} + {( \phibar_{i,p} \timesf \I_{i}^{C}- \I_{i}^{C}(\phibar_{i,p} \times) )}  (\psibardot_{j,t} + \phibardot_{j,t} )  \Big) \\
      &~~~~~~~~~~~~~~~~~~~~~~~~~~~~~~~, (k \prec j \prec i)
      \end{aligned}
\end{equation}

\item [4.] FO Partials of $\M(\q) $w.r.t $\q$: All the expressions for FO partial of $\M(\q)$ are converted to scalar forms by considering them only for the $p^{th}$ DoF of joint $i$, $t^{th}$ DoF of joint $j$, and $r^{th}$ DoF of joint $k$.\\

\begin{enumerate}
    \item [1A)] Eq.~\ref{M_FO_case_1A} becomes
    
    \begin{equation}
     \frac{\partial \M_{(j,t) (i,p)}}{\partial \q_{k,r}}  = \mathbf{0},  (k \preceq j \preceq i)
\end{equation}

        \item [2A)]  Eq.~\ref{M_FO_case_2A} becomes
\begin{equation}
     \frac{\partial \M_{(i,p)(j,t)}}{\partial \q_{k,r}}  =  \mathbf{0},  (k \preceq j \preceq i)
\end{equation}

    \item [1B)] Eq.~\ref{M_FO_case_1B} becomes:
    \begin{equation}
       \frac{\partial \M_{(k,r)(i,p)}}{\partial \q_{j,t}} =  \phibar_{k,r}\T ((\I_{i}^{C}\phibar_{i,p}) \crff \phibar_{j,t})  ,  (k \prec j \preceq i)
\end{equation}
    
    \item [2B)] Eq.~\ref{M_FO_case_2B} becomes
    
    \begin{equation}
      \frac{\partial \M_{(i,p)(k,r)}}{\partial \q_{j,t}} = \frac{\partial \M_{(k,r)(i,p)}}{\partial \q_{j,t}}  ,  (k \prec j \preceq i)
\end{equation}

The tensor 1-2 rotation falls out in this case, since the quantity $\frac{\partial \M_{(k,r)(i,p)}}{\partial \q_{j,t}}  $ is a scalar, and 1-2 tensor rotation has no effect on it.

    \item [1C)] Eq.~\ref{M_FO_case_1C} becomes:
    
    \begin{equation}
      \frac{\partial \M_{(k,r)(j,t)}}{\partial \q_{i,p}} = \phibar_{k,r}\T ( \phibar_{i,p} \timesf \I_{i}^{C}- \I_{i}^{C}(\phibar_{i,p} \times)) \phibar_{j,t}, (k \preceq  j \prec i)
\end{equation}

    \item [2C)] Eq.~\ref{M_FO_case_2C} becomes:

\begin{equation}
     \frac{\partial \M_{(j,t)(k,r)}}{\partial \q_{i,p}} = \frac{\partial \M_{(k,r)(j,t)}}{\partial \q_{i,p}} , (k \preceq  j \prec i)
\end{equation}

In this case, since $\frac{\partial \M_{(k,r)(j,t)}}{\partial \q_{i,p}} $ is a scalar, the 1-2 tensor rotation falls off.

\end{enumerate}

     \end{enumerate}
   
\end{enumerate}

\newpage

\section{Second-Order relation between ID and FD}
\label{sec_SO_eqn}
In this section, the second-order relation between the partial derivatives of ID and FD is derived. We start with the formulation in Eq~\ref{jain_FO_eqn}~\cite{jain1993linearization}.

  \begin{equation}
      \frac{\partial~\textrm{\textrm{FD}}}{\partial \boldsymbol u}\biggr\rvert_{\q_{0},  \qd_{0}, \taubar_{0}} = -\M^{-1}(\q) \frac{\partial~ \textrm{ID} }{\partial \u}\biggr\rvert_{\q_{0},  \qd_{0},  \qdd_{0}}
      \label{jain_FO_eqn}
  \end{equation}
Re-arranging the equation:
 \begin{equation}
     \frac{\partial~  \textrm{ID}(model, \q, \qd, \textrm{FD}) }{\partial \u}\biggr\rvert_{\q, \qd_{0}, \qdd_{0}} + \frac{\partial~ \textrm{ID}(model, \q, \qd, \textrm{FD}) }{\partial~\textrm{FD}} \biggr\rvert_{\q_{0}, \qd_{0}, \qdd_{0}}  \frac{\partial~\textrm{FD}}{\partial \u}\biggr\rvert_{\q_{0}, \qd_{0}, \boldsymbol{\tau}_{0}} = 0
    \label{eqn_FD_SO_1}
\end{equation}
where $\frac{\partial \textrm{ID}}{\partial \textrm{FD}}=\M$.
Now, we take another partial derivative of Eq.~\ref{eqn_FD_SO_1} w.r.t $\w$.
 \begin{equation}
     \frac{\partial }{\partial \w} \Bigg( \frac{\partial~  \textrm{ID}(model, \q, \qd, \textrm{FD}) }{\partial \u}\biggr\rvert_{\q_{0}, \qd_{0}, \qdd_{0}} \Bigg ) + \frac{\partial }{\partial \w}\Bigg ( \frac{\partial~ \textrm{ID}(model, \q, \qd, \textrm{FD}) }{\partial~\textrm{FD}} \biggr\rvert_{\q_{0}, \qd_{0}, \qdd_{0}}  \frac{\partial~\textrm{FD}}{\partial \u}\biggr\rvert_{\q_{0}, \qd_{0}, \taubar_{0}} \Bigg)= 0
    \label{eqn_FD_SO_2}
\end{equation}
Moving forward, we lose the terms $(model, \q, \qd, \textrm{FD})$ for ease of reading (Eq~\ref{eqn_FD_SO_3}). We use the chain rule for differentiation again to get Eq~\ref{eqn_FD_SO_4}.
\begin{equation}
   \frac{\partial }{\partial \w} \Bigg(  \frac{\partial \textrm{ID}}{\partial \u} \Bigg) +   \frac{\partial }{\partial \w} \Bigg( \frac{\partial \textrm{ID}}{\partial \textrm{FD}} \frac{\partial \textrm{FD}}{\partial \u}  \Bigg ) = 0
       \label{eqn_FD_SO_3}
\end{equation}
\small
\begin{equation}
   \frac{\partial }{\partial \w} \Bigg(  \frac{\partial \textrm{ID}}{\partial \u} \Bigg) +     \Bigg( \frac{\partial^2 \textrm{ID}}{\partial \u \partial \textrm{FD}}  \Bigg ) \frac{\partial \textrm{FD}}{\partial \w}+ \Bigg ( \frac{\partial }{\partial \w} \Bigg( \frac{\partial \textrm{ID}}{\partial \textrm{FD}}  \Bigg) +  \Bigg( \frac{\partial^2 \textrm{ID}}{\partial \textrm{FD}^2} \Bigg )  \frac{\partial \textrm{FD}}{\partial \w} \Bigg) \frac{\partial \textrm{FD}}{\partial \u}  +
    \frac{\partial \textrm{ID}}{\partial \textrm{FD}} \Bigg ( \frac{\partial }{\partial \w} \Bigg(  \frac{\partial \textrm{FD}}{\partial \u} \Bigg) + \frac{\partial}{ \partial \textrm{FD}} \Bigg( \frac{\partial \textrm{FD}}{\partial \u} \Bigg)\frac{\partial \textrm{FD}}{\partial \w} \Bigg)= 0
    \label{eqn_FD_SO_4}
\end{equation}
The last term $\frac{\partial}{\textrm{FD}} \Big( \frac{\partial \textrm{FD}}{\partial \u} \Big)=0$, since $\frac{\partial \textrm{FD}}{\partial \u}$ does not have a dependency on FD. Simplifying:
\begin{equation}
  \frac{\partial^2 \textrm{ID}}{\partial \u \partial \w}  +     \Bigg( \frac{\partial^2 \textrm{ID}}{\partial \u \partial \textrm{FD}}  \Bigg ) \frac{\partial \textrm{FD}}{\partial \w}+ \Bigg ( \frac{\partial }{\partial \w} \Bigg( \frac{\partial \textrm{ID}}{\partial \textrm{FD}}  \Bigg) +  \Bigg( \frac{\partial^2 \textrm{ID}}{\partial \textrm{FD}^2} \Bigg )  \frac{\partial \textrm{FD}}{\partial \w} \Bigg) \frac{\partial \textrm{FD}}{\partial \u}  +
    \frac{\partial \textrm{ID}}{\partial \textrm{FD}} \Bigg ( \frac{\partial }{\partial \w} \Bigg(  \frac{\partial \textrm{FD}}{\partial \u} \Bigg)  \Bigg)= 0
    \label{eqn_FD_SO_41}
\end{equation}


Noting that $\frac{\partial \textrm{ID}}{\partial \textrm{FD}}=\M$, and $\frac{\partial^2 \textrm{ID}}{\partial \textrm{FD}^2}=0$, we get:
\begin{equation}
  \frac{\partial^2 \textrm{ID}}{\partial \u \partial \w}  +     \Bigg( \frac{\partial^2 \textrm{ID}}{\partial \u \partial \textrm{FD}}  \Bigg ) \frac{\partial \textrm{FD}}{\partial \w}+ \frac{\partial \M}{\partial \w}  \frac{\partial \textrm{FD}}{\partial \u}  +
    \M   \frac{\partial^2 \textrm{FD}}{\partial \u \partial \w} = 0
    \label{eqn_FD_SO_42}
\end{equation}
Simplifying, the second term:
\begin{equation}
  \frac{\partial^2 \textrm{ID}}{\partial \u \partial \w}  +     \Bigg( \frac{\partial \M}{ \partial \u}  \Bigg )\Rten \frac{\partial \textrm{FD}}{\partial \w}+ \frac{\partial \M}{\partial \w}  \frac{\partial \textrm{FD}}{\partial \u}  +
    \M   \frac{\partial^2 \textrm{FD}}{\partial \u \partial \w} = 0
    \label{eqn_FD_SO_42}
\end{equation}
The final form of the equation is:

\begin{equation}
  \frac{\partial^{2} \textrm{FD}}{\partial \u \partial \w}   = -\M^{-1} \left(\frac{\partial^{2} \textrm{ID}}{\partial \u \partial \w} +\frac{\partial \M}{\partial \w}  \frac{\partial \textrm{FD}}{\partial \u} +\Bigg(  \frac{\partial \M}{\partial \u}  \frac{\partial \textrm{FD}}{\partial \w}\Bigg)\Rten\right)
    \label{eqn_FD_SO_43}
\end{equation}

The 2-3 rotation $(\Rten)$ on the last term here is for keeping the tensor order storage consistent with the rest of the terms in the equation. For example, for the term $\frac{\partial^{2} \textrm{FD}}{\partial \u \partial \w}$, the elements of $\textrm{FD}$ are along the rows, $\u$ along the columns, and $\w$ along the pages. For the term $\frac{\partial \M}{\partial \u}  \frac{\partial \textrm{FD}}{\partial \w}$ on the other hand leads to the elements of $\u$ along the pages. Hence to keep the order intact for all the terms, this term is rotated in the 2-3 dimension.

\newpage
\section{Extension of ID FO/SO Derivatives with External Forces}
\label{sec_ID_FD_contacts}
In this section, we provide the extensions of Inverse Dynamics (ID) partial derivatives for rigid-body systems with external forces. The ``constrained Inverse Dynamics'' is given by:
\begin{equation}
\begin{aligned}
     \S\taubar &= \M(\q)\qdd+ \b-\Jc\T \lambda \\
   &=  \textrm{ID}_c({\rm model},\q,  \qd, \qdd,\lambda)
     \end{aligned}
     \label{eqn_IDc}
\end{equation}

where $\Jc \in \R^{n_{c}\times n}$ is the contact Jacobian, $\lambda \in \R^{n_{c}}$ is the vector of the contact forces, and $\mathbf{S}$ is the selector matrix to map the actuated joint torques to the entire tree. The Coriolis term $\C\qd$ and the generalized gravitational forces $\g$ are combined together in $\b=\C\qd+\g$. Using a similar approach used for getting the ID FO derivatives~\cite{singh2022efficient}, we consider the joint torque for the joint $i$ for the constrained-ID~\eqref{eqn_IDc} as:
\begin{equation}
    [\S\taubar]_{i} =   [ \M(\q) \qdd]_{i}+   [\C(\q,\qd) \qd]_{i}  + g_{i}(\q) - [\taubar_{ext}]_{i}
    \label{eqn_ID_con}
\end{equation}
 where $\taubar_{ext}=\Jc\T \lambda$ from \eqref{eqn_IDc}. From the RNEA~\cite{Featherstone08}, the  torque for joint $i$ is given via the spatial force across joint $i$:
\begin{equation}
    [\S\taubar]_{i} = \Phibar_{i}\T \sum_{l \succeq i}  ( \I_{l} \a_{l} + \v_{l} \times^{*} \I_{l} \v_{l} -\f_{l,ext}) 
        \label{eqn_RNEA_ext}
\end{equation}
where $\f_{l,ext}$ is the external spatial force on body $i$. Comparing~\eqref{eqn_ID_con} and \eqref{eqn_RNEA_ext},

\begin{equation}
    [\taubar_{ext}]_{i} =\Phibar_{i}\T \sum_{l \succeq i}\f_{l,ext}
    \label{eqn_tau_ext}
\end{equation}

\subsection{Constrained ID FO Derivatives}
\label{sec:IDc_FO}

Similar to the approach for ID FO derivatives~\cite{singh2022efficient}, we first take the partial derivative of Eq.~\eqref{eqn_tau_ext} w.r.t. $\q_{j}$ for the condition $j \preceq i$. Using the identity \ref{J3}, and chain-rule product:

\begin{equation}
    \frac{\partial [\taubar_{ext}]_{i}}{\partial \q_{j}} =-\Phibar_{i}\T (\f_{i,ext}^{C} \crff  \Phibar_{j})+ \Phibar_{i}\T\frac{\partial \f_{i,ext}^{C}}{\partial \q_{j}}, (j \preceq i)
   \label{JTlam_i_term_1}
\end{equation}
where, $\f_{i,ext}^{C}=\sum_{l \succeq i}\f_{l,ext}$ is the cumulative external spatial force on body $i$.

On the other hand, using \ref{J3} for the case $j \succ i$:

\begin{equation}
    \frac{\partial [\taubar_{ext}]_{i}}{\partial \q_{j}} =\Phibar_{i}\T\frac{\partial \f_{i,ext}^{C}}{\partial \q_{j}}, (j \succ i)
\label{eqn_IDc_case1}
\end{equation}
Now, switching the indices $i$ and $j$ in \eqref{eqn_IDc_case1} to get $\frac{\partial [\taubar_{ext}]_{j}}{\partial \q_{i}}$ for the case $j \prec i$:
\begin{equation}
        \frac{\partial [\taubar_{ext}]_{j}}{\partial \q_{i}} =\Phibar_{j}\T\frac{\partial \f_{j,ext}^{C}}{\partial \q_{i}}, (j \prec i)
        \label{eqn_IDc_case2}
\end{equation}
Hence, the updated FO derivatives of ID with the new terms are:
\begin{equation}
\begin{aligned}
        \frac{\partial \taubar_{i}}{\partial \q_{j}} &=  \Phibar_{i} \T \big[ 2  \B_{i}^{C} \big]\Psibardot{}_{j} + \Phibar_{i} \T \I_{i}^{C} \Psibarddot{}_{j}  +\Phibar_{i}\T (\f_{i,ext}^{C} \crff  \Phibar_{j}) -\blue{\Phibar_{i}\T\frac{\partial \f_{i,ext}^{C}}{\partial \q_j}}, (j \preceq i)    \\
      \frac{\partial \taubar_{j}}{\partial \q_{i}} &= \Phibar_{j}\T [ 2 \B_{i}^{C} \Psibardot_{i} +  \I_{i}^{C}  \Psibarddot_{i}+(\f_{i}^{C} )\crff \Phibar_{i}  ]-\red{\Phibar_{j}\T \frac{\partial \f_{j,ext}^{C}}{\partial \q_{i}}} , (j \prec i)
\end{aligned}
\label{eqn_ID_FO_contact}
\end{equation}

Note that the quantity $\f_{i,ext}^{C}$ and it's partial derivative w.r.t $\q$ depends on the external force $\lambda$ and the geometry of the contact point on the connectivity tree. To explain this, we assume a rigid-body system with $N$ bodies and a point contact on a link numbered $m$. Then, the external spatial force on $m$ is:
\begin{equation}
    \f_{m,ext} = \left[\begin{array}{c} {}^{0}{\bar{p}_{c}} \times {}^{0}\bar{\lambda}_{c} \\ {}^{0}\bar{\lambda}_{c} \end{array}\right]
\end{equation}
where, ${}^{0}\bar{\lambda}_{c} \in \R^3$ is the 3D cartesian force vector on the link $m$,  while ${}^{0}{\bar{p}_{c}}$ is the cartesian position vector of the point contact, both represented in the ground frame. It's partial derivative w.r.t $\q_{j}$, where $j\preceq m$ is:
\begin{equation}
   \frac{\partial \f_{m,ext}}{\partial \q_{j}} =  \left[\begin{array}{c} -{}^{0}\bar{\lambda}_{c}\times \J_c(:,j) \\ \mathbf{0}_{3 \times n_j} \end{array}\right]
\end{equation}
where, $\J_c(:,j)$ represent the columns of the contact Jacobian $\J_c$ corresponding to the joint $j$, and $n_j$ is the number of DoFs for joint $j$. The cumulative spatial external force $\f_{i,ext}^{C}=\f_{m,ext}$ if $i \preceq m$, and zero otherwise. Hence, the partial derivative of $\f_{i,ext}^{C}$ w.r.t $\q_{j}$ can now be calculated as:
\begin{align}
  \blue{\frac{\partial \f_{i,ext}^{C}}{\partial \q_{j}}}=
    \begin{cases}
       \left[\begin{array}{c} -{}^{0}\bar{\lambda}_{c}\times \J_c(:,j) \\ \mathbf{0}_{3 \times n_j} \end{array}\right] ,  &~~~~ \text{if}\ i \preceq m \\
      \mathbf{0}_{6 \times n_j},  &~~~~ \text{otherwise}
    \end{cases}
    \label{eqn_fic_ext_FO_1}
\end{align}

Switching the indices $i$ and $j$ to get:
\begin{align}
  \red{\frac{\partial \f_{j,ext}^{C}}{\partial \q_{i}}}=
    \begin{cases}
       \left[\begin{array}{c} -{}^{0}\bar{\lambda}_{c}\times \J_c(:,i) \\ \mathbf{0}_{3 \times n_i} \end{array}\right] ,  &~~~~ \text{if}\  j \preceq m \\
      \mathbf{0}_{6 \times n_i},  &~~~~ \text{otherwise}
    \end{cases}
        \label{eqn_fic_ext_FO_2}
\end{align}
The IDSVA algorithm~\cite[Alg.~1]{singh2022efficient} is now updated with the new terms to return the FO partial derivatives of constrained-ID~\eqref{eqn_ID_FO_contact}. The updated algorithm (Alg.~\ref{alg:IDSVAc}) presented only shows the additional terms for the original IDSVA~\cite[Alg.~1]{singh2022efficient} algorithm. Note the additional backward pass to compute the derivatives of $\f_{ext}^{C}$ w.r.t $\q$.

\begin{algorithm}[t]
\small
\caption{IDSVAc}

\begin{algorithmic}[1]  
\REQUIRE $ \q, \,\qd ,\, \qdd,\, {\rm model},\, \f_{ext}, \Jc$
\STATE $...$
\FOR{$i=1$ to $N$}
 \STATE $...$ \\
\STATE $ \f_{i,ext}^{C} =  \f_{i,ext}  $\\[.5ex]  

\ENDFOR

\FOR{$i=N$ to $1$}
 \STATE $j=i$
 \WHILE {$j>0$}
    \STATE CALCULATE $\Phibar_{i}\T\blue{\frac{\partial \f_{i,ext}^{C}}{\partial \q_{j}}},\Phibar_{j}\T\red{\frac{\partial \f_{j,ext}^{C}}{\partial \q_{i}}}$
 \ENDWHILE
    \IF {$\lambda(i) > 0$}
        \STATE $ \f_{\lambda(i),ext}^C = \f_{\lambda(i),ext}^C +  \f_{i,ext}^C $ 
    \ENDIF

\ENDFOR


\FOR{$i=N$ to $1$}
\STATE $...$
\STATE $\t_5[i] = (\f_{i,ext}^{C}\crff)\T  \Phibar_i$\\[.5ex] 

\STATE $\frac{\partial \boldsymbol\tau}{\partial \q}[i, \subtreeb(i)] = ...-\Phibar\T\frac{\partial \f_{ext}^{C}}{\partial \q}[i,\subtreeb(i)]$ \\[1ex]

\STATE $\frac{\partial \boldsymbol\tau}{\partial \q}[\subtree(i),i] =... +  \t_5[\subtree(i)]\T \Phibar_i-\Phibar\T\frac{\partial \f_{ext}^{C}}{\partial \q}[\subtree(i),i]$ \\[1ex]

\STATE $...$

   \STATE $...$
\ENDFOR
\RETURN $\frac{\partial \taubar }{\partial{\q}},\frac{\partial \taubar }{\partial{\qd}}$

\end{algorithmic}
\label{alg:IDSVAc}
\end{algorithm}

\subsection{Constrained ID SO Derivatives}
\label{sec:IDc_SO}

The ID-SO partial derivatives (Sec.~\ref{SO_q_sec}) w.r.t $\q$ are also extended for systems with external forces. The only extra term needed for these derivatives is the SO partial derivative of $[\taubar_{ext}]_{i}$ in Eq.~\eqref{eqn_ID_con}. 
\vspace{15px}
\subsubsection{Case-A}

\vspace{15px}
We consider the Case~\ref{SO_q_case_A}, for the case $k \preceq j \preceq i$. Taking another partial derivative of Eq.~\eqref{JTlam_i_term_1} w.r.t $\q_{k}$:

\begin{equation}
\begin{aligned}
      &  \frac{\partial^2 [\taubar_{ext}]_{i}}{\partial \q_{j} \partial \q_{k}} = -\frac{\partial \Phibar_{i}\T}{\partial \q_{k}}(\f_{i,ext}^{C} \crff \Phibar_{j})-\Phibar_{i}\T\Bigg[\bigg(\frac{\partial \f_{i,ext}^{C}}{\partial \q_{k}}\crffM\bigg) \Phibar_{j}  \Bigg]-\Phibar_{i}\T \Bigg[\f_{i,ext}^{C} \crff \frac{\partial \Phibar_{j}}{\partial \q_{k}} \Bigg] +  \\&~~\frac{\partial \Phibar_{i}\T}{\partial \q_{k}} \frac{\partial \f_{i,ext}^{C}}{\partial \q_{j}} +
      \Phibar_{i}\T \frac{\partial^2 \f_{i,ext}^{C}}{\partial \q_{j} \partial \q_{k}}
\end{aligned}
\end{equation}
Using \ref{PhiT} for the first and fourth term, while \ref{dphii} for the third term:
\begin{equation}
\begin{aligned}
        &\frac{\partial^2 [\taubar_{ext}]_{i}}{\partial \q_{j} \partial \q_{k}} = \Phibar_{i}\T \Phibar_{k} \timesfM (\f_{i,ext}^{C} \crff \Phibar_{j})-\Phibar_{i}\T\Bigg[\bigg(\frac{\partial \f_{i,ext}^{C}}{\partial \q_{k}}\crffM\bigg) \Phibar_{j}  \Bigg]-\Phibar_{i}\T \Bigg[\f_{i,ext}^{C} \crff (\Phibar_{k} \timesM \Phibar_{j}) \Bigg] - \\&~~
        \Phibar_{i}\T \Phibar_{k} \timesfM \frac{\partial \f_{i,ext}^{C}}{\partial \q_{j}} + \Phibar_{i}\T \frac{\partial^2 \f_{i,ext}^{C}}{\partial \q_{j} \partial \q_{k}}
\end{aligned}
\end{equation}
Combining the first and the third terms:
\begin{equation}
\begin{aligned}
       & \frac{\partial^2 [\taubar_{ext}]_{i}}{\partial \q_{j} \partial \q_{k}} = \Phibar_{i}\T \bigg[ \Phibar_{k} \timesfM (\f_{i,ext}^{C} \crff)-\f_{i,ext}^{C} \crff (\Phibar_{k} \timesM)  \bigg] \Phibar_{j}-\Phibar_{i}\T\Bigg[\bigg(\frac{\partial \f_{i,ext}^{C}}{\partial \q_{k}}\crffM\bigg) \Phibar_{j}  \Bigg]- \\&~~
       \Phibar_{i}\T \Phibar_{k} \timesfM \frac{\partial \f_{i,ext}^{C}}{\partial \q_{j}} + \Phibar_{i}\T \frac{\partial^2 \f_{i,ext}^{C}}{\partial \q_{j} \partial \q_{k}}
\end{aligned}
\end{equation}
Using \ref{m11} for the first term and collecting the second and the third term now:
\begin{equation}
    \frac{\partial^2 [\taubar_{ext}]_{i}}{\partial \q_{j} \partial \q_{k}} = \Phibar_{i}\T \Big((\Phibar_{k}\timesfM \f_{i,ext}^{C})\Rten \Big)\crffM \Phibar_{j} -\Phibar_{i}\T\Bigg[\bigg(\frac{\partial \f_{i,ext}^{C}}{\partial \q_{k}}\crffM\bigg) \Phibar_{j}+ \Phibar_{k} \timesfM \frac{\partial \f_{i,ext}^{C}}{\partial \q_{j}}  \Bigg] + \Phibar_{i}\T \frac{\partial^2 \f_{i,ext}^{C}}{\partial \q_{j} \partial \q_{k}}
\end{equation}
Using \ref{m15} for the first term:

\begin{equation}
    \frac{\partial^2 [\taubar_{ext}]_{i}}{\partial \q_{j} \partial \q_{k}} = \Phibar_{i}\T (\f_{i,ext}^{C} \crff \Phibar_{k} )\crffM \Phibar_{j} -\Phibar_{i}\T\Bigg[\bigg(\frac{\partial \f_{i,ext}^{C}}{\partial \q_{k}}\crffM\bigg) \Phibar_{j}+ \Phibar_{k} \timesfM \frac{\partial \f_{i,ext}^{C}}{\partial \q_{j}}  \Bigg] + \Phibar_{i}\T \frac{\partial^2 \f_{i,ext}^{C}}{\partial \q_{j} \partial \q_{k}}
    \label{ID_SO_contact_1}
\end{equation}

Hence the updated derivative $\frac{\partial^2 \taubar_{i}}{\partial \q_{j} \partial \q_{k}}$ (Eq.~\ref{tau_SO_q_1A}) is:

\begin{equation}
\begin{aligned}
     \frac{\partial^2 \taubar_{i}}{\partial \q_{j} \partial \q_{k}} &=  -\Big[\Psibardot_{j}\T [2\Bten{\I_{i}^{C}}{\Phibar_{i}}\Psibardot_{k}]\Rten \Big]\Tten - 2\big[\Phibar_{j}\T ((\B_{i}^{C,T} \Phibar_{i}) \crffM\Psibardot_{k} )\Rten\big] \Tten   - \Big[  \Phibar_{j}\T ((\I_{i}^{C}\Phibar_{i} ) \crffM\Psibarddot_{k} )\Rten \Big]\Tten \\
     &-\Phibar_{i}\T (\f_{i,ext}^{C} \crff \Phibar_{k} )\crffM \Phibar_{j} +\Phibar_{i}\T\Bigg[\bigg(\frac{\partial \f_{i,ext}^{C}}{\partial \q_{k}}\crffM\bigg) \Phibar_{j}+ \Phibar_{k} \timesfM \frac{\partial \f_{i,ext}^{C}}{\partial \q_{j}}  \Bigg] - \Phibar_{i}\T \frac{\partial^2 \f_{i,ext}^{C}}{\partial \q_{j} \partial \q_{k}}, (k \preceq j \preceq i)
     \end{aligned}
     \label{eqn_ID_SO_contact_1}
\end{equation}

Similar to the efficient implementation for the ID-SO derivatives in Sec.~\ref{eff_impl_sec}, we consider Eq.~\eqref{eqn_ID_SO_contact_1} for the $p^{th}$ DoF of joint $i$, $t^{th}$ DoF of joint $j$, and $r^{th}$ DoF of joint $k$ to get:

\begin{equation}
\begin{aligned}
     \frac{\partial^2 \taubar_{i,p}}{\partial \q_{j,t} \partial \q_{k,r}} &=  -\psibardot_{j,t}\T 2\Bmat{\I_i^{C}}{\phibar_{i,p}}\psibardot_{k,r}  - 2\phibar_{j,t}\T ((\B_{i}^{C\T} \phibar_{i,p}) \crff \psibardot_{k,r} )   -  \phibar_{j,t}\T ((\I_{i}^{C}\phibar_{i,p} ) \crff \psibarddot_{k,r} )  \\
     &-\phibar_{i,p}\T (\f_{i,p,ext}^{C} \crff \phibar_{k,r} )\crff \phibar_{j,t} +\phibar_{i,p}\T\Bigg[\bigg(\frac{\partial \f_{i,p,ext}^{C}}{\partial \q_{k,r}}\crff\bigg) \phibar_{j,t}+ \phibar_{k,r} \timesf \frac{\partial \f_{i,p,ext}^{C}}{\partial \q_{j,t}}  \Bigg] - \phibar_{i,p}\T \frac{\partial^2 \f_{i,p,ext}^{C}}{\partial \q_{j,t} \partial \q_{k,r}}, (k \preceq j \preceq i)
     \end{aligned}
\end{equation}

\vspace{15px}
\subsubsection{Case-B}
\vspace{15px}

Here we consider the case-B ($j \prec k \preceq i$) (Sec.~\ref{SO_q_case_B}). We take the partial derivative of Eq.~\eqref{eqn_IDc_case2} w.r.t $\q_{k}$ as:

\begin{equation}
    \frac{\partial^2 [\taubar_{ext}]_{j}}{\partial \q_{i} \partial \q_{k}} = \frac{\partial \Phibar_{j}\T}{\partial \q_{k}} \frac{\partial \f_{j,ext}^{C}}{\partial \q_{i}}+\Phibar_{j}\T \frac{\partial^2 \f_{j,ext}^{C}}{\partial \q_{i} \partial \q_{k}}
\end{equation}
Using \ref{PhiT} results in:
\begin{equation}
    \frac{\partial^2 [\taubar_{ext}]_{j}}{\partial \q_{i} \partial \q_{k}} = \Phibar_{j}\T \frac{\partial^2 \f_{j,ext}^{C}}{\partial \q_{i} \partial \q_{k}}, (j \prec k \preceq i)
\end{equation}
Switching indices $j$ and $k$ to get $\frac{\partial^2 [\taubar_{ext}]_{k}}{\partial \q_{i} \partial \q_{j}}$ for the case $k \prec j \preceq i$:
\begin{equation}
    \frac{\partial^2 [\taubar_{ext}]_{k}}{\partial \q_{i} \partial \q_{j}} = \Phibar_{k}\T \frac{\partial^2 \f_{k,ext}^{C}}{\partial \q_{i} \partial \q_{j}}, (k \prec j \preceq i)
\end{equation}
Hence, the updated term (Eq.~\eqref{tau_SO_q_1B}) now is:

\begin{equation}
    \begin{aligned}
         \frac{\partial^2 \taubar_{k}}{\partial \q_{i} \partial \q_{j}} &= \Phibar_{k}\T \Big( \big[2(\Bten{\I_{i}^{C}}{\Psibardot_{i}}+ \Phibar_{i} \timesfM \B_{i}^{C} - \B_{i}^{C}\Phibar_{i}  \timesM)\Psibardot_{j} + \big( \Phibar_{i} \timesfM \I_{i}^{C}   -\I_{i}^{C}   \Phibar_{i}\timesM \big)\Psibarddot_{j} \big]\Rten +\\&~~~~
          \Phibar_{j} \timesfM \big( 2\B_{i}^{C}  \Psibardot_{i} +\I_{i}^{C} \Psibarddot_{i}+\f_{i}^{C} \crff \Phibar_{i}\big) \Big) -\Phibar_{k}\T \frac{\partial^2 \f_{k,ext}^{C}}{\partial \q_{i} \partial \q_{j}} , (k \prec j \preceq i)
    \end{aligned}
         \label{eqn_ID_SO_contact_2}
\end{equation}

Considering Eq.~\eqref{eqn_ID_SO_contact_2} for the $p^{th}$ DoF of joint $i$, $t^{th}$ DoF of joint $j$, and $r^{th}$ DoF of joint $k$ to get:

\begin{equation}
    \begin{aligned}
         \frac{\partial^2 \taubar_{k,r}}{\partial \q_{i,p} \partial \q_{j,t}} &= \phibar_{k,r}\T \Big( 2(\Bmat{\I_{i}^C}{\psibardot_{i,p}}+ {\phibar_{i,p} \timesf \B_{i}^{C} - \B_{i}^{C}\phibar_{i,p}  \times})\psibardot_{j,t} + {\big( \phibar_{i,p} \timesf \I_{i}^{C}   -\I_{i}^{C}   \phibar_{i,p}\times \big)}\psibarddot_{j,t}  +\\&~~~~
          \phibar_{j,t} \timesf \big( 2\B_{i}^{C}  \psibardot_{i,p} +\I_{i}^{C} \psibarddot_{i,p}+\f_{i}^{C} \crff \phibar_{i,p}\big) \Big)  -\phibar_{k,r}\T \frac{\partial^2 \f_{k,r,ext}^{C}}{\partial \q_{i,p} \partial \q_{j,t}} , (k \prec j \preceq i) 
    \end{aligned}
\end{equation}

\vspace{15px}
\subsubsection{Case-C}
\vspace{15px}

Here we consider the case-C $(j \preceq i \prec k)$ (Sec.~\ref{SO_q_case_C}) for taking the partial derivative of Eq.~\eqref{JTlam_i_term_1} w.r.t. $\q_{k}$ as:
\begin{equation}
\begin{aligned}
        &\frac{[\taubar_{ext}]_{i}}{\partial \q_{j} \partial \q_{k}} = -\frac{\partial \Phibar_{i}\T}{\partial \q_{k}}(\f_{i,ext}^{C} \crff \Phibar_{j})-\Phibar_{i}\T\Bigg[\bigg(\frac{\partial \f_{i,ext}^{C}}{\partial \q_{k}}\crffM\bigg) \Phibar_{j}  \Bigg]-\Phibar_{i}\T \Bigg[\f_{i,ext}^{C} \crff \frac{\partial \Phibar_{j}}{\partial \q_{k}} \Bigg] + \\&~~~~~~\frac{\partial \Phibar_{i}\T}{\partial \q_{k}} \frac{\partial \f_{i,ext}^{C}}{\partial \q_{j}} + 
        \Phibar_{i}\T \frac{\partial^2 \f_{i,ext}^{C}}{\partial \q_{j} \partial \q_{k}}
\end{aligned}
\end{equation}

Using \ref{PhiT} for the first and fourth term, while \ref{dphii} for the third term, results in:

\begin{equation}
    \frac{[\taubar_{ext}]_{i}}{\partial \q_{j} \partial \q_{k}} = -\Phibar_{i}\T\Bigg[\bigg(\frac{\partial \f_{i,ext}^{C}}{\partial \q_{k}}\crffM\bigg) \Phibar_{j}  \Bigg] + \Phibar_{i}\T \frac{\partial^2 \f_{i,ext}^{C}}{\partial \q_{j} \partial \q_{k}}, (j \preceq i \prec k)
\end{equation}

Switching indices $j$ and $k$, and then $i$ and $j$ to get $\frac{\partial^2 [\taubar_{ext}]_{j}}{\partial \q_{k} \partial \q_{i}}$ for the case $k \preceq j \prec i$:

\begin{equation}
    \frac{\partial^2 [\taubar_{ext}]_{j}}{\partial \q_{k} \partial \q_{i}} = -\Phibar_{j}\T\Bigg[\bigg(\frac{\partial  \f_{j,ext}^{C}}{\partial \q_{i}}\crffM\bigg) \Phibar_{k}  \Bigg] + \Phibar_{j}\T \frac{\partial^2 \f_{j,ext}^{C}}{\partial \q_{k} \partial \q_{i}}, (k \preceq j \prec i)
\end{equation}

Hence, the updated term (Eq.~\eqref{tau_SO_q_1C}) now is:

\begin{equation}
\begin{aligned}
  &\frac{\partial^2 \taubar_{j}}{\partial \q_{k} \partial \q_{i}} = 2\Phibar_{j} \T \big(   \Bten{\I_{i}^{C}}{\Psibardot_{i}}+\Phibar_{i} \timesfM \B_{i}^{C}- \B_{i}^{C}(\Phibar_{i} \timesM) \big)\Psibardot_{k} + 
  \Phibar_{j} \T  \big( \Phibar_{i} \timesfM \I_{i}^{C}- \I_{i}^{C}(\Phibar_{i} \timesM)\big) \Psibarddot_{k} +\\
  &~~~~~~~~~~~~~~~~\Phibar_{j}\T\Bigg[\bigg(\frac{\partial  \f_{j,ext}^{C}}{\partial \q_{i}}\crffM\bigg) \Phibar_{k}  \Bigg] - \Phibar_{j}\T \frac{\partial^2 \f_{j,ext}^{C}}{\partial \q_{k} \partial \q_{i}}, (k \preceq j \prec i)
  \end{aligned}
           \label{eqn_ID_SO_contact_3}
\end{equation}

Considering Eq.~\eqref{eqn_ID_SO_contact_3} for the $p^{th}$ DoF of joint $i$, $t^{th}$ DoF of joint $j$, and $r^{th}$ DoF of joint $k$ to get:

\begin{equation}
\begin{aligned}
  &\frac{\partial^2 \taubar_{j,t}}{\partial \q_{k,r} \partial \q_{i,p}} = 2\phibar_{j,t} \T \Big(   \Bmat{\I_{i}^C}{\psibardot_{i,p}}+{\phibar_{i,p} \timesf \B_{i}^{C}- \B_{i}^{C}\phibar_{i,p} \times } \Big)\psibardot_{k,r} + 
  \phibar_{j,t} \T  \Big( {\phibar_{i,p} \timesf \I_{i}^{C}- \I_{i}^{C}(\phibar_{i,p} \times)}\Big) \psibarddot_{k,r}+ \\
  &~~~~~~~~~~~~\phibar_{j,t}\T\Bigg[\bigg(\frac{\partial  \f_{j,t,ext}^{C}}{\partial \q_{i,p}}\crff\bigg) \phibar_{k,r}  \Bigg] - \phibar_{j,t}\T \frac{\partial^2 \f_{j,t,ext}^{C}}{\partial \q_{k,r} \partial \q_{i,p}}, (k \preceq j \prec i)
\end{aligned}
\end{equation}

\vspace{15px}
\subsection{Second-Order Derivative of $\f_{ext}^{C}$:}
\vspace{15px}

To get the SO derivative of $\f_{ext}^{C}$ required in Eq.~\eqref{eqn_ID_SO_contact_1},\eqref{eqn_ID_SO_contact_2},\eqref{eqn_ID_SO_contact_3}, we follow the process similar to the FO case. We assume a single point contact on the link $m$ of a rigid-body system with $N$ joints and $N$ bodies. We take  partial derivative of Eq.~\ref{eqn_fic_ext_FO_1} w.r.t $\q_{k}$ to get: 

\begin{align}
 \frac{\partial^2 \f_{i,ext}^{C}}{\partial \q_{j} \partial \q_{k}}=
    \begin{cases}
       \left[\begin{array}{c} -{}^{0}\bar{\lambda}_{c}\times \frac{\partial \Jc}{\partial \q}(:,j,k) \\ \mathbf{0}_{3 \times n_j \times n_{k}} \end{array}\right] ,  &~~~~ \text{if}\  i \preceq m \\
      \mathbf{0}_{6 \times n_j \times n_{k}},  &~~~~ \text{otherwise}
    \end{cases}
    \label{eqn_fic_SO_1}
\end{align}
where $\frac{\partial \Jc}{\partial \q}(:,j,k) $ is the $j^{th}$ column of the $k^{th}$ page of the tensor $\frac{\partial \Jc}{\partial \q}$. \\

Similarly other cases can be obtained by switching the indices $i$, $j$, and $k$:

\begin{align}
 \frac{\partial^2 \f_{k,ext}^{C}}{\partial \q_{i} \partial \q_{j}}=
    \begin{cases}
       \left[\begin{array}{c} -{}^{0}\bar{\lambda}_{c}\times \frac{\partial \Jc}{\partial \q}(:,i,j) \\ \mathbf{0}_{3 \times n_i \times n_{j}} \end{array}\right] ,  &~~~~ \text{if}\  k \preceq m \\
      \mathbf{0}_{6 \times n_i \times n_{j}},  &~~~~ \text{otherwise}
    \end{cases}
    \label{eqn_fic_SO_2}
\end{align}

\begin{align}
 \frac{\partial^2 \f_{j,ext}^{C}}{\partial \q_{k} \partial \q_{i}}=
    \begin{cases}
       \left[\begin{array}{c} -{}^{0}\bar{\lambda}_{c}\times \frac{\partial \Jc}{\partial \q}(:,k,i) \\ \mathbf{0}_{3 \times n_k\times n_{i}} \end{array}\right] ,  &~~~~ \text{if}\  j \preceq m \\
      \mathbf{0}_{6 \times n_k \times n_{i}},  &~~~~ \text{otherwise}
    \end{cases}
    \label{eqn_fic_SO_3}
\end{align}

Alg.~1 from Ref.~\cite{singh2023second} provides the ID-SO derivatives for ID. This is modified to account for the extra terms developed in this section. Alg.~\ref{alg:IDSVA_SOc} shows the modified IDSVA-SO algorithm to handle the external forces. This is referred to as IDSVA-SOc. Only the modified parts of the algorithm are shown, keeping the rest of the IDSVA-SO algorithm intact. An additional backward pass (in blue) is added to compute the $\f_{ext}^{C}$ and its FO/SO partial derivatives w.r.t~$\q$. In the derivative backward pass, only the modifications are required for the SO derivatives w.r.t $\q$ in the form of three additional terms $p_{3},p_{4},p_{5}$ which are given as: \\
\vspace{25px}
\noindent $p_{3} = \phibar_{i,p}\T\left( (\f_{i,p,ext}^{C} \crff \phibar_{k,r} )\crff \phibar_{j,t} -\Bigg[\bigg(\frac{\partial \f_{i,p,ext}^{C}}{\partial \q_{k,r}}\crff\bigg) \phibar_{j,t}+ \phibar_{k,r} \timesf \frac{\partial \f_{i,p,ext}^{C}}{\partial \q_{j,t}}  \Bigg]+  \frac{\partial^2 \f_{i,p,ext}^{C}}{\partial \q_{j,t} \partial \q_{k,r}}\right)$\\
\vspace{10px}
\noindent $p_{4}=\phibar_{j,t}\T\left(-\frac{\partial  \f_{j,t,ext}^{C}}{\partial \q_{i,p}}\crff \phibar_{k,r}   +  \frac{\partial^2 \f_{j,t,ext}^{C}}{\partial \q_{k,r} \partial \q_{i,p}}\right)$\\
\vspace{10px}
\noindent $p_{5}= \phibar_{k,r}\T \frac{\partial^2 \f_{k,r,ext}^{C}}{\partial \q_{i,p} \partial \q_{j,t}} $
\vspace{10px}

{\normalsize
\begin{algorithm*}[h]
\normalsize
\caption{IDSVA-SOc Algorithm. Only modifications to the original IDSVA-SO algorithm~\cite[Alg.~1]{singh2023second} are shown here.}
\begin{multicols}{2}
\begin{algorithmic}[1]  
\REQUIRE $ \q, \,\qd ,\, \qdd,\, {\rm model},\, \f_{ext}, \Jc$
\STATE $...$
\FOR{$i=1$ to $N$}
 \STATE $...$ \\
\STATE $ \f_{i,ext}^{C} =  \f_{i,ext}  $\\[.5ex]  

\ENDFOR
\blue{\FOR{$i=N$ to $1$}
 \STATE $j=i$
 \WHILE {$j>0$}
    \STATE CALCULATE $\frac{\partial \f_{i,ext}^{C}}{\partial \q_{j}},\frac{\partial \f_{j,ext}^{C}}{\partial \q_{i}}$
        \STATE $k=j$
     \WHILE {$k>0$}
        \STATE CALCULATE $\frac{\partial^2 \f_{i,ext}^{C}}{\partial \q_{j}\partial \q_{k}},\frac{\partial^2 \f_{k,ext}^{C}}{\partial \q_{i}\partial \q_{j}},\frac{\partial^2 \f_{j,ext}^{C}}{\partial \q_{k}\partial \q_{i}}$
     \ENDWHILE
 
 \ENDWHILE
    \IF {$\lambda(i) > 0$}
        \STATE $ \f_{\lambda(i),ext}^C = \f_{\lambda(i),ext}^C +  \f_{i,ext}^C $ 
    \ENDIF

\ENDFOR}


\FOR{$i=N$ to $1$} \label{alg:1a_start}

\FOR{$p=1$ to $n_{i}$} \label{alg:1b_start}
 \STATE $...$ \\

\STATE $j=i$
 \WHILE{$j>0$} \label{alg:2a_start}  
 \FOR{$t=1$ to $n_{j}$} \label{alg:2b_start}
 \STATE $...$ \\

         \STATE $k=j$

         \WHILE{$k>0$} \label{alg:3a_start} 
          \FOR{$r=1$ to $n_{k}$} \label{alg:3b_start}
            \STATE $...$ \\
            \STATE calculate $\red{p_{3}}$
             \STATE $\dtaudqSO{i,p}{j,t}{k,r}=p_{2}-\red{p_{3}}$ \\[\mysp]
              \IF{$j \neq i$}
              \STATE calculate $\red{p_{4}}$\\

                \STATE$\dtaudqSO{j,t}{k,r}{i,p}=\dtaudqSO{j,t}{i,p}{k,r}=\u_{1}\T \psibardot_{r}+\u_{2}\T \psibarddot_{r}-\red{p_{4}}$ \\[\mysp]
                 \STATE $...$ \\

              \ENDIF
              
          
            \IF{$k \neq j$}
               \STATE calculate $\red{p_{5}}$\\
              \STATE $\dtaudqSO{i,p}{k,r}{j,t}=p_{2}-\red{p_{3}}$; \\
              \STATE $\dtaudqSO{k,r}{i,p}{j,t}= \phibar_{r}\T\u_{3}-\red{p_{5}} $ \\[\mysp]
            
                \STATE $...$ \\

            \ELSE
                \STATE $...$ \\
            \ENDIF
    
        \ENDFOR \label{alg:3b_end}

          \STATE $k = \lambda(k)$ 
        
         \ENDWHILE \label{alg:3a_end}
        
     \ENDFOR \label{alg:2b_end}
  \STATE $j = \lambda(j)$ 
 \ENDWHILE \label{alg:2a_end}
    \ENDFOR \label{alg:1b_end}

                  \STATE $...$ \\

\ENDFOR \label{alg:1a_end}

\RETURN $\frac{\partial^{2}\taubar}{\partial \q^{2}},\frac{\partial^{2}\taubar}{\partial \qd^{2}},\frac{\partial^{2}\taubar}{\partial \q \partial \qd}, \frac{\partial \M}{\partial \q}$
\end{algorithmic}
\end{multicols}

\label{alg:IDSVA_SOc}
\end{algorithm*}
}

\newpage
\section{KKT Dynamics FO/SO Derivatives}
\label{sec_KKT_derivs}
The FO/SO derivatives of the KKT dynamics~\cite{budhiraja2019multi} are calculated for multi-DoF joints and models with floating bases. The KKT equation is given as:

\begin{equation}
\underbrace{\left[\begin{array}{cc}
\M & \Jc^{\top} \\
\Jc & \mathbf{0}
\end{array}\right]}_{\K}\left[\begin{array}{c}
\qdd \\
-\lambda
\end{array}\right]=\left[\begin{array}{c}
\S \boldsymbol{\tau}-\b \\
-\Jdc\, \qd
\end{array}\right]
\label{KKT_eqn}
\end{equation}
where $\Jc \in \R^{n_{c}\times n}$ is the contact Jacobian, $\lambda \in \R^{n_{c}}$ is the vector of the contact forces, and $\mathbf{S}$ is the selector matrix to map the actuated joint torques to the entire tree. The Coriolis term $\C\qd$ and the generalized gravitational forces $\g$ are combined together in $\b=\C\qd+\g$, $\K$ is referred to as the KKT Matrix. Eq.~\ref{KKT_eqn} is simplified to get the KKT-Dynamics as:
\begin{equation}
\left[\begin{array}{c}
\qdd \\
-\lambda
\end{array}\right]=\K^{-1} \left[\begin{array}{c}
\S \boldsymbol{\tau}-\b \\
-\Jdc\, \qd
\end{array}\right]
\label{KKT_dyn}
\end{equation}
In the sections below we compute the partial derivative of Eq.~\eqref{KKT_dyn} w.r.t $\u=\{\q,\qd,\taubar\}$.

\vspace{15px}
\subsection{FO Partial Derivative of KKT Dynamics}
\vspace{15px}

Expanding Eq.~\ref{KKT_eqn}:
\begin{align}
    \M\qdd -\Jc\T\lambda = \S\taubar-\b \\
    \Jc\qdd +\dot{\Jc}\qd=0
\end{align}

Taking the partial derivative w.r.t $\u$:

\begin{align}
    \frac{\partial \M}{\partial \u}\qdd+\M\frac{\partial \qdd}{\partial \u} - \frac{\partial \Jc\T}{\partial \u}\lambda - \Jc\T \frac{\partial \lambda}{\partial \u} =  \S\frac{\partial \taubar}{\partial \u}-\frac{\partial \b}{\partial \u}\\
    \frac{\partial \Jc}{\partial \u}\qdd + \Jc\frac{\partial \qdd}{\partial \u} + \frac{\partial (\dot{\Jc}\qd)}{\partial \u} =0
    \label{FO_crude_eqn}
\end{align}

Re-arranging terms

\begin{equation}
\left[\begin{array}{cc}
\M & \Jc^{\top} \\
\Jc & \mathbf{0}
\end{array}\right]\left[\begin{array}{c}
\frac{\partial \qdd}{\partial \u} \\
-\frac{\partial \lambda}{\partial \u}
\end{array}\right] = -\left[\begin{array}{c}
\frac{\partial \M}{\partial \u}\qdd+\frac{\partial \b}{\partial \u}-\frac{\partial \Jc\T}{\partial \u}\lambda-\S\frac{\partial \taubar}{\partial \u}\\
\frac{\partial \Jc}{\partial \u}\qdd+\frac{\partial (\dot{\Jc}\qd)}{\partial \u}
\end{array}\right]
\end{equation}

Upon simplification, we get:

\begin{equation}
\left[\begin{array}{c}
\frac{\partial \qdd}{\partial \u} \\
-\frac{\partial \lambda}{\partial \u}
\end{array}\right] = -\K^{-1}\left[\begin{array}{c}
\frac{\partial \M}{\partial \u}\qdd+\frac{\partial \b}{\partial \u}-\frac{\partial \Jc\T}{\partial \u}\lambda-\S\frac{\partial \taubar}{\partial \u}\\
\frac{\partial \Jc}{\partial \u}\qdd+\frac{\partial (\dot{\Jc}\qd)}{\partial \u}
\end{array}\right]
\label{FO_general_eqn}
\end{equation}
\vspace{25px}

\subsubsection{FO Partial derivative w.r.t $\q$}
\label{sec:FO_stance_q}
\vspace{15px}

Considering $\u=\q$ in Eq.~\ref{FO_general_eqn}:
\begin{equation}
\left[\begin{array}{c}
\frac{\partial \qdd}{\partial \q} \\
-\frac{\partial \lambda}{\partial \q}
\end{array}\right] = -\K^{-1}\left[\begin{array}{c}
\frac{\partial (\M+\b-\Jc\T\lambda)}{\partial \q}-\cancel{\S\frac{\partial \taubar}{\partial \q}}\\
\frac{\partial \Jc}{\partial \q}\qdd+\frac{\partial (\dot{\Jc}\qd)}{\partial \q}
\end{array}\right]
\end{equation}

where the last term cancels out since $\taubar$ is an input and doesn't depend on $\q$. Hence, we get:

\begin{equation}
\left[\begin{array}{c}
\frac{\partial \qdd}{\partial \q} \\
-\frac{\partial \lambda}{\partial \q}
\end{array}\right] = -\K^{-1}\left[\begin{array}{c}
\frac{\partial \textrm{ID}_{c}}{\partial \q}\\
\frac{\partial (\Jc \qdd+\Jdc\qd)}{\partial \q}
\end{array}\right]
\label{ID_FO_q_eqn1}
\end{equation}

where the FO derivative of $\textrm{ID}_c$ function is given by the IDSVAc algorithm (Alg.~\ref{alg:IDSVAc}).

\vspace{15px}

\subsubsection{FO Partials w.r.t $\qd$}
\vspace{15px}

Simplifying Eq.~\ref{FO_general_eqn} for $\u = \qd$, this case results in:

\begin{equation}
\left[\begin{array}{c}
\frac{\partial \qdd}{\partial \qd} \\
-\frac{\partial \lambda}{\partial \qd}
\end{array}\right] = -\K^{-1}\left[\begin{array}{c}
\frac{\partial \textrm{ID}_{c}}{\partial \qd}\\
\frac{\partial (\dot{\Jc}\qd)}{\partial \qd}
\end{array}\right]
\label{ID_FO_v_eqn1}
\end{equation}

\subsubsection{FO Partials w.r.t $\taubar$}
\vspace{15px}

Considering Eq.~\ref{FO_general_eqn} for $\u = \taubar$:
\vspace{15px}

\begin{equation}
\left[\begin{array}{c}
\frac{\partial \qdd}{\partial \taubar} \\
-\frac{\partial \lambda}{\partial \taubar}
\end{array}\right] = -\K^{-1}\left[\begin{array}{c}
\cancel{\frac{\partial \M}{\partial \taubar}\qdd}+\cancel{\frac{\partial \b}{\partial \taubar}}-\cancel{\frac{\partial \Jc\T}{\partial \taubar}\lambda}-\S\frac{\partial \taubar}{\partial \taubar}\\
\cancel{\frac{\partial \Jc}{\partial \taubar}\qdd}+\cancel{\frac{\partial (\dot{\Jc}\qd)}{\partial \taubar}}
\end{array}\right]
\end{equation}

Simplifying terms,

\begin{equation}
\left[\begin{array}{c}
\frac{\partial \qdd}{\partial \taubar} \\
-\frac{\partial \lambda}{\partial \taubar}
\end{array}\right] = \K^{-1}\left[\begin{array}{c}
 \S\mathbf{I}_{n\times n}\\
\mathbf{0}_{n_{c} \times n}
\end{array}\right]
\label{ID_FO_tau_eqn}
\end{equation}

\subsection{SO Partial derivatives of KKT Dynamics}
\vspace{15px}

Taking partial derivative of Eq.~\ref{FO_crude_eqn} w.r.t $\w$:

\begin{equation}
\begin{aligned}
   & \frac{\partial^2 \M}{\partial \u \partial \w}\qdd+\frac{\partial \M}{\partial \w}\frac{\partial \qdd}{\partial \u} +\Bigg( \frac{\partial \M}{\partial \u}\frac{\partial \qdd}{\partial \w}\Bigg) \Rten +\M\frac{\partial^2 \qdd}{\partial \u\partial \w}+\frac{\partial^2 \b}{\partial \u \partial \w} =   \frac{\partial \Jc\T}{\partial \w} \frac{\partial \lambda}{\partial \u}+\Jc\T\frac{\partial ^2 \lambda}{\partial \u \partial \w}+\frac{\partial^2 \Jc\T}{\partial \u \partial \w}\lambda +\\
   &~~~~~~~~~~~~~~~~~~~~~~~~~~~~~~\Bigg(\frac{\partial \Jc\T}{\partial \u} \frac{\partial \lambda}{\partial \w}\Bigg)\Rten +     \S\frac{\partial^2 \taubar}{\partial \u \partial \w} \\
    &\frac{\partial^2 \Jc}{\partial \u \partial \w}\qdd +\Bigg(\frac{\partial \Jc}{\partial \u}\frac{\partial \qdd}{\partial \w}\Bigg)\Rten+\frac{\partial \Jc}{\partial \w}\frac{\partial \qdd}{\partial \u}+\Jc\frac{\partial^2 \qdd}{\partial \u \partial \w} =-\frac{\partial^2 (\dot{\Jc}\qd)}{\partial \u \partial \w}
\end{aligned}
\end{equation}

Re-arranging some terms:
\begin{equation}
\begin{aligned}
   &\M\frac{\partial^2 \qdd}{\partial \u\partial \w}-\Jc\T\frac{\partial ^2 \lambda}{\partial \u \partial \w}+\frac{\partial \M}{\partial \w}\frac{\partial \qdd}{\partial \u} +\Bigg( \frac{\partial \M}{\partial \u}\frac{\partial \qdd}{\partial \w}\Bigg)\Rten - \frac{\partial \Jc\T}{\partial \w} \frac{\partial \lambda}{\partial \u} -\Bigg(\frac{\partial \Jc\T}{\partial \u} \frac{\partial \lambda}{\partial \w}\Bigg)\Rten=  - \frac{\partial^2 \M}{\partial \u \partial \w}\qdd-\\
   &~~~~~~~~~~~~~~~~~~~~~~~~~~\frac{\partial^2 \b}{\partial \u \partial \w}+\frac{\partial^2 \Jc\T}{\partial \u \partial \w}\lambda  +\S\frac{\partial^2 \taubar}{\partial \u \partial \w} \\
    &\Bigg(\frac{\partial \Jc}{\partial \u}\frac{\partial \qdd}{\partial \w}\Bigg)\Rten+\frac{\partial \Jc}{\partial \w}\frac{\partial \qdd}{\partial \u}+\Jc\frac{\partial^2 \qdd}{\partial \u \partial \w} =-\frac{\partial^2 \Jc}{\partial \u \partial \w}\qdd -\frac{\partial^2 (\dot{\Jc}\qd)}{\partial \u \partial \w}
\end{aligned}
\end{equation}

Writing in a simpler form:
\begin{equation}
    \begin{aligned}
  &  \left[\begin{array}{cc}
\M & \Jc^{\top} \\
\Jc & \mathbf{0}
\end{array}\right]\left[\begin{array}{c}
\frac{\partial^2 \qdd}{\partial \u \partial \w} \\
-\frac{\partial^2 \lambda}{\partial \u \partial \w}
\end{array}\right]+   \left[\begin{array}{cccc}
\frac{\partial \M}{\partial \u} & \frac{\partial \M}{\partial \w} & \frac{\partial \Jc\T}{\partial \u} & \frac{\partial \Jc\T}{\partial \w}\\
\frac{\partial \Jc}{\partial \u} & \frac{\partial \Jc}{\partial \w}& \mathbf{0}& \mathbf{0}
\end{array}\right] \left[\begin{array}{c}
\frac{\partial \qdd}{\partial \w} \\
\frac{\partial \qdd}{\partial \u} \\
-\frac{\partial \lambda}{\partial \w} \\
-\frac{\partial \lambda}{\partial \u}
\end{array}\right] = \\
&~~~~~~~~~~~~~~-\left[\begin{array}{c}
\frac{\partial^2 \M}{\partial \u \partial \w}\qdd+\frac{\partial^2 \b}{\partial \u \partial \w}-\frac{\partial^2 \Jc\T}{\partial \u \partial \w}\lambda  -\S\frac{\partial^2 \taubar}{\partial \u \partial \w}\\
\frac{\partial^2 \Jc}{\partial \u \partial \w}\qdd +\frac{\partial^2 (\dot{\Jc}\qd)}{\partial \u \partial \w}
\end{array}\right]
\label{SO_general_eqn}
    \end{aligned}
\end{equation}

Now, different cases by assuming $\u$ and $\w$ from $\{\q,\qd,\taubar \}$ are considered.
\vspace{25px}

\subsubsection{SO partial derivative w.r.t $\q$}
\vspace{15px}

Assuming $\u,\w = \q$ for Eq.~\ref{SO_general_eqn}:

\begin{align}
    \left[\begin{array}{cc}
\M & \Jc^{\top} \\
\Jc & \mathbf{0}
\end{array}\right]\left[\begin{array}{c}
\frac{\partial^2 \qdd}{\partial \q^2 } \\
-\frac{\partial^2 \lambda}{\partial \q^2 }
\end{array}\right]+   \left[\begin{array}{cccc}
\frac{\partial \M}{\partial \q} & \frac{\partial \M}{\partial \q} & \frac{\partial \Jc\T}{\partial \q} & \frac{\partial \Jc\T}{\partial \q}\\
\frac{\partial \Jc}{\partial \q} & \frac{\partial \Jc}{\partial \q}& \mathbf{0}& \mathbf{0}
\end{array}\right] \left[\begin{array}{c}
\frac{\partial \qdd}{\partial \q} \\
\frac{\partial \qdd}{\partial \q} \\
-\frac{\partial \lambda}{\partial \q} \\
-\frac{\partial \lambda}{\partial \q}
\end{array}\right] = -\left[\begin{array}{c}
\frac{\partial^2 \M}{\partial \q^2}\qdd+\frac{\partial^2 \b}{\partial \q^2}-\frac{\partial \Jc\T}{\partial \q^2}\lambda  -\cancel{\S\frac{\partial^2 \taubar}{\partial \q^2}}\\
\frac{\partial^2 \Jc}{\partial \q^2 }\qdd +\frac{\partial^2 (\dot{\Jc}\qd)}{\partial \q^2}
\end{array}\right]
\end{align}

Simplifying:

\begin{align}
    \left[\begin{array}{cc}
\M & \Jc^{\top} \\
\Jc & \mathbf{0}
\end{array}\right]\left[\begin{array}{c}
\frac{\partial^2 \qdd}{\partial \q^2 } \\
-\frac{\partial^2 \lambda}{\partial \q^2 }
\end{array}\right]+   \left[\begin{array}{c}
\frac{\partial \M}{\partial \q}\frac{\partial \qdd}{\partial \q}+  \Big[\frac{\partial \M}{\partial \q}\frac{\partial \qdd}{\partial \q} \Big]\Rten -  \frac{\partial \Jc\T}{\partial \q}\frac{\partial \lambda}{\partial \q} -\Big[ \frac{\partial \Jc\T}{\partial \q}\frac{\partial \lambda}{\partial \q} \Big]\Rten\\
\frac{\partial \Jc}{\partial \q}\frac{\partial \qdd}{\partial \q}+  \Big[\frac{\partial \Jc}{\partial \q}\frac{\partial \qdd}{\partial \q}\Big]\Rten
\end{array}\right]  = -\left[\begin{array}{c}
\frac{\partial^2 \M\qdd}{\partial \q^2}+\frac{\partial^2 \b}{\partial \q^2}-\frac{\partial \Jc\T\lambda}{\partial \q^2} \\
\frac{\partial^2 \Jc\qdd}{\partial \q^2 } +\frac{\partial^2 (\dot{\Jc}\qd)}{\partial \q^2}
\end{array}\right]
\end{align}

Simplifying:

\begin{align}
\left[\begin{array}{c}
\frac{\partial^2 \qdd}{\partial \q^2 } \\
-\frac{\partial^2 \lambda}{\partial \q^2 }
\end{array}\right]   = -\K^{-1}\left[\begin{array}{c}
\frac{\partial^2 \textrm{ID}_{c}}{\partial \q^2} +\frac{\partial \M}{\partial \q}\frac{\partial \qdd}{\partial \q}+  \Big[\frac{\partial \M}{\partial \q}\frac{\partial \qdd}{\partial \q} \Big]\Rten -  \frac{\partial \Jc\T}{\partial \q}\frac{\partial \lambda}{\partial \q} -\Big[ \frac{\partial \Jc\T}{\partial \q}\frac{\partial \lambda}{\partial \q} \Big]\Rten\\
\frac{\partial^2 \Jc\qdd}{\partial \q^2 } +\frac{\partial^2 (\dot{\Jc}\qd)}{\partial \q^2}+\frac{\partial \Jc}{\partial \q}\frac{\partial \qdd}{\partial \q}+  \Big[\frac{\partial \Jc}{\partial \q}\frac{\partial \qdd}{\partial \q}\Big]\Rten
\end{array} \right]
\end{align}
where the SO partial derivative of $\textrm{ID}_{c}$ can be computed using the IDSVA-SOc~\ref{alg:IDSVA_SOc} algorithm. Instead of the $\mathcal{O(N^4)}$ Direct tensor-matrix (DTM) product of $\frac{\partial \M}{\partial \q}$ and $\frac{\partial \qdd}{\partial \q}$ can be computed using the $\mathcal{O}(N^2d)$ IDFOZA~\cite{singh2023second} algorithm.

\vspace{25px}

\subsubsection{SO partial w.r.t $\qd$}
\vspace{15px}

Assuming $\u,\w = \qd$ in Eq.~\ref{SO_general_eqn}:

\begin{align}
    \left[\begin{array}{cc}
\M & \Jc^{\top} \\
\Jc & \mathbf{0}
\end{array}\right]\left[\begin{array}{c}
\frac{\partial^2 \qdd}{\partial \qd^2} \\
-\frac{\partial^2 \lambda}{\partial \qd^2}
\end{array}\right]+   \left[\begin{array}{cccc}
\cancel{\frac{\partial \M}{\partial \qd}} & \cancel{\frac{\partial \M}{\partial \qd}} & \cancel{\frac{\partial \Jc\T}{\partial \qd}} & \cancel{\frac{\partial \Jc\T}{\partial \qd}}\\
\cancel{\frac{\partial \Jc}{\partial \qd}} & \cancel{\frac{\partial \Jc}{\partial \qd}}& \mathbf{0}& \mathbf{0}
\end{array}\right] \left[\begin{array}{c}
\frac{\partial \qdd}{\partial \qd} \\
\frac{\partial \qdd}{\partial \qd} \\
-\frac{\partial \lambda}{\partial \qd} \\
-\frac{\partial \lambda}{\partial \qd}
\end{array}\right] = -\left[\begin{array}{c}
\frac{\partial^2 \M\qdd}{\partial \qd^2}+\frac{\partial^2 \b}{\partial \qd^2}-\cancel{\frac{\partial \Jc\T}{\partial \qd^2}\lambda}  -\cancel{\S\frac{\partial^2 \taubar}{\partial \qd^2}}\\
\cancel{\frac{\partial^2 \Jc}{\partial \qd^2}\qdd} +\frac{\partial^2 (\dot{\Jc}\qd)}{\partial \qd^2}
\end{array}\right]
\end{align}

Simplifying:

\begin{align}
\left[\begin{array}{c}
\frac{\partial^2 \qdd}{\partial \qd^2 } \\
-\frac{\partial^2 \lambda}{\partial \qd^2 }
\end{array}\right]   = -\K^{-1}\left[\begin{array}{c}
\frac{\partial^2 \textrm{ID}_{c}}{\partial \qd^2} \\
\frac{\partial^2 (\dot{\Jc}\qd)}{\partial \qd^2}\end{array}\right]
\end{align}

\vspace{25px}

\subsubsection{Cross SO partial derivatives w.r.t $\qd$, $\q$}
\vspace{15px}

Assuming $\u=\qd$, $\w = \q$ in Eq.~\ref{SO_general_eqn}

\begin{align}
    \left[\begin{array}{cc}
\M & \Jc^{\top} \\
\Jc & \mathbf{0}
\end{array}\right]\left[\begin{array}{c}
\frac{\partial^2 \qdd}{\partial \qd \partial \q} \\
-\frac{\partial^2 \lambda}{\partial \qd \partial \q}
\end{array}\right]+   \left[\begin{array}{cccc}
\cancel{\frac{\partial \M}{\partial \qd}} & \frac{\partial \M}{\partial \q} & \cancel{\frac{\partial \Jc\T}{\partial \qd}} & \frac{\partial \Jc\T}{\partial \q}\\
\cancel{\frac{\partial \Jc}{\partial \qd}} & \frac{\partial \Jc}{\partial \q}& \mathbf{0}& \mathbf{0}
\end{array}\right] \left[\begin{array}{c}
\frac{\partial \qdd}{\partial \q} \\
\frac{\partial \qdd}{\partial \qd} \\
-\frac{\partial \lambda}{\partial \q} \\
-\frac{\partial \lambda}{\partial \qd}
\end{array}\right] = -\left[\begin{array}{c}
\frac{\partial^2 \M}{\partial \qd \partial \q}\qdd+\frac{\partial^2 \b}{\partial \qd \partial \q}-\cancel{\frac{\partial \Jc\T}{\partial \qd \partial \q}\lambda}  -\cancel{\S\frac{\partial^2 \taubar}{\partial \qd \partial \q}}\\
\cancel{\frac{\partial^2 \Jc}{\partial \qd \partial \q}\qdd} +\frac{\partial^2 (\dot{\Jc}\qd)}{\partial \qd \partial \q}
\end{array}\right]
\end{align}

Simplifying:

\begin{align}
    \left[\begin{array}{cc}
\M & \Jc^{\top} \\
\Jc & \mathbf{0}
\end{array}\right]\left[\begin{array}{c}
\frac{\partial^2 \qdd}{\partial \qd \partial \q} \\
-\frac{\partial^2 \lambda}{\partial \qd \partial \q}
\end{array}\right]+   \left[\begin{array}{c}
 \frac{\partial \M}{\partial \q}\frac{\partial \qdd}{\partial \qd} - \frac{\partial \Jc\T}{\partial \q}\frac{\partial \lambda}{\partial \qd}\\
\frac{\partial \Jc}{\partial \q}\frac{\partial \qdd}{\partial \qd}
\end{array}\right] = -\left[\begin{array}{c}
\frac{\partial^2 \textrm{ID}_{c}}{\partial \qd \partial \q}\\
\frac{\partial^2 (\dot{\Jc}\qd)}{\partial \qd \partial \q}
\end{array}\right]
\end{align}

Hence, the final result is:

\begin{align}
\left[\begin{array}{c}
\frac{\partial^2 \qdd}{\partial \qd \partial \q} \\
-\frac{\partial^2 \lambda}{\partial \qd \partial \q}
\end{array}\right]    = -\K^{-1}\left[\begin{array}{c}
\frac{\partial^2 \textrm{ID}_c}{\partial \qd \partial \q} +\frac{\partial \M}{\partial \q}\frac{\partial \qdd}{\partial \qd} - \frac{\partial \Jc\T}{\partial \q}\frac{\partial \lambda}{\partial \qd}\\
\frac{\partial^2 (\dot{\Jc}\qd)}{\partial \qd \partial \q}+\frac{\partial \Jc}{\partial \q}\frac{\partial \qdd}{\partial \qd}
\end{array}\right]
\end{align}
To get the cross-derivative where $\u=\q$ and $\w=\qd$, we 
 use the symmetry property of the Hessian as:
\begin{equation}
      \left[\begin{array}{c}
    \frac{\partial^2 \qdd}{\partial \q \partial \qd} \\[.75ex]
    -\frac{\partial^2 \lambda}{\partial \q \partial \qd}
    \end{array}\right]  =  \Bigg( \left[\begin{array}{c}
    \frac{\partial^2 \qdd}{\partial \qd \partial \q} \\[.75ex]
    -\frac{\partial^2 \lambda}{\partial \qd \partial \q}
    \end{array}\right]  \Bigg)\Rten
\end{equation}
\vspace{25px}

\subsubsection{Cross SO partial derivatives w.r.t $\taubar$ and $\q$}
\vspace{15px}

Assuming $\u=\taubar$ and $\w = \q$ in Eq.~\ref{SO_general_eqn}:

\begin{align}
    \left[\begin{array}{cc}
\M & \Jc^{\top} \\
\Jc & \mathbf{0}
\end{array}\right]\left[\begin{array}{c}
\frac{\partial^2 \qdd}{\partial \taubar \partial \q} \\
-\frac{\partial^2 \lambda}{\partial \taubar \partial \q}
\end{array}\right]+   \left[\begin{array}{cccc}
\cancel{\frac{\partial \M}{\partial \taubar}} & \frac{\partial \M}{\partial \q} & \cancel{\frac{\partial \Jc\T}{\partial \taubar}} & \frac{\partial \Jc\T}{\partial \q}\\
\cancel{\frac{\partial \Jc}{\partial \taubar}} & \frac{\partial \Jc}{\partial \q}& \mathbf{0}& \mathbf{0}
\end{array}\right] \left[\begin{array}{c}
\frac{\partial \qdd}{\partial \q} \\
\frac{\partial \qdd}{\partial \taubar} \\
-\frac{\partial \lambda}{\partial \q} \\
-\frac{\partial \lambda}{\partial \taubar}
\end{array}\right] = -\left[\begin{array}{c}
\cancel{\frac{\partial^2 \M}{\partial \taubar \partial \q}\qdd}+\cancel{\frac{\partial^2 \b}{\partial \taubar \partial \q}}-\cancel{\frac{\partial \Jc\T}{\partial \taubar \partial \q}\lambda}  -\cancel{\S\frac{\partial^2 \taubar}{\partial \taubar \partial \q}}\\
\cancel{\frac{\partial^2 \Jc}{\partial \taubar \partial \q}\qdd} +\cancel{\frac{\partial^2 (\dot{\Jc}\qd)}{\partial \taubar \partial \q}}
\end{array}\right]
\end{align}

Simplifying:

\begin{align}
    \left[\begin{array}{cc}
\M & \Jc^{\top} \\
\Jc & \mathbf{0}
\end{array}\right]\left[\begin{array}{c}
\frac{\partial^2 \qdd}{\partial \taubar \partial \q} \\
-\frac{\partial^2 \lambda}{\partial \taubar \partial \q}
\end{array}\right]+   \left[\begin{array}{c}
 \frac{\partial \M}{\partial \q} \frac{\partial \qdd}{\partial \taubar}-\frac{\partial \Jc\T}{\partial \q}\frac{\partial \lambda}{\partial \taubar}\\
 \frac{\partial \Jc}{\partial \q}\frac{\partial \qdd}{\partial \taubar}\end{array}\right]  = \mathbf{0}
\end{align}

Finally:

\begin{align}
\left[\begin{array}{c}
\frac{\partial^2 \qdd}{\partial \taubar \partial \q} \\
-\frac{\partial^2 \lambda}{\partial \taubar \partial \q}
\end{array}\right]     = -\K^{-1}\left[\begin{array}{c}
 \frac{\partial \M}{\partial \q} \frac{\partial \qdd}{\partial \taubar}-\frac{\partial \Jc\T}{\partial \q}\frac{\partial \lambda}{\partial \taubar}\\
 \frac{\partial \Jc}{\partial \q}\frac{\partial \qdd}{\partial \taubar}\end{array}\right]
\end{align}

To get the cross-derivative where $\u=\q$ and $\w=\taubar$, we 
 use the symmetry property of the Hessian as:
\begin{equation}
      \left[\begin{array}{c}
    \frac{\partial^2 \qdd}{\partial \q \partial \taubar} \\[.75ex]
    -\frac{\partial^2 \lambda}{\partial \q \partial \taubar}
    \end{array}\right]  =  \left( \left[\begin{array}{c}
    \frac{\partial^2 \qdd}{\partial \taubar \partial \q} \\[.75ex]
    -\frac{\partial^2 \lambda}{\partial \taubar \partial \q}
    \end{array}\right]  \right)\Rten
\end{equation}

\newpage

\section{Impact Dynamics FO/SO Derivatives}
\label{sec_impact}
The impact dynamics on a rigid-body system, assuming a coefficient of restitution as e is modeled as~\cite{li2020hybrid}:

\begin{equation}
\left[\begin{array}{cc}
\M & \Jc^{\top} \\
\Jc & \mathbf{0}
\end{array}\right]\left[\begin{array}{c}
\qd_{+} \\
-\lambdah
\end{array}\right]=\left[\begin{array}{c}
\M\qd_{-} \\
e \Jc \qd_{-}
\end{array}\right]
\label{im_eqn_1}
\end{equation}
where $\qd_{-}$ and $\qd_{+}$ are the system velocities before and after the impact respectively, while $\lambdah$ is the impact force at the contact. We assume $e=0$ for rigid contacts, and simplify:
\begin{equation}
\begin{aligned}
    \M\qd_{+} - \Jc\T\lambdah &= \M\qd_{-} \\
    \Jc\qd_{+} &= \mathbf{0}
    \end{aligned}
        \label{im_eqn_4}
\end{equation}
In the following sections, we take the partial derivative of Eq.~\ref{im_eqn_4} w.r.t $\u = \{\q,\qd_{-}\}$ .

\subsection{FO Derivatives}
\vspace{15px}
Taking partial derivatives of \eqref{im_eqn_4} w.r.t $\u$:
\begin{equation}
\begin{aligned}
    \frac{\partial \M}{\partial \u}\qd_{+} + \M \frac{\partial \qd_{+}}{\partial \u} - \frac{\partial \Jc\T}{\partial \u}\lambdah - \Jc\T\frac{\partial \lambdah}{\partial \u} =   \frac{\partial \M}{\partial \u}\qd_{-} +  \M \frac{\partial \qd_{-}}{\partial \u}\\ 
    \frac{\partial \Jc}{\partial \u}\qd_{+} + \Jc\frac{\partial \qd_{+}}{\partial \u}=\mathbf{0} 
\end{aligned}
    \label{FO_im_eqn}
\end{equation}
Combining the two equations to get:
\begin{equation}
\left[\begin{array}{cc}
\M & \Jc^{\top} \\
\Jc & \mathbf{0}
\end{array}\right]\left[\begin{array}{c}
\frac{\partial \qd_{+}}{\partial \u} \\
-\frac{\partial \lambdah}{\partial \u}
\end{array}\right] = -\left[\begin{array}{c}
\frac{\partial \M}{\partial \u}\qd_{+}-\frac{\partial \M}{\partial \u}\qd_{-} -\frac{\partial \Jc\T}{\partial \u}\lambdah-\M \frac{\partial \qd_{-}}{\partial \u}\\
\frac{\partial \Jc}{\partial \u}(\qd_{+})\end{array}\right]
\end{equation}
Finally the general equation is:
\begin{equation}
\left[\begin{array}{c}
\frac{\partial \qd_{+}}{\partial \u} \\
-\frac{\partial \lambdah}{\partial \u}
\end{array}\right] = -\K^{-1}\left[\begin{array}{c}
\frac{\partial \M}{\partial \u}\qd_{+}-\frac{\partial \M}{\partial \u}\qd_{-} -\frac{\partial \Jc\T}{\partial \u}\lambdah-\M \frac{\partial \qd_{-}}{\partial \u}\\
\frac{\partial \Jc}{\partial \u}(\qd_{+})\end{array}\right]
\label{FO_general_eqn_im}
\end{equation}
\vspace{15px}

\subsubsection{Derivatives w.r.t $\q$}
\vspace{15px}
Using $\u=\q$ in Eq.~\ref{FO_general_eqn_im}:
\begin{equation}
\left[\begin{array}{c}
\frac{\partial \qd_{+}}{\partial \q} \\
-\frac{\partial \lambdah}{\partial \q}
\end{array}\right] = -\K^{-1}\left[\begin{array}{c}
\frac{\partial \M}{\partial \q}\qd_{+}-\frac{\partial \M}{\partial \q}\qd_{-} -\frac{\partial \Jc\T}{\partial \q}\lambdah-\cancel{\M \frac{\partial \qd_{-}}{\partial \q}}\\
\frac{\partial \Jc}{\partial \q}(\qd_{+})\end{array}\right]
\end{equation}
Simplifying the expression:

\begin{equation}
\left[\begin{array}{c}
\frac{\partial \qd_{+}}{\partial \q} \\
-\frac{\partial \lambdah}{\partial \q}
\end{array}\right] = -\K^{-1}\left[\begin{array}{c}
\frac{\partial  \M }{\partial \q}(\qd_{+}-\qd_{-})- \frac{\partial (\Jc\T \lambdah)}{\partial \q}\\
\frac{\partial (\Jc \qd_{+})}{\partial \q}
\end{array}\right]
\label{ID_FO_q_im_1}
\end{equation}
Instead of using a tensor-matrix product for the term $\frac{\partial  \M }{\partial \q}(\qd_{+}-\qd_{-})$ in Eq.~\ref{ID_FO_q_im_1}, the IDFOZA~\cite{singh2023second} algorithm with inputs $\qdd = \qd_{+}-\qd_{-}$, $\qd=0$, $\g=0$ can be used. This would reduce the computational complexity from $\mathcal{O}(N^4)$ to $\mathcal{O}(Nd^2)$.

\vspace{15px}

\subsubsection{Derivatives w.r.t $\qd_{-}$}
\vspace{15px}

Using $u=\qd_{-}$ in Eq.~\ref{FO_general_eqn_im}:

\begin{equation}
\left[\begin{array}{c}
\frac{\partial \qd_{+}}{\partial \qd_{-}} \\
-\frac{\partial \lambdah}{\partial \qd_{-}}
\end{array}\right] = -\K^{-1}\left[\begin{array}{c}
\cancel{\frac{\partial \M}{\partial \qd_{-}}\qd_{+}}-\cancel{\frac{\partial \M}{\partial \qd_{-}}\qd_{-}} -\cancel{\frac{\partial \Jc\T}{\partial \qd_{-}}\lambdah}-\M \frac{\partial \qd_{-}}{\partial \qd_{-}}\\
\cancel{\frac{\partial \Jc}{\partial \qd_{-}}(\qd_{+})}\end{array}\right]
\end{equation}

or,

\begin{equation}
\left[\begin{array}{c}
\frac{\partial \qd_{+}}{\partial \qd_{-}} \\
-\frac{\partial \lambdah}{\partial \qd_{-}}
\end{array}\right] = \left[\begin{array}{cc}
\M & \Jc^{\top} \\
\Jc & \mathbf{0}
\end{array}\right]^{-1}\left[\begin{array}{c}
\M\\
\mathbf{0}
\end{array}\right]
\label{ID_FO_v_im_1}
\end{equation}

\subsection{SO Derivatives }
\vspace{15px}

Taking another partial derivative of Eq.~\ref{FO_im_eqn} w.r.t $\w$:

\begin{equation}
\begin{aligned}
    \frac{\partial^2 \M}{\partial \u \partial \w}\qd_{+}+\frac{\partial \M}{\partial \w}\frac{\partial \qd_{+}}{\partial \u} +\left(\frac{\partial \M}{\partial \u}\frac{\partial \qd_{+}}{\partial \w}\right)\Rten +\M\frac{\partial^2 \qd_{+}}{\partial \u\partial \w} -\frac{\partial^2 \Jc\T}{\partial \u \partial \w}\lambdah -\frac{\partial \Jc\T}{\partial \w} \frac{\partial \lambdah}{\partial \u}-\Jc\T\frac{\partial ^2 \lambdah}{\partial \u \partial \w}-\left(\frac{\partial \Jc\T}{\partial \u} \frac{\partial \lambdah}{\partial \w}\right)\Rten=  \\  \frac{\partial ^2 \M}{\partial \u \partial \w}\qd_{-}+\left(\frac{\partial \M}{\partial \u}\frac{\partial \qd_{-}}{\partial \w}\right)\Rten+\frac{\partial \M}{\partial \w}\frac{\partial \qd_{-}}{\partial \u}+\M\frac{\partial^2 \qd_{-}}{\partial \u \partial \w}
\end{aligned}
\end{equation}

\begin{equation}
\frac{\partial^2 \Jc}{\partial \u \partial \w} \qd_{+} +\left(\frac{\partial \Jc}{\partial \u}\frac{\partial \qd_{+} }{\partial \w}\right)\Rten+\frac{\partial \Jc}{\partial \w}\frac{\partial \qd_{+} }{\partial \u}+\Jc\frac{\partial^2 \qd_{+}}{\partial \u \partial \w} = \mathbf{0} 
\end{equation}

Writing in a simpler form:

\begin{equation}
\begin{aligned}
    \left[\begin{array}{cc}
\M & \Jc^{\top} \\
\Jc & \mathbf{0}
\end{array}\right]\left[\begin{array}{c}
\frac{\partial^2 \qd_{+}}{\partial \u \partial \w} \\
-\frac{\partial^2 \lambdah}{\partial \u \partial \w}
\end{array}\right]+   \left[\begin{array}{cccc}
\frac{\partial \M}{\partial \u} & \frac{\partial \M}{\partial \w} & \frac{\partial \Jc\T}{\partial \u} & \frac{\partial \Jc\T}{\partial \w}\\
\frac{\partial \Jc}{\partial \u} & \frac{\partial \Jc}{\partial \w}& \mathbf{0}& \mathbf{0}
\end{array}\right] \left[\begin{array}{c}
\frac{\partial \qd_{+}}{\partial \w} \\
\frac{\partial \qd_{+}}{\partial \u} \\
-\frac{\partial \lambdah}{\partial \w} \\
-\frac{\partial \lambdah}{\partial \u}
\end{array}\right] = \\
-\left[\begin{array}{c}
\frac{\partial^2 \M}{\partial \u \partial \w}(\qd_{+}-\qd_{-})-\frac{\partial^2 \Jc\T}{\partial \u \partial \w}\lambdah  -\left(\frac{\partial \M}{\partial \u}\frac{\partial \qd_{-}}{\partial \w}\right)\Rten-\frac{\partial \M}{\partial \w}\frac{\partial \qd_{-}}{\partial \u}-\M\frac{\partial^2 \qd_{-}}{\partial \u \partial \w}\\
\frac{\partial^2 \Jc}{\partial \u \partial \w}\qd_{+} 
\end{array}\right]
\end{aligned}
\label{SO_im_general_eqn}
\end{equation}
\vspace{15px}

\subsubsection{Partial derivative w.r.t $\q$}
\vspace{15px}

Assuming $\u,\w=\q$, in Eq.~\ref{SO_im_general_eqn}:

\begin{equation}
\begin{aligned}
    \left[\begin{array}{cc}
\M & \Jc^{\top} \\
\Jc & \mathbf{0}
\end{array}\right]\left[\begin{array}{c}
\frac{\partial^2 \qd_{+}}{\partial \q^2} \\
-\frac{\partial^2 \lambdah}{\partial \q^2}
\end{array}\right]+   \left[\begin{array}{cccc}
\frac{\partial \M}{\partial \q} & \frac{\partial \M}{\partial \q} & \frac{\partial \Jc\T}{\partial \q} & \frac{\partial \Jc\T}{\partial \q}\\
\frac{\partial \Jc}{\partial \q} & \frac{\partial \Jc}{\partial \q}& \mathbf{0}& \mathbf{0}
\end{array}\right] \left[\begin{array}{c}
\frac{\partial \qd_{+}}{\partial \q} \\
\frac{\partial \qd_{+}}{\partial \q} \\
-\frac{\partial \lambdah}{\partial \q} \\
-\frac{\partial \lambdah}{\partial \q}
\end{array}\right] = \\
-\left[\begin{array}{c}
\frac{\partial^2 \M}{\partial \q^2}(\qd_{+}-\qd_{-})-\frac{\partial \Jc\T}{\partial \q^2}\lambdah  -\cancel{\left(\frac{\partial \M}{\partial \q}\frac{\partial \qd_{-}}{\partial \q}\right)\Rten}-\cancel{\frac{\partial \M}{\partial \q}\frac{\partial \qd_{-}}{\partial \q}}-\cancel{\M\frac{\partial^2 \qd_{-}}{\partial \q^2}}\\
\frac{\partial^2 \Jc}{\partial \q^2}\qd_{+} 
\end{array}\right]
\end{aligned}
\end{equation}

Simplifying:

\begin{equation}
\begin{aligned}
    \left[\begin{array}{cc}
\M & \Jc^{\top} \\
\Jc & \mathbf{0}
\end{array}\right]\left[\begin{array}{c}
\frac{\partial^2 \qd_{+}}{\partial \q^2} \\
-\frac{\partial^2 \lambdah}{\partial \q^2}
\end{array}\right]+   \left[\begin{array}{c}
\frac{\partial \M}{\partial \q}\frac{\partial \qd_{+}}{\partial \q}+ \Big[ \frac{\partial \M}{\partial \q}\frac{\partial \qd_{+}}{\partial \q} \Big]\Rten-  \frac{\partial \Jc\T}{\partial \q}\frac{\partial \lambdah}{\partial \q} -\Big[ \frac{\partial \Jc\T}{\partial \q}\frac{\partial \lambdah}{\partial \q}\Big]\Rten\\
\frac{\partial \Jc}{\partial \q} \frac{\partial \qd_{+}}{\partial \q}+ \Big[\frac{\partial \Jc}{\partial \q}\frac{\partial \qd_{+}}{\partial \q}\Big]\Rten
\end{array}\right]= 
-\left[\begin{array}{c}
\frac{\partial^2 \M}{\partial \q^2}(\qd_{+}-\qd_{-})-\frac{\partial \Jc\T}{\partial \q^2}\lambdah \\
\frac{\partial^2 \Jc}{\partial \q^2}\qd_{+} 
\end{array}\right]
\end{aligned}
\end{equation}

Finally:

\begin{equation}
\begin{aligned}
\left[\begin{array}{c}
\frac{\partial^2 \qd_{+}}{\partial \q^2} \\
-\frac{\partial^2 \lambdah}{\partial \q^2}
\end{array}\right]   = -\K^{-1}
\left[\begin{array}{c}
\frac{\partial^2 \M}{\partial \q^2}(\qd_{+}-\qd_{-})-\frac{\partial \Jc\T}{\partial \q^2}\lambdah +\frac{\partial \M}{\partial \q}\frac{\partial \qd_{+}}{\partial \q}+ \Big[ \frac{\partial \M}{\partial \q}\frac{\partial \qd_{+}}{\partial \q} \Big]\Rten-  \frac{\partial \Jc\T}{\partial \q}\frac{\partial \lambdah}{\partial \q} -\Big[ \frac{\partial \Jc\T}{\partial \q}\frac{\partial \lambdah}{\partial \q}\Big]\Rten\\
\frac{\partial^2 \Jc}{\partial \q^2}\qd_{+}+\frac{\partial \Jc}{\partial \q} \frac{\partial \qd_{+}}{\partial \q}+ \Big[\frac{\partial \Jc}{\partial \q}\frac{\partial \qd_{+}}{\partial \q}\Big]\Rten
\end{array}\right]
\end{aligned}
\label{eqn_im_SO_q}
\end{equation}

The IDSVA-SOc algorithm (Alg.~\ref{alg:IDSVA_SOc}) gives the SO derivatives of the constrained-ID~\eqref{eqn_IDc} wr.t.~ $\q$. For a fully actuated system, we get:

\begin{equation}
         \frac{\partial^2 \taubar}{\partial \q^2} = \frac{\partial^2 \M}{\partial \q^2}\qdd+ \frac{\partial^2 \b}{\partial \q^2}-\frac{\partial^2 ( \Jc\T \lambda)}{\partial \q^2} = \textrm{IDSVA-SOc} (\q,\qd,\qdd,\lambda)
\end{equation}
For zero $\qd$, and gravity, $\b=\mathbf{0}$, and some arbitrary vector $\qdd=\m$, we get:

\begin{equation}
         \frac{\partial^2 \M}{\partial \q^2}\m -\frac{\partial^2 ( \Jc\T \lambda)}{\partial \q^2} = \textrm{IDSVA-SOc}(\q,\mathbf{0},\m,\lambda) = \textrm{IDSOZAc}(\q,\mathbf{0},\m,\lambda)
         \label{eqn_IDSOZAc}
\end{equation}
This approach is referred to as the \underline{I}nverse \underline{D}ynamics \underline{S}econd-\underline{O}rder \underline{Z}ero \underline{A}lgorithm-\underline{c}onstrained (IDSOZAc), with a complexity of $\mathcal{O}(N^3)$. Eq.~\eqref{eqn_IDSOZAc} replaces the expensive 4D tensor product of $\frac{\partial^2 \M}{\partial \q^2}$ with a vector by this cheaper $\mathcal{O}(N^3)$ IDSOZAc approach. Hence, IDSOZAc with inputs $\qdd = \qd_{+}-\qd_{-}$ can be used in Eq.~\ref{eqn_im_SO_q} to get the term $\frac{\partial^2 \M}{\partial \q^2}(\qd_{+}-\qd_{-})-\frac{\partial \Jc\T}{\partial \q^2}\lambdah $. To get the terms with product of $\frac{\partial \M}{\partial \q}$ with matrices in Eq.~\ref{eqn_im_SO_q}, the IDFOZA~\cite{singh2023second} approach can be used.

\vspace{15px}

\subsubsection{Partial derivatives w.r.t $\qd_{-}$}
\vspace{15px}

Using $\u,\w = \qd_{-}$ in Eq.~\ref{SO_im_general_eqn}:

\begin{equation}
\begin{aligned}
    \left[\begin{array}{cc}
\M & \Jc^{\top} \\
\Jc & \mathbf{0}
\end{array}\right]\left[\begin{array}{c}
\frac{\partial^2 \qd_{+}}{\partial \qd_{-}^2} \\
-\frac{\partial^2 \lambdah}{\partial \qd_{-}^2}
\end{array}\right]+   \cancel{\left[\begin{array}{cccc}
\frac{\partial \M}{\partial \qd_{-}} & \frac{\partial \M}{\partial \qd_{-}} & \frac{\partial \Jc\T}{\partial \qd_{-}} & \frac{\partial \Jc\T}{\partial \qd_{-}}\\
\frac{\partial \Jc}{\partial \qd_{-}} & \frac{\partial \Jc}{\partial \qd_{-}}& \mathbf{0}& \mathbf{0}
\end{array}\right]} \left[\begin{array}{c}
\frac{\partial \qd_{+}}{\partial \qd_{-}} \\
\frac{\partial \qd_{+}}{\partial \qd_{-}} \\
-\frac{\partial \lambdah}{\partial \qd_{-}} \\
-\frac{\partial \lambdah}{\partial \qd_{-}}
\end{array}\right] = \\
-\cancel{\left[\begin{array}{c}
\frac{\partial^2 \M}{\partial \qd_{-}^2}(\qd_{+}-\qd_{-})-\frac{\partial \Jc\T}{\partial \qd_{-}^2}\lambdah  -\frac{\partial \M}{\partial \qd_{-}}\frac{\partial \qd_{-}}{\partial \qd_{-}}-\frac{\partial \M}{\partial \qd_{-}}\frac{\partial \qd_{-}}{\partial \qd_{-}}-\M\frac{\partial^2 \qd_{-}}{\partial \qd_{-}^2}\\
\frac{\partial^2 \Jc}{\partial \qd_{-}^2}\qd_{+} 
\end{array}\right]}
\end{aligned}
\end{equation}

Finally,
\begin{equation}
    \left[\begin{array}{c}
\frac{\partial^2 \qd_{+}}{\partial \qd_{-}^2} \\
-\frac{\partial^2 \lambdah}{\partial \qd_{-}^2}
\end{array}\right]=\mathbf{0}
\end{equation}
This trivial result can also be achieved by taking the partial derivative of the FO derivative w.r.t $\qd$ (Eq.~\ref{ID_FO_v_im_1}). Since Eq.~\ref{ID_FO_v_im_1} depends purely on $\q$, the partial derivative w.r.t~$\qd$ results to zero.

\vspace{15px}

\subsubsection{Cross Partial Derivative w.r.t $\qd_{-}$,$\q$}
\vspace{15px}

Using $\u=\qd_{-}$ and $\w=\q$ in Eq.~\ref{SO_im_general_eqn}:

\begin{equation}
\begin{aligned}
    \left[\begin{array}{cc}
\M & \Jc^{\top} \\
\Jc & \mathbf{0}
\end{array}\right]\left[\begin{array}{c}
\frac{\partial^2 \qd_{+}}{\partial \qd_{-} \partial \q} \\
-\frac{\partial^2 \lambdah}{\partial \qd_{-} \partial \q}
\end{array}\right]+   \left[\begin{array}{cccc}
\cancel{\frac{\partial \M}{\partial \qd_{-}}} & \frac{\partial \M}{\partial \q} & \cancel{\frac{\partial \Jc\T}{\partial \qd_{-}}} & \frac{\partial \Jc\T}{\partial \q}\\
\cancel{\frac{\partial \Jc}{\partial \qd_{-}}} & \frac{\partial \Jc}{\partial \q}& \mathbf{0}& \mathbf{0}
\end{array}\right] \left[\begin{array}{c}
\frac{\partial \qd_{+}}{\partial \q} \\
\frac{\partial \qd_{+}}{\partial \qd_{-}} \\
-\frac{\partial \lambdah}{\partial \q} \\
-\frac{\partial \lambdah}{\partial \qd_{-}}
\end{array}\right] = \\
-\left[\begin{array}{c}
\cancel{\frac{\partial^2 \M}{\partial \qd_{-} \partial \q}(\qd_{+}-\qd_{-})}-\cancel{\frac{\partial \Jc\T}{\partial \qd_{-} \partial \q}\lambdah}  -\cancel{\frac{\partial \M}{\partial \qd_{-}}\frac{\partial \qd_{-}}{\partial \q}}-\frac{\partial \M}{\partial \q}\mathbf{I}-\cancel{\M\frac{\partial^2 \qd_{-}}{\partial \qd_{-} \partial \q}}\\
\cancel{\frac{\partial^2 \Jc}{\partial \qd_{-} \partial \q}\qd_{+}} 
\end{array}\right]
\end{aligned}
\end{equation}

Simplifying:

\begin{equation}
\begin{aligned}
    \left[\begin{array}{cc}
\M & \Jc^{\top} \\
\Jc & \mathbf{0}
\end{array}\right]\left[\begin{array}{c}
\frac{\partial^2 \qd_{+}}{\partial \qd_{-} \partial \q} \\
-\frac{\partial^2 \lambdah}{\partial \qd_{-} \partial \q}
\end{array}\right]+   \left[\begin{array}{c}
 \frac{\partial \M}{\partial \q}\frac{\partial \qd_{+}}{\partial \qd_{-}}- \frac{\partial \Jc\T}{\partial \q}\frac{\partial \lambdah}{\partial \qd_{-}}\\
 \frac{\partial \Jc}{\partial \q}\frac{\partial \qd_{+}}{\partial  \qd_{-}}
\end{array}\right]  =
\left[\begin{array}{c}
\frac{\partial \M}{\partial \q}\mathbf{I}\\
\mathbf{0} 
\end{array}\right]
\end{aligned}
\end{equation}

Finally,

\begin{equation}
\begin{aligned}
   \left[\begin{array}{c}
\frac{\partial^2 \qd_{+}}{\partial \qd_{-} \partial \q} \\
-\frac{\partial^2 \lambdah}{\partial \qd_{-} \partial \q}
\end{array}\right]   = \K^{-1}
\left[\begin{array}{c}
\frac{\partial \M}{\partial \q}\mathbf{I}_{n \times n}- \frac{\partial \M}{\partial \q}\frac{\partial \qd_{+}}{\partial \qd_{-}}+ \frac{\partial \Jc\T}{\partial \q}\frac{\partial \lambdah}{\partial \qd_{-}}\\
-\frac{\partial \Jc}{\partial \q}\frac{\partial \qd_{+}}{\partial \qd_{-}}
\end{array}\right]
\end{aligned}
\end{equation}

The cross-derivative when $\u=\q$, $\w=\qd_{-}$ can be obtained by using the Hessian symmetry as:

\begin{equation}
     \left[\begin{array}{c}
\frac{\partial^2 \qd_{+}}{\partial \q \partial \qd_{-}} \\
-\frac{\partial^2 \lambdah}{\partial \q \partial \qd_{-}}
\end{array}\right]  = \left(  \left[\begin{array}{c}
\frac{\partial^2 \qd_{+}}{\partial \qd_{-} \partial \q} \\
-\frac{\partial^2 \lambdah}{\partial \qd_{-} \partial \q}
\end{array}\right] \right)\Rten
\end{equation}

\bibliographystyle{IEEEtran}
\bibliography{ref.bib}

\newpage

\section{Summary}
The details of SO partial derivatives of Inverse/Forward dynamics are provided in these notes. An efficient implementation is described and is used to develop an associated algorithm, that can be found in the paper~\cite{singh2023second} associated with these notes. The derivatives of the KKT dynamics and impact dynamics are also provided.\\

\bibliographystyle{IEEEtran}


\end{document}